\documentclass{article}

\usepackage[preprint]{neurips_2022}

\usepackage{microtype}
\usepackage[utf8]{inputenc} 
\usepackage[T1]{fontenc}    
\usepackage{hyperref}       
\usepackage{url}            
\usepackage{booktabs}       
\usepackage{amsfonts}       
\usepackage{nicefrac}       
\usepackage{microtype}      
\usepackage{xcolor}         
\usepackage{graphicx}
\usepackage{subcaption}
\usepackage{algorithm}
\usepackage{algpseudocode}
\usepackage{wrapfig}
\usepackage{enumitem}

\title{Continual Pre-Training Mitigates Forgetting in Language and Vision}

\author{%
  Andrea Cossu\thanks{Corresponding author} \\
  Scuola Normale Superiore\\
  \texttt{andrea.cossu@sns.it} \\
  \And
  Tinne Tuytelaars\\
  PSI, ESAT\\
  KU Leuven\\
  \texttt{tinne.tuytelaars@kuleuven.be}\\
  \And
  Antonio Carta
  \qquad
  Lucia Passaro
  \qquad
  Vincenzo Lomonaco
  \qquad
  Davide Bacciu\\
  {Computer Science Department} \\
  {University of Pisa}\\
  \texttt{\{antonio.carta, lucia.passaro, vincenzo.lomonaco, davide.bacciu\}@unipi.it}
}

\begin{document}

\maketitle

\begin{abstract}
Pre-trained models are nowadays a fundamental component of machine learning research. In continual learning, they are commonly used to initialize the model before training on the stream of non-stationary data. However, pre-training is rarely applied during continual learning. We formalize and investigate the characteristics of the continual pre-training scenario in both language and vision environments, where a model is continually pre-trained on a stream of incoming data and only later fine-tuned to different downstream tasks. We show that continually pre-trained models are robust against catastrophic forgetting and we provide strong empirical evidence supporting the fact that self-supervised pre-training is more effective in retaining previous knowledge than supervised protocols. Code is provided at \url{https://github.com/AndreaCossu/continual-pretraining-nlp-vision}.
\end{abstract}

\section{Introduction}
Continual Learning (CL) \citep{lesort2020} focuses on the design of agents able to learn from a stream of non-stationary data while preserving previously acquired knowledge. The tendency of neural networks to catastrophically forget when confronted with new data has been the subject of many studies \citep{mccloskey1989, french1999}, mostly focused on the design of new CL strategies that mitigate such problem \citep{delange2021}. The traditional CL scenario currently used in the literature considers a single model tackling a sequence of tasks, one after the other \citep{parisi2019}. In this setting, the CL model needs to learn its features while, \emph{at the same time}, leveraging the same features to solve the supervised task.
However, this scenario is not the only conceivable one. Natural Language Processing (NLP), for example, often exploits Transfer Learning techniques \citep{ruder2019} implemented through the so-called {\it pre-training fine-tuning} setup. In this setting, the more general linguistic knowledge acquired with pre-training is leveraged as a starting point to target specific downstream tasks. Specifically: 1) during pre-training, language models focus on unsupervised learning tasks (e.g. predicting masked words based on the surrounding context), and 2) during fine-tuning, the pre-trained model is further trained on supervised learning tasks (e.g. sequence labeling). 
Pre-trained models are widespread also in CL \citep{mehta2021, wu2021a}, where they are mostly used to conveniently initialize the model weights before learning from the non-stationary stream of data. However, the generality and robustness of the neural representations features may be greatly impaired during the continual training on the sequence of tasks, since the model will tend to overfit to the tasks objective. 
By separating the goal of building robust features from that of solving the task \emph{during} the continual training, we provide a new way to design continual learning models which are 1) kept continuously up-to-date over time and 2) more robust to catastrophic forgetting since pre-trained features have been reported to be subjected to softer drifts during adaptation to the task \citep{mehta2020, ramasesh2021}.
\begin{wrapfigure}{r}{0.55\linewidth}
    \centering
  \includegraphics[width=0.55\textwidth]{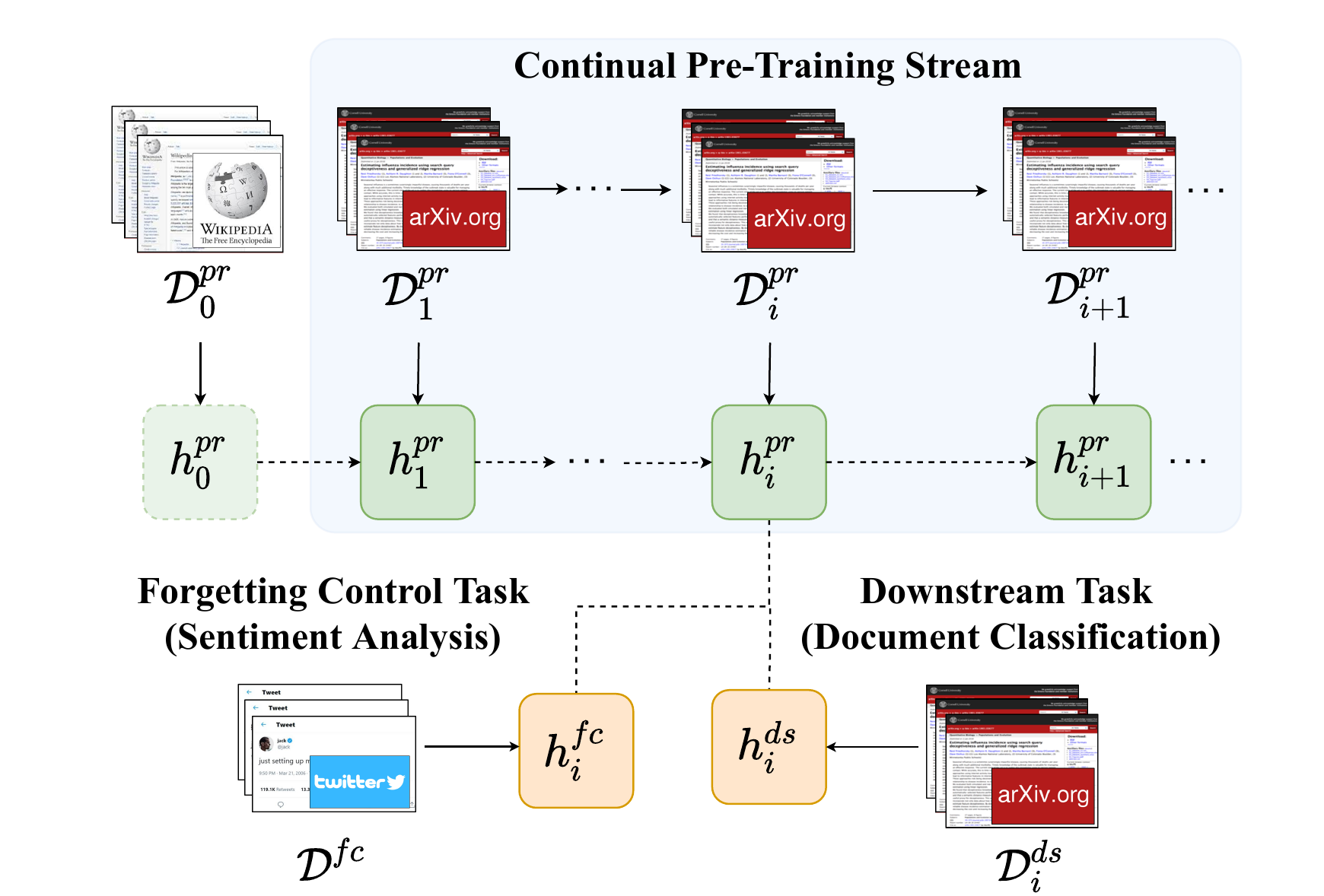}
  \caption{The Continual Pre-training scenario. During each stage (experience) $i$ of continual pre-training (top), the model $h^{pr}_i$ is pre-trained (center) on the dataset $\mathcal{D}^{pr}_i$ (e.g., \emph{scientific abstracts}). Subsequently (bottom), the model is fine-tuned against one (or more) downstream task $\mathcal{D}^{ds}_i$ (e.g. \emph{scientific abstracts} classification). Forgetting is measure by fine-tuning on $\mathcal{D}^{fc}$ (e.g. \emph{sentiment analysis}). At each stage, only the current pre-trained and downstream datasets/models are available.}
  \label{fig:scenario}
\end{wrapfigure}
The former point can be better understood with an example: let us consider the case in which a model is pre-trained on a snapshot of Wikipedia containing articles up to 2018. Part of the knowledge contained inside the model will soon become outdated: on one hand, the information contained in the original articles is likely to be replaced with up-to-date versions (e.g., changes in public figures such as a new President). On the other hand, outdated models do not incorporate the semantics of concepts related to more recent events. For example, the semantics of a term like COVID-19, which becomes important in a short amount of time, cannot be incorporated in the model without additional pre-training. As a consequence, an outdated language model may perform worse on tasks like language generation and Question Answering (Q/A), since it will not be able to generate sentences related to recent events \citep{jang2022}. \\
In this paper, we formalize and study the continual pre-training scenario (Figure \ref{fig:scenario}), where the model is continuously updated via an appropriate pre-training objective on a non-stationary stream of (possibly unlabeled) data. After each stage of pre-training, we build a new model from the pre-trained one (e.g., by substituting its final classifier head) and we train it on a number of downstream tasks. We monitor whether continual pre-training improves/worsens the performance on tasks which are similar/different with respect to the ones encountered during continual pre-training. We are particularly interested in studying the possible deterioration, which represents catastrophic forgetting. For the sake of the evaluation, we specifically introduce a Forgetting Control (FC) dataset as one of the downstream tasks. The FC dataset contains samples different from the ones present in the non-stationary stream and more similar to the dataset used for the original pre-training phase prior to continual training. Against this FC dataset we compare the performance of the pre-trained model at the beginning of the sequence of tasks with the performance of the model after each stage of continual pre-training. 
Our aim is \emph{to investigate the behavior of different architectures, pre-training protocols and input modalities in the continual pre-training scenario and how these factors impact on catastrophic forgetting}. In order to explore this broad research question:
\begin{enumerate}[noitemsep,topsep=0pt,parsep=0pt,partopsep=0pt]
    \item we formally define the continual pre-training scenario and we describe an evaluation methodology to assess the impact of catastrophic forgetting (Section \ref{sec:scenario});
    \item we build two evaluation environments based on Natural Language Processing (NLP) and Computer Vision (CV) tasks (Sections \ref{sec:nlp-env} and \ref{sec:cv-env}, respectively). We thoroughly study them by using different datasets, models architectures and pre-training protocols;
    \item we show that unsupervised/self-supervised pre-training protocols play a fundamental role in the mitigation of forgetting, while supervised protocols hurt the performance. The role of the architecture type and depth does not have an equivalent impact (Section \ref{results-architecture});
    \item we study the feature space of our pre-trained models by using linear evaluation and Centered Kernel Alignment \citep{kornblith2019} (Section \ref{results-features}). We observe that keeping the hidden features fixed during linear evaluation exacerbates forgetting for supervised pre-training. Supervised pre-training also causes a larger drift in the feature space compared to self-supervised pre-training.
\end{enumerate}

\section{Related Works} \label{sec:related}
The ability of pre-trained models to solve a diverse set of tasks through fine-tuning has led to consider them as almost static models.
However, it was recently shown that taking a pre-trained model and performing an additional step of pre-training on domain-specific data is beneficial for the downstream performance in that domain (e.g., Q/A in bio-medicine as showed by \cite{gururangan2020, lee2020}). Pre-trained models are helpful also in CL, where leveraging a pre-trained model as the starting point for the continual training leads to better results with respect to forgetting both in CV \citep{mehta2021, ramasesh2021} and NLP \citep{wu2021a}, especially when combined with CL strategies. An additional pre-training step before the continual training also provides advantages in terms of downstream performance on domain-specific tasks \citep{rongali2021}.\\
The need to perform continual pre-training is present in many different applications, where updating the pre-trained model is fundamental to incorporate new knowledge and update or erase outdated information \citep{lazaridou2021, han2021, jang2022}. While models trained directly on a domain task may achieve similar or even better performance on downstream tasks \citep{gu2021}, the cost of starting from scratch each time is large and mitigating it is one of the objectives of CL. 
Continual pre-training has been recently explored in the context of NLP by leveraging either domain-specific datasets (like multi-domain research papers) \citep{jin2021a} or news/tweets corpora split into different temporal segments \citep{loureiro2022, jang2021}. The results show that continual pre-training is beneficial to the downstream performance and that forgetting on the tasks stream can be effectively mitigated by employing CL strategies. Moreover, continual pre-training is also able to provide advantages in terms of temporal generalization on unseen future data \citep{loureiro2022} and event temporal reasoning \citep{han2021}. 
The work by \cite{hu2021a} focuses on the performance difference between contrastive self-supervised (MoCo-v2 by \cite{chen2020b}) and supervised pre-training in CV, showing that self-supervised leads to robust features in terms of forgetting. A more detailed discussion of related works is presented in Appendix \ref{app:extended-rel}.
Our work provides new evidence of the behavior of pre-trained models in the continual pre-training scenario. We propose to evaluate the performance in terms of catastrophic forgetting on a FC dataset not present in the CL stream. We provided results for both CV and NLP, with experiments on longer streams than most of the existing studies (with the exception of \cite{qin2022}). Unlike prior works, we did not use any CL strategy, but we just employed naive fine-tuning.

\section{Continual Pre-Training Scenario} \label{sec:scenario}
The traditional CL scenario \citep{lomonaco2021} trains a model $h_0$ on a (possibly infinite) stream of experiences $\mathcal{S} = (e_1, e_2, e_3, \ldots)$, where each experience $e_i$ brings a dataset $\mathcal{D}_i$, representing the current task. The model is trained on $\mathcal{S}$, one experience after the other, and needs to address the non-stationarity and drifts occurring between experiences without having access to the previously encountered data. The model $h_0$ is sometimes initialized with the weights of a pre-trained model. The pre-training phase is conducted on the dataset $\mathcal{D}^{pr}$ which is however not available during CL.

We provide a formal characterization of the continual pre-training scenario (pseudo-code in Appendix \ref{app:pseudo}) and highlight the differences with respect to the traditional CL setup. 
The continual pre-training scenario leverages a model $h^{pr}_0$ originally pre-trained on dataset $\mathcal{D}_0^{pr}$, not available anymore. The model is presented with a (possibly infinite) stream of experiences, where each experience $e_i$ brings a dataset $\mathcal{D}_i^{pr}$ for pre-training and a downstream dataset $\mathcal{D}_i^{ds}$ for fine-tuning.
For each experience $e_i$, the last pre-trained model $h_{i-1}^{pr}$ is further pre-trained on $\mathcal{D}_i^{pr}$. After the pre-training step, the model $h_i^{pr}$ is fine-tuned on $\mathcal{D}_i^{ds}$, resulting in $h_i^{ds}$. We adopt naive fine-tuning, without any CL strategies.
In order to measure catastrophic forgetting, we leverage a FC dataset $\mathcal{D}^{fc}$ in place of the $\mathcal{D}_0^{pr}$ originally used during the first pre-training phase. 
While each $\mathcal{D}_i^{ds}$ contains samples similar to the ones encountered during pre-training, the FC dataset contains knowledge more similar to the one in $\mathcal{D}_0^{pr}$ than the one in $\bigcup_{i=1,2,3,\ldots} \mathcal{D}_i^{pr}$. Forgetting is assessed after each experience $e_i$ by comparing the performance of $h^{pr}_0$ fine-tuned on $\mathcal{D}^{fc}$ with the performance of $h^{pr}_i$ fine-tuned on the same dataset. We use $h_i^{ds}$ to verify that the continual pre-training step actually contributes to learning meaningful features for the downstream task. In this way we avoid the uninteresting case where pre-training leaves features (mostly) unchanged, resulting in no catastrophic forgetting of previous knowledge but also in a lower performance on the downstream task. It is important to note that the head (last layer of the model) used during pre-training is not the one used during fine-tuning. In fact, the pre-training and downstream tasks are different ones and they therefore require different heads. Before fine-tuning on each downstream task, the head of $h_{i}^{pr}$ is replaced with a randomly initialized head. The model is then trained until convergence to obtain $h_{i}^{ds}$. During the continual pre-training step instead, the head is not replaced.
\begin{wraptable}{r}{0.45\linewidth}
    \caption{Combinations for the main components of the continual pre-training scenario explored in this paper. MLM=Masked Language modeling, MIM=Masked Image Modeling, CLF=Image Classification.}
    \label{tab:combinations}
    \centering
    \small
    \setlength{\tabcolsep}{3pt}
    \begin{tabular}{ccc}
\toprule
\textbf{Pre-training} & \textbf{Architecture} & \textbf{Data} \\ \midrule
Unsupervised (MLM) & Transformer & Words \\ \midrule
Unsupervised (MIM) & Transformer & Images \\ \midrule
Supervised (CLF) & Transformer & Images \\ \midrule
Supervised (CLF) & CNN & Images  \\
\bottomrule
\end{tabular}
\end{wraptable}
The continual pre-training scenario has different characteristics with respect to the traditional CL setup. Firstly, the continual pre-training scenario updates continuously the pre-trained model and then adapts it to specific tasks. The traditional CL setup does not consider this important distinction, using the same model both for representation learning and to solve incoming tasks.  
Secondly, model evaluation in continual pre-training requires an additional training phase on the target task, while CL usually requires the model to be readily able to tackle all tasks seen so far without any additional training. Therefore, the model has to focus on the new task without the opportunity to build robust, general features via pre-training protocols. As our results will show, the additional cost of a training phase in continual pre-training can be largely mitigated by a quick adaptation phase (e.g., one epoch of training). In fact, fast remembering of previous knowledge is considered one of the objectives of CL \citep{hadsell2020}. \\
Ultimately, \emph{our continual pre-training scenario aims at building models which are general learners, able to quickly adapt to unseen data while still preserving the original knowledge}. 
We studied continual pre-training by introducing two evaluation environments: one for NLP and one for CV. They are designed to investigate the impact on forgetting of specific components of the scenario (Table \ref{tab:combinations}), namely the input modality, the pre-training protocol and the model architecture.

\subsection{Natural Language Processing Environment} \label{sec:nlp-env}
Current NLP applications are all based on the idea of leveraging large-scale pre-trained models to then solve different tasks under fine-tuning, few- or even zero-shot learning settings. Therefore, NLP applications based on the traditional pre-training fine-tuning setting seem to be the most natural field for evaluating our continual pre-training scenario. For example, when dealing with a stream of news, it is important to keep the language model updated \citep{lazaridou2021} so that it can incorporate information which was not previously available. 
Our NLP environment employs an unsupervised/self-supervised pre-training protocol and different Transformer architectures \citep{vaswaniAttentionAllYou2017}. These components are standard ones in NLP and represent the state of the art of the field. We uses the popular pre-trained Transformers RoBERTa \citep{liu2019b} and BERT \citep{devlin2019}, pre-trained on Wikipedia. In addition, we study a variant of RoBERTa in which the vocabulary is dynamically expanded with the addition of new tokens. We select the most frequent tokens of the continual pre-training dataset which were not present in the pre-trained tokenizer. Vocabulary expansion is beneficial for downstream performance, as showed by recent works on dynamic token expansion in both CV \citep{douillard2021} and NLP \citep{zhang2020, han2021}. Our aim is to understand whether the addition of new tokens may result in a larger forgetting of existing knowledge. 
We apply continual pre-training on a dataset of \texttt{scientific abstracts} from arXiv \citep{geiger2019}. The motivation behind the choice of this dataset is that \texttt{scientific abstracts} represent a very specific domain for NLP both in terms of syntactic structures and domain-specific terminology. Indeed, updating the language model before fine-tuning is particularly beneficial under these circumstances. The downstream task is modeled as a document classification problem aiming to associate \texttt{scientific abstracts} to their corresponding arXiv classes. The CL stream includes $5$ experiences, with $2$ scientific domains (classes) in each experience (as in common CL benchmarks like Split-MNIST/CIFAR-10). Please, refer to Appendix \ref{app:setup} for a complete description of the split used for pretraining and downstream fine-tuning. We test two different FC datasets to measure forgetting: \texttt{sentiment analysis} from tweets and Question Answering Natural Language Inference (\texttt{QNLI}). The idea behind these choices is that the dataset of \texttt{scientific abstracts} should not contain much knowledge neither about sentiments, nor about generic facts for language inference. Pre-training on scientific abstracts may therefore disrupt the knowledge contained in the original language model. 
We additionally expand our analysis by using the $20$ datasets present in the \texttt{SentEval} benchmark \citep{CONNEAU18.757} as FC datasets. 
\begin{table}[ht]
    \caption{Accuracy on the entire dataset of \texttt{sentiment analysis} with RoBERTa model. Continual pre-training has been performed sequentially over each experience of \texttt{scientific abstracts}. Base refers to the model pre-trained on Wikipedia, while NT refers to the model with vocabulary expansion.}
    \label{tab:cl-tweets}
    \centering
    \small
    \begin{tabular}{lccccc|ccccc}
\toprule
\textbf{RoBERTa} & \multicolumn{5}{c}{\textbf{Accuracy}} & \multicolumn{5}{c}{\textbf{1-epoch Accuracy}}  \\ \toprule
\textbf{Base} & \multicolumn{5}{c}{93.40} &  \multicolumn{5}{c}{92.40} \\ \midrule
Exp.      & e1 & e2 & e3 & e4 & e5         & e1 & e2 & e3 & e4 & e5        \\ \midrule
\textbf{Pretr}            & 93.40 &  93.15  & 93.35  &  93.20  & 92.90             & 92.40  & 91.80   &  92.30  &   91.85 &  92.20                \\ \midrule
\textbf{Pretr. NT}            & 93.75 &  93.70  &  93.75  &  93.60  & 94.10            & 91.75   &  91.15  &  92.00  &  92.30  &    92.45                 \\ \bottomrule
\end{tabular}
\end{table}
\begin{table}
    \caption{Accuracy on the entire dataset of \texttt{QNLI} with RoBERTa model. Continual pre-training has been performed sequentially over each experience of \texttt{scientific abstracts}. Base refers to the model pre-trained on Wikipedia, while NT refers to the model with expanding vocabulary.}
    \label{tab:cl-qnli}
    \small
\centering
\begin{tabular}{lccccc|ccccc}
\toprule
\textbf{RoBERTa} & \multicolumn{5}{c}{\textbf{Accuracy}} & \multicolumn{5}{c}{\textbf{1-epoch Accuracy}}  \\ \toprule
\textbf{Base} & \multicolumn{5}{c}{92.73} &  \multicolumn{5}{c}{91.76} \\ \midrule
Exp.      & e1 & e2 & e3 & e4 & e5         & e1 & e2 & e3 & e4 & e5        \\ \midrule
\textbf{Pretr.}            & 91.96   &  91.87  & 91.96   &  91.76  &  92.07          &  90.68  & 91.32    &  90.70  & 90.83    & 90.85                     \\ \midrule
\textbf{Pretr. NT} &  92.09  &  91.62  & 91.31    &  91.45  & 91.51           &  91.49  & 91.05 & 91.31   & 89.99    & 90.99 \\ \bottomrule
\end{tabular}
\end{table}
\subsection{Computer Vision Environment} \label{sec:cv-env}
We found CV to be a useful test-bed to disentangle the importance of the three components in our continual pre-training scenario. In particular, we design the CV environment to understand \emph{to what extent forgetting depends on the input modality (natural language against vision), on the architecture (Transformer against CNN) and on the pre-training protocol (unsupervised/self-supervised against supervised)}.
To limit the large number of experiments needed to explore these three factors, in the CV environment we do not measure the performance on the downstream task after each step of continual pre-training. Instead, we focus on the study of forgetting on the FC dataset.
In fact, the impact of pre-training on downstream tasks similar to the ones in the pre-training stream is assessed both in the discussion of related works (Section \ref{sec:related} above) and in the experiments with \texttt{scientific abstracts} classification in NLP environment (results presented below in Section \ref{sec:results} and Appendix \ref{app:downstream}).\\
The CV environment uses \texttt{iNaturalist} \citep{vanhorn2018} for continual pre-training and \texttt{CORe50} \citep{lomonaco2017} as FC dataset for catastrophic forgetting. We use ResNet101, Vision Transformer (ViT) and BEiT originally pre-trained on ImageNet. The choice of ResNet and ViT is fundamental to disentangle the role of the architecture (NLP uses only Transformers) and the pre-training protocol (NLP uses only self-supervised pre-training). In fact, ResNet \citep{he2016} and ViT \citep{dosovitskiy2020} are pre-trained via \emph{supervised image classification}. The choice of BEiT \citep{bao2021}, instead, allows to understand the role of the input modality. BEiT uses the recent self-supervised \emph{masked image modeling} pre-training, which closely resembles the masked language modeling one used in NLP. The proposed setup allows to run experiments by changing one factor at a time among the three we studied and to keep fixed the other two. In this way, we are able to properly compare results between the NLP and CV environments. 

\begin{table}
    \caption{Accuracy on the entire dataset of \texttt{sentiment analysis} (\texttt{ER}) and \texttt{QNLI} with BERT model. Continual pre-training has been performed sequentially over each experience of \texttt{scientific abstracts}. Base refers to the model pre-trained on Wikipedia.}
    \label{tab:cl-qnli-tweets-bert}
    \small
\centering
\begin{tabular}{lccccc|ccccc}
\toprule
\textbf{BERT} & \multicolumn{5}{c}{\textbf{Accuracy}} & \multicolumn{5}{c}{\textbf{1-epoch Accuracy}}  \\ \toprule
\textbf{Base \texttt{ER}} & \multicolumn{5}{c}{93.05} &  \multicolumn{5}{c}{92.70} \\ \midrule
\textbf{Base \texttt{QNLI}} & \multicolumn{5}{c}{90.43} &  \multicolumn{5}{c}{90.43} \\ \midrule
Exp.      & e1 & e2 & e3 & e4 & e5         & e1 & e2 & e3 & e4 & e5        \\ \midrule
\textbf{Pr. \texttt{ER}}            & 92.95   &  92.90  &  92.90  & 92.65 & 92.45            & 92.25   & 92.35  & 91.90 & 92.15 &   91.90                   \\ \midrule
\textbf{Pr. \texttt{QNLI}} &  90.28  & 89.75   &  90.50   & 89.93  &   90.01        & 90.01   & 89.49 & 89.31   & 89.11   &  89.29 \\ \bottomrule
\end{tabular}
\end{table}
\begin{figure}
    \centering
  \includegraphics[width=\textwidth]{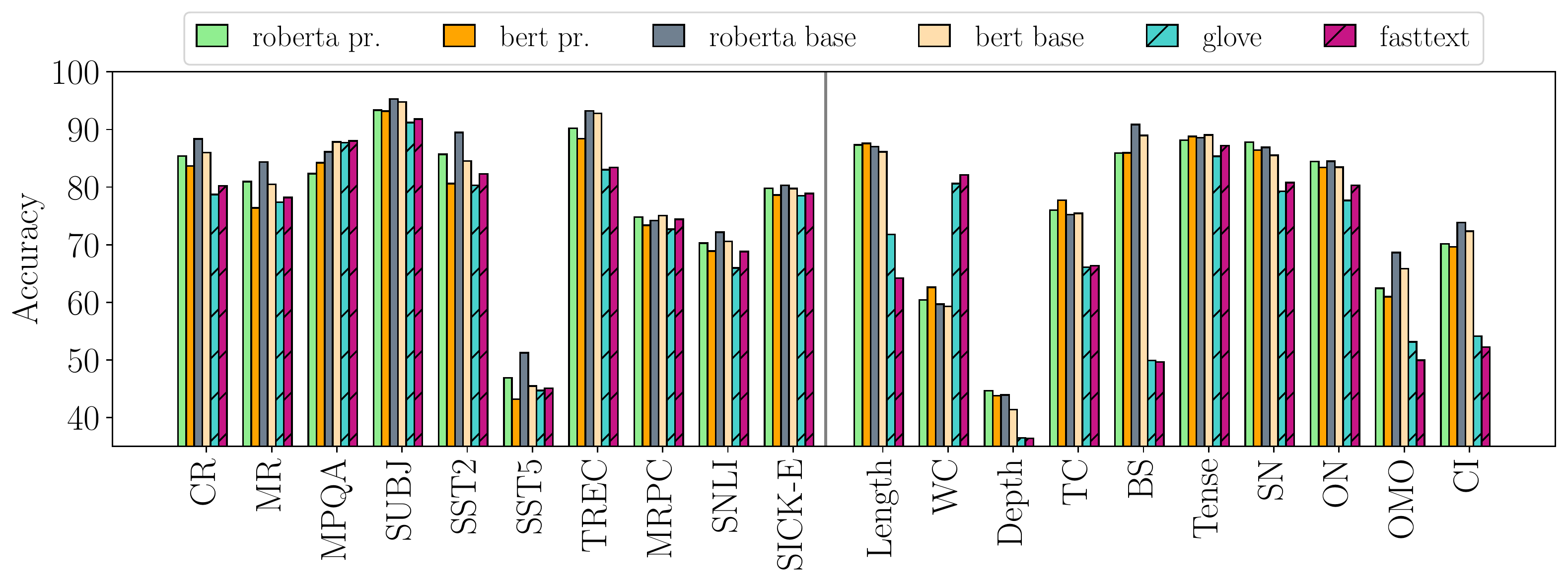}
  \caption{Accuracy on the $10$ transfer tasks (left) and $10$ probing tasks (right) of \texttt{SentEval}. Transformers are fine-tuned after $5$ experiences of pre-training on the \texttt{scientific abstracts}. Base refers to the model pre-trained on Wikipedia. Full results in Appendix \ref{app:senteval},}
  \label{fig:senteval}
\end{figure}

\subsection{Experimental Setup} \label{sec:setup}
For NLP, we use the Huggingface's pre-trained BERT and RoBERTa with $12$ layers. The NLP datasets, \texttt{SentEval} excluded, are also taken from Huggingface. For \texttt{SentEval}, we train our models using the original code. We use the same pre-training protocol across all experiments, with a learning rate of $5$e-$5$ and $30$ epochs with early stopping with $2$ epochs patience. For fine-tuning, we adopt a similar setup but with a learning rate of $1$e-$5$ and $20$ epochs.
For CV, we use ResNet101 and iNaturalist from Torchvision, while we retrieve ViT and BEiT models from Huggingface, using the version with $12$ layers in order to properly compare results with NLP experiments. We use Avalanche \citep{lomonaco2021} to run the continual pre-training and fine-tuning. For fine-tuning on FC task, we try few combinations of learning rates ($1e-5$, $1e-4$, $1e-3$) and batch sizes ($64$, $128$, $256$) on a held-out validation set built from \texttt{CORe50}. We report the best performance in terms of accuracy on the test set. The experimental setup is described in detail in Appendix \ref{app:setup}.

\section{Results} \label{sec:results}
We provide strong empirical evidence supporting the hypothesis that \emph{the continual pre-training scenario is less impacted by catastrophic forgetting than the traditional one}. In particular, we found the \emph{unsupervised pre-training objective to be the common factor for the resistance to forgetting} in the proposed environments. Our result adds to the evidences discussed in Section \ref{sec:related} for the robustness of unsupervised and self-supervised protocols with respect to catastrophic forgetting. Our evaluation offers similar conclusions for the novel continual pre-training scenario.
\begin{table}
    \caption{Fine-tuning accuracy on the entire dataset of \texttt{CORe50}. Pre-training has been performed sequentially over each experience of \texttt{iNaturalist}.}
    \label{tab:cl-inaturalist2core}
    \centering
    \small
    \begin{tabular}{lccccc|ccccc}
\toprule
\textbf{Model} & \multicolumn{5}{c}{\textbf{Accuracy}} & \multicolumn{5}{c}{\textbf{1-epoch Accuracy}}  \\ \toprule
\textbf{ResNet} & \multicolumn{5}{c}{94.72} &  \multicolumn{5}{c}{94.28} \\ \midrule
\textbf{ViT Base} & \multicolumn{5}{c}{90.56} &  \multicolumn{5}{c}{90.56} \\ \midrule
\textbf{BEiT Base} & \multicolumn{5}{c}{90.15} &  \multicolumn{5}{c}{82.51} \\ \midrule
Exp.      & e1 & e2 & e3 & e4 & e5         & e1 & e2 & e3 & e4 & e5        \\ \midrule
\textbf{ResNet Pr.}            & 89.88 & 81.29 & 80.82 & 77.78 & 74.35            & 88.40 & 69.93 & 70.43 & 65.91 & 57.60                 \\ \midrule
\textbf{ViT Pr.}            & 90.29 & 81.36 & 81.47 & 79.71 & 77.42           & 88.48  & 79.33   & 78.60 & 75.01 & 75.72                 \\ \midrule
\textbf{BEiT Pr.}            & 88.37 & 86.45 & 86.73 & 87.07 & 86.46               & 80.55 & 78.06 & 78.88 & 77.27 & 77.06                \\ \bottomrule
\end{tabular}
\end{table}
\paragraph{Continual pre-training improves performance on the downstream task without forgetting on FC datasets.} \label{results-nlp}
We verified that continual pre-training positively impacts on the performance on the downstream \texttt{scientific abstracts} classification task. That is, we observed that acquiring domain knowledge on \texttt{scientific abstracts} helps when solving the classification task (on held-out data). Appendix \ref{app:downstream} shows that continual pre-training on $5$ experiences improves the downstream classification performance (Table \ref{tab:cl-entire-abstract}). Performance is improved also with one step of pre-training on the entire \texttt{scientific abstracts} dataset. As discussed in Appendix \ref{app:downstream}, while the improvement is relatively small, we were able to achieve it by using a smaller number of samples with respect to the common pre-training datasets (e.g. Wikipedia): this points to the fact that continual pre-training does not necessarily need enormous datasets to actually be beneficial (a very useful aspect for continual learning). \\
In terms of catastrophic forgetting on the FC dataset after continual pre-training, we show that, quite surprisingly, both RoBERTa (Table \ref{tab:cl-tweets} and \ref{tab:cl-qnli}) and BERT (Table \ref{tab:cl-qnli-tweets-bert}) achieves almost zero forgetting, reaching an accuracy comparable to the one originally obtained by the model before continual pre-training. This happens both for \texttt{sentiment analysis} and \texttt{QNLI}. Moreover, a single epoch of gradient descent is sufficient to retain most of the original performance, showing the quick adaptation capabilities of the pre-trained models. Notably, the additional pretraining steps on domain-specific texts along with the expansion of the RoBERTa vocabulary does not worsen the effects of catastrophic forgetting. We conducted a broader empirical assessment on a diverse set of NLP tasks by using the \texttt{SentEval} benchmark. Figure \ref{fig:senteval} shows the downstream performance of BERT and RoBERTa after the entire continual pre-training stream. GloVe and fastText results are used as baselines and are taken from \cite{CONNEAU18.757}, except on \texttt{SNLI} and on all probing tasks, for which they were not available. Therefore, we computed these results using original code. The results confirm our findings: BERT and RoBERTa do not not show clear signs of forgetting, neither with respect to their original pre-trained version, nor with respect to the baselines.
\paragraph{Self-supervised continual pre-training mitigates forgetting.} \label{results-self}
We found out that self-supervised continual pre-training is the main responsible for the mitigation of forgetting in continual pre-training.\\
Since all NLP models use the self-supervised masked language modeling task for pre-training, we turned our attention to the CV environment. In fact, ResNet and ViT both use a supervised image classification during pre-training. In contrast, BEiT uses the recent self-supervised protocol of masked image modeling \citep{bao2021} (mirroring the NLP setting).
We show (Table \ref{tab:cl-inaturalist2core}) that BEiT shares the same properties of the NLP transformers, showing little forgetting with respect to the original version on the FC dataset (and one epoch of fine-tuning is sufficient to recover the original performance). Interestingly, ResNet and ViT exhibit a qualitatively different trend, with a substantial accuracy drop of around $20\%$ and $13\%$, respectively. This difference in performance hints towards the fact that \emph{supervised pre-training} in both ResNet and ViT is the main responsible of forgetting.

\begin{figure}
     \centering
     \begin{subfigure}[b]{0.19\textwidth}
         \centering
         \includegraphics[width=\textwidth]{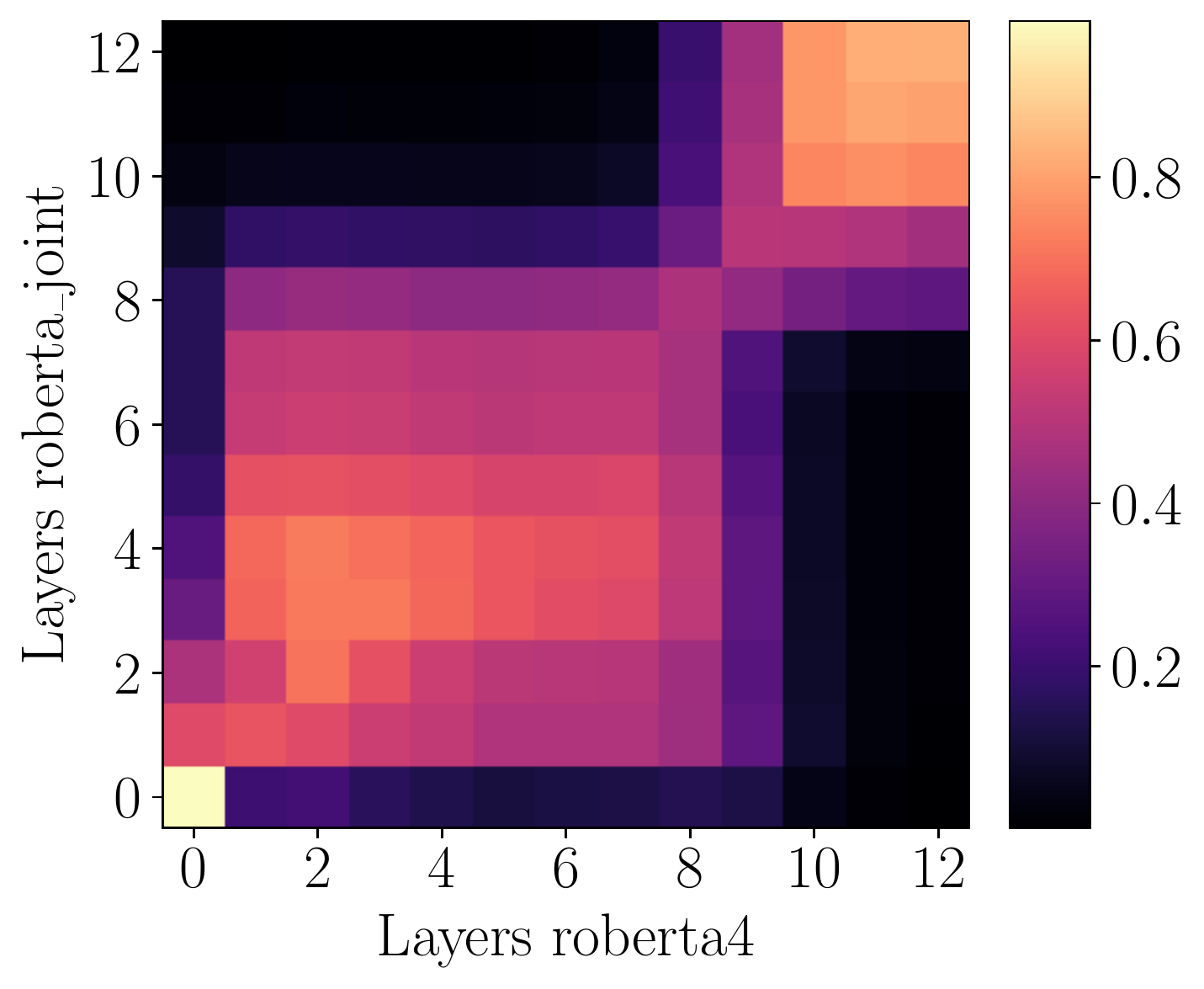}
         \caption{RoBERTa QNLI}
     \end{subfigure}
     \hfill
     \begin{subfigure}[b]{0.19\textwidth}
         \centering
         \includegraphics[width=\textwidth]{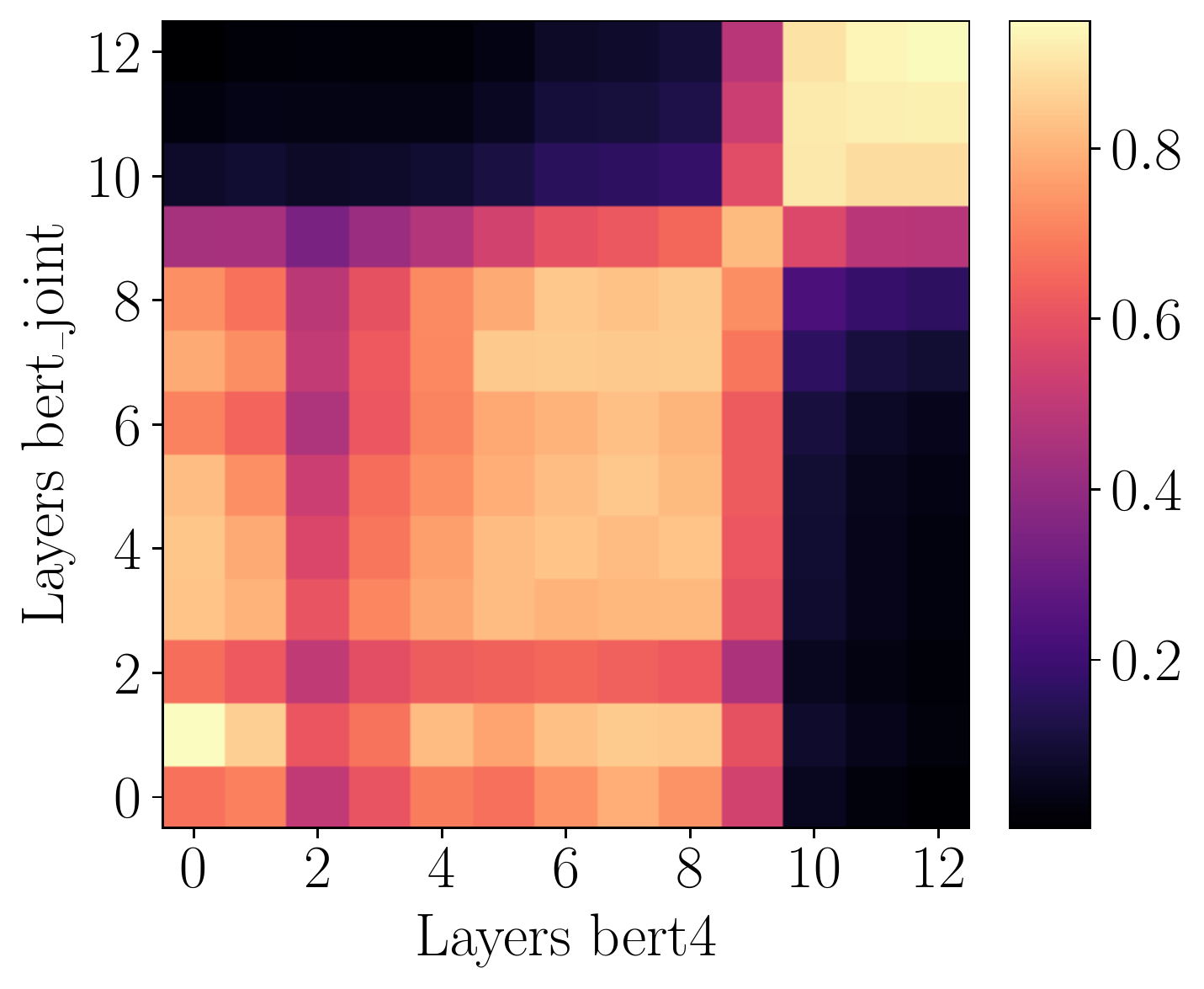}
         \caption{BERT Tweets}
     \end{subfigure}
     \hfill
     \begin{subfigure}[b]{0.19\textwidth}
         \centering
         \includegraphics[width=\textwidth]{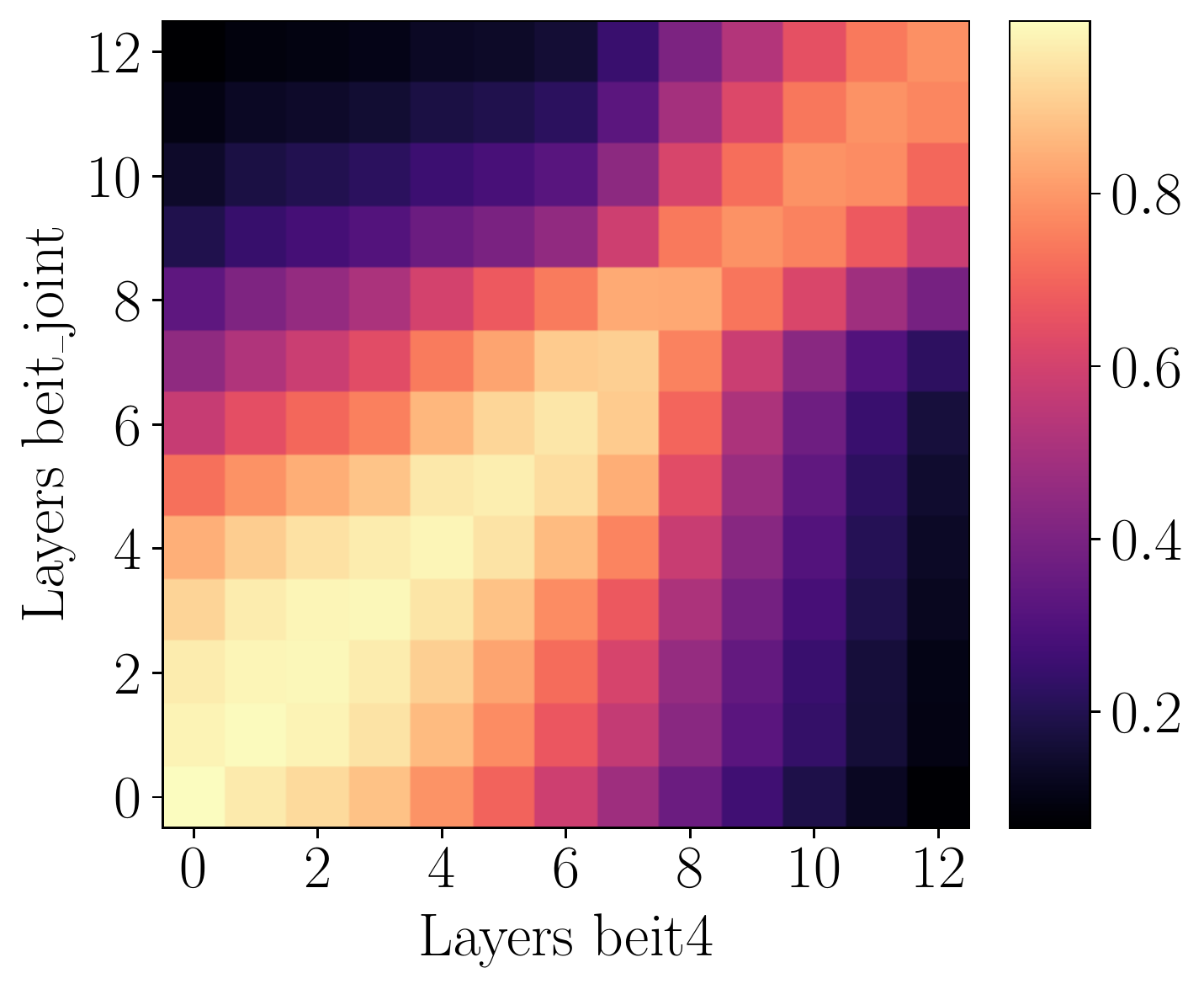}
         \caption{BEiT}
     \end{subfigure}
     \hfill
     \begin{subfigure}[b]{0.19\textwidth}
         \centering
         \includegraphics[width=\textwidth]{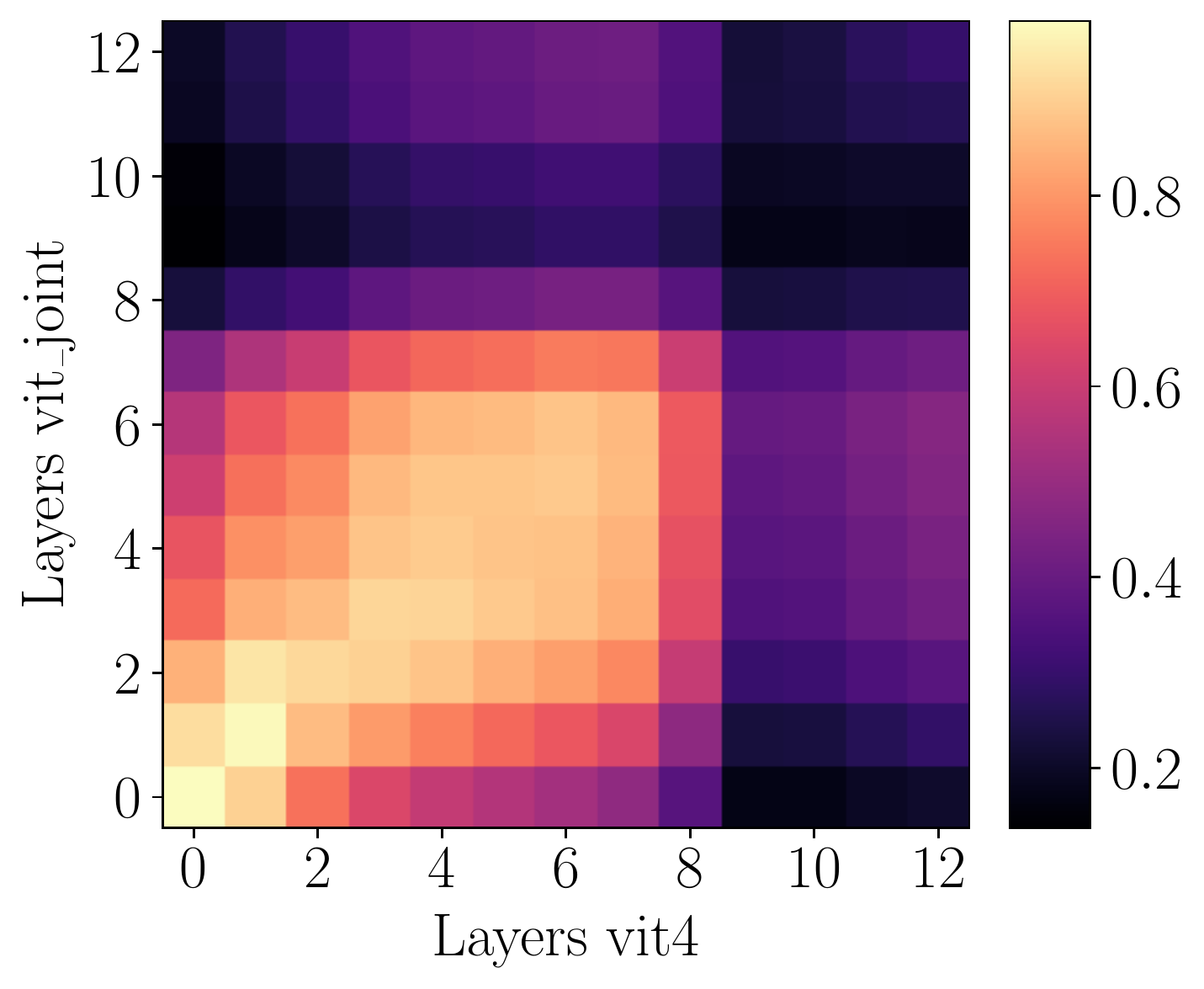}
         \caption{ViT}
     \end{subfigure}
     \begin{subfigure}[b]{0.19\textwidth}
         \centering
         \includegraphics[width=\textwidth]{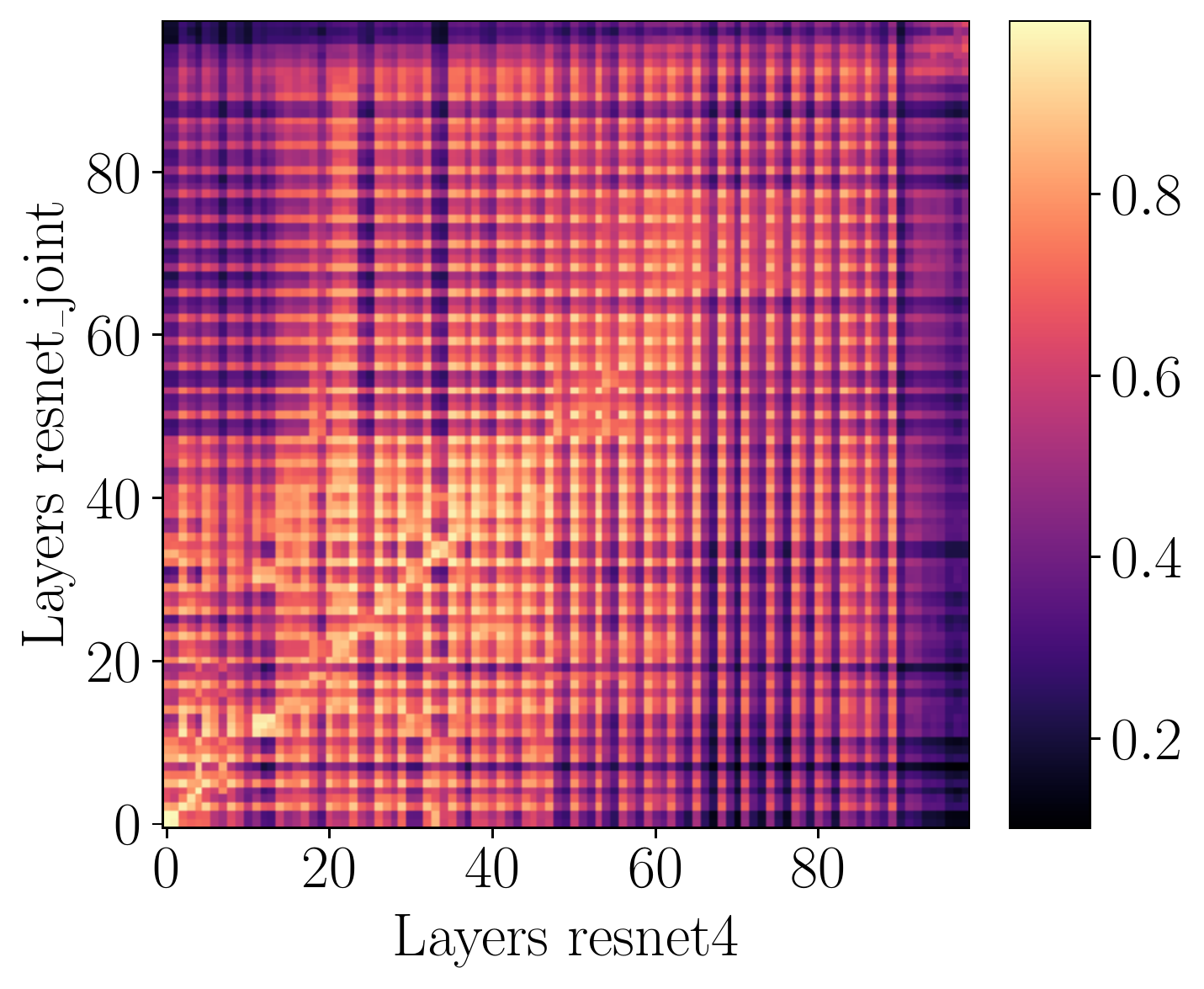}
         \caption{ResNet}
     \end{subfigure}
        \caption{CKA for RoBERTa, BERT, BEiT, Vit and ResNet. Pre-trained model $h^{ds}_5$ after the last experience (x axis) is compared with the original pre-trained model $h^{ds}_0$ (y axis). Each row is the similarity of a layer with respect to each layer of the other model.}
        \label{fig:cka}
\end{figure}

\paragraph{The negligible role of the architecture.} \label{results-architecture}
The type of Transformer used in the experiments does not appear to be a fundamental component: we experimented with larger vision models with $24$ layers instead of $12$ (Appendix \ref{app:larger}) without being able to appreciate any significant difference. The difference between convolutional networks like ResNet and attention-based transformers does not seem to have an impact, either. While ResNet sometimes exhibits worse performance than ViT, there is no clear evidence that this kind of model is more susceptible to forgetting.
\begin{table}
    \caption{Linear evaluation accuracy on the \texttt{sentiment analysis} (\texttt{ER}) and \texttt{QNLI} datasets. Pre-training has been performed sequentially over each experience of \texttt{scientific abstracts}.}
    \label{tab:nlp-linear}
    \centering
    \small
    \begin{tabular}{lccccc|ccccc}
\toprule
\textbf{Model} & \multicolumn{5}{c}{\textbf{\texttt{ER}}} & \multicolumn{5}{c}{\textbf{\texttt{QNLI}}}  \\ \toprule
\textbf{RoBERTa Base} & \multicolumn{5}{c}{60.05} &  \multicolumn{5}{c}{69.43} \\ \midrule
\textbf{BERT Base} & \multicolumn{5}{c}{59.85} &  \multicolumn{5}{c}{77.87} \\ \midrule
Exp.      & e1 & e2 & e3 & e4 & e5         & e1 & e2 & e3 & e4 & e5        \\ \midrule
\textbf{RoBERTa Pr.}            & 59.15 & 59.85 & 57.00 & 54.10 & 58.05            & 68.88 & 68.97 & 67.16 & 68.08 & 67.55                \\ \midrule
\textbf{BERT Pr.}            & 60.15 & 59.15 & 59.35 & 58.20 & 56.70            & 75.62  & 74.15 & 72.93  & 73.37 & 73.44                \\ \bottomrule
\end{tabular}
\end{table}
\paragraph{Feature Space Analysis: supervised pre-training induces larger drifts.} \label{results-features}
\begin{wraptable}{r}{0.45\linewidth}
    \caption{Linear evaluation accuracy on the entire dataset of \texttt{CORe50}. Pre-training has been performed sequentially over each experience of \texttt{iNaturalist}.}
    \label{tab:cl-inaturalist2core-linear}
    \setlength{\tabcolsep}{3pt}
    \small
    \begin{tabular}{lccccc}
\toprule
\textbf{Model} & \multicolumn{5}{c}{\textbf{Accuracy}}  \\ \toprule
\textbf{ResNet} & \multicolumn{5}{c}{82.50} \\ \midrule
\textbf{ViT Base} & \multicolumn{5}{c}{91.90} \\ \midrule
\textbf{BEiT Base} & \multicolumn{5}{c}{52.75} \\ \midrule
Exp.      & e1 & e2 & e3 & e4 & e5         \\ \midrule
\textbf{ResNet Pr.}            & 61.99 & 31.02 & 34.71 & 26.41 & 22.01                 \\ \midrule
\textbf{ViT Pr.}            & 79.38 & 55.20 & 57.98 & 60.49 & 48.25                    \\ \midrule
\textbf{BEiT Pr.}            & 52.34 & 51.71 & 51.31 & 53.12 & 52.51                \\ \bottomrule
\end{tabular}
\end{wraptable}
We verified the coherence of our findings by studying the feature space of the models. We leveraged linear evaluation for a quantitative analysis and Centered Kernel Alignment (CKA) \citep{kornblith2019} for a qualitatively analysis. Linear evaluation (i.e., training only the linear classifier and keeping the rest of the model fixed) is a powerful tool to understand the impact of the learned model representations in terms of catastrophic forgetting \citep{davari2022}. A model which exhibits forgetting during linear evaluation is likely to posses features which are not representative of the task. Conversely, a good linear evaluation performance points to a set of strong features, since it means that the task is linearly separable in that feature space. We adopted this approach for our continual pre-training scenario. In the NLP environment (Table \ref{tab:nlp-linear}), the features built by the models during continual pre-training are robust and do not cause a large deviation of performance with respect to the original pre-trained model. The lower training accuracy with respect to fine-tuning is expected, since we train only a subset of all parameters. In the CV environment (Table \ref{tab:cl-inaturalist2core-linear}), both ResNet and Vit suffer from forgetting, while BEiT does not (although it reaches a lower absolute accuracy).\\
Following \cite{hu2021a}, we used CKA with linear kernel \cite{kornblith2019} to compute layers similarity between the original pre-trained model and its continually pre-trained versions. From Figure \ref{fig:cka}, we can see that all models show large correlations across bottom layers (features are not drifting much). Instead, ViT and ResNet show lower correlation values for the final layers than BEiT. This corresponds to a larger drift (full set of results in Appendix \ref{app:cka}) in those layers. These results are compatible with what showed by \cite{madaan2021} for unsupervised CL, namely that unsupervised models in the traditional CL scenario have larger correlations in the lower layers than supervised ones. Our results further extend this conclusion to continual pre-training, supporting the idea that pre-training acts mainly in the upper layer of the networks (the ones containing more specific domain knowledge) and that heavy changes in these layers are enough to cause a deterioration of performance on the FC dataset, resulting in forgetting.

\section{Discussion and Limitations} \label{sec:discussion}
\begin{wraptable}{r}{0.45\linewidth}
    \caption{Main takeaways from the experiments presented in this paper. Only the supervised pre-training protocols showed clear signs of forgetting, while unsupervised/self-supervised protocols did not.}
    \label{tab:summary}
    \centering
    \setlength{\tabcolsep}{2pt}
    \small
    \begin{tabular}{cccc}
\toprule
\textbf{Pre-training} & \textbf{Architecture} & \textbf{Data} & \textbf{Forgetting} \\ \midrule
Unsupervised & Transformer & Words & $\times$\\ \midrule
Unsupervised & Transformer & Images & $\times$ \\ \midrule
Supervised & Transformer & Images & $\checkmark$ \\ \midrule
Supervised & CNN & Images & $\checkmark$  \\
\bottomrule
\end{tabular}
\end{wraptable}
Our empirical evaluation provides evidence that forgetting is mitigated in continual pre-training by the usage of self-supervised pre-training protocols (Table \ref{tab:summary}). 
Fine-tuning for only one epoch allows to recover most of the performance: this is important since an expensive fine-tuning phase might reduce the applicability of continual pre-training in environments with constrained resources.
Deciding when to use continual pre-training and when to use the traditional CL scenario is an open question. As previously discussed, the properties of continual pre-training do not fit the case in which a single model has to be readily applicable to different tasks without a step of fine-tuning. Nonetheless, we believe that whenever knowledge must be kept updated over time, continual pre-training can deliver a superior solution, less affected by forgetting (see Appendix \ref{app:cl} for a comparison with the traditional CL scenario). Continual pre-training offers the possibility to shift the focus from the mitigation of forgetting to other CL objectives like quick adaptation and knowledge reuse and transfer.\\
The main limitation of our study is related to the scale of the experiments, as we were able to experiment with only a limited number of datasets for each environment. While the computational cost of each experiment was reasonable (each experiment took from few hours - fine-tuning - to around one day - continual pre-training on a single A100), the number of experiments per environment was large. We preferred to thoroughly evaluate few environments rather than trying to address a wide range of different datasets without being able to properly explore them (Table \ref{tab:combinations}).  
We are well aware that a comprehensive exploration of continual pre-training in both NLP and CV domains is an ambitious objective, possible only in the context of a broad research program. However, we are confident of the fact that this study has shed some light on the subject and clearly pointed towards promising research directions.

\section{Conclusion}
Continual pre-training represents a novel CL scenario with promising opportunities and unexpected characteristics. In this work, we formally defined the continual pre-training scenario and we showed the effect that pre-training has on catastrophic forgetting, for both NLP and CV environments and with different architectures. Our results show that forgetting can be effectively mitigated by means of self-supervised pretraining, even with a single epoch of fine-tuning on the FC dataset. Ultimately, this work opens up the possibility to continually train large pre-trained models in a scalable and efficient way. Much like Deep Learning has advanced by disentangling the representation learning objective from the solution to specific tasks, continual pre-training aims to focus on the incremental development of robust features which are kept updated over time. This is a fundamental property towards the achievement of agents who can truly learn continuously in the real-world.


\begin{ack}
This work has been partially supported by the H2020 TEACHING project (GA 871385).
\end{ack}

\medskip

\bibliography{mybib, CL}

\begin{thebibliography}{43}
\providecommand{\natexlab}[1]{#1}
\providecommand{\EM}{\em}
\providecommand{\RNtxt}{\relax}
\RNtxt{}

\bibitem[Bao et~al.(2021)H.~Bao, L.~Dong, S.~Piao, F.~Wei]{bao2021}
{\EM Bao Hangbo, Dong Li, Piao Songhao, Wei Furu}.
\newblock {{BEiT}}: {{BERT Pre-Training}} of {{Image Transformers}}
  \allowbreak\newblock// International {{Conference}} on {{Learning
  Representations}}. 2021.

\bibitem[Chen et~al.(2020)X.~Chen, H.~Fan, R.~Girshick, K.~He]{chen2020b}
{\EM Chen Xinlei, Fan Haoqi, Girshick Ross, He~Kaiming}.
\newblock Improved {{Baselines}} with {{Momentum Contrastive Learning}}
  \allowbreak\newblock// arXiv:2003.04297 [cs]. 2020.

\bibitem[Conneau, Kiela(May 7-12, 2018 2018)A.~Conneau,
  D.~Kiela]{CONNEAU18.757}
{\EM Conneau Alexis, Kiela Douwe}.
\newblock {{SentEval}}: {{An}} Evaluation Toolkit for Universal Sentence
  Representations \allowbreak\newblock// Proceedings of the Eleventh
  International Conference on Language Resources and Evaluation ({{LREC}}
  2018). {Miyazaki, Japan}: {European Language Resources Association (ELRA)},
  May 7-12, 2018 2018.

\bibitem[Davari et~al.(2022)M.~Davari, N.~Asadi, S.~Mudur, R.~Aljundi,
  E.~Belilovsky]{davari2022}
{\EM Davari MohammadReza, Asadi Nader, Mudur Sudhir, Aljundi Rahaf, Belilovsky
  Eugene}.
\newblock Probing {{Representation Forgetting}} in {{Supervised}} and
  {{Unsupervised Continual Learning}} \allowbreak\newblock// Proceedings of the
  {{IEEE}}/{{CVF Conference}} on {{Computer Vision}} and {{Pattern
  Recognition}}. 2022.

\bibitem[De~Lange et~al.(2021)M.~De~Lange, R.~Aljundi, M.~Masana, S.~Parisot,
  X.~Jia, A.~Leonardis, G.~Slabaugh, T.~Tuytelaars]{delange2021}
{\EM De~Lange Matthias, Aljundi Rahaf, Masana Marc, Parisot Sarah, Jia Xu,
  Leonardis Ales, Slabaugh Gregory, Tuytelaars Tinne}.
\newblock A Continual Learning Survey: {{Defying}} Forgetting in Classification
  Tasks \allowbreak\newblock// IEEE Transactions on Pattern Analysis and
  Machine Intelligence. 2021.

\bibitem[Devlin et~al.(2019)J.~Devlin, M.-W. Chang, K.~Lee,
  K.~Toutanova]{devlin2019}
{\EM Devlin Jacob, Chang Ming-Wei, Lee Kenton, Toutanova Kristina}.
\newblock {{BERT}}: {{Pre-training}} of {{Deep Bidirectional Transformers}} for
  {{Language Understanding}} \allowbreak\newblock// Proceedings of the 2019
  {{Conference}} of the {{North American Chapter}} of the {{Association}} for
  {{Computational Linguistics}}: {{Human Language Technologies}}, {{Volume}} 1
  ({{Long}} and {{Short Papers}}). {Minneapolis, Minnesota}: {Association for
  Computational Linguistics}, 2019.  4171--4186.

\bibitem[Dosovitskiy et~al.(2020)A.~Dosovitskiy, L.~Beyer, A.~Kolesnikov,
  D.~Weissenborn, X.~Zhai, T.~Unterthiner, M.~Dehghani, M.~Minderer,
  G.~Heigold, S.~Gelly, J.~Uszkoreit, N.~Houlsby]{dosovitskiy2020}
{\EM Dosovitskiy Alexey, Beyer Lucas, Kolesnikov Alexander, Weissenborn Dirk,
  Zhai Xiaohua, Unterthiner Thomas, Dehghani Mostafa, Minderer Matthias,
  Heigold Georg, Gelly Sylvain, Uszkoreit Jakob, Houlsby Neil}.
\newblock An {{Image}} Is {{Worth}} 16x16 {{Words}}: {{Transformers}} for
  {{Image Recognition}} at {{Scale}} \allowbreak\newblock// International
  {{Conference}} on {{Learning Representations}}. 2020.

\bibitem[Douillard et~al.(2022)A.~Douillard, A.~Ram{\'e}, G.~Couairon,
  M.~Cord]{douillard2021}
{\EM Douillard Arthur, Ram{\'e} Alexandre, Couairon Guillaume, Cord Matthieu}.
\newblock {{DyTox}}: {{Transformers}} for {{Continual Learning}} with {{DYnamic
  TOken eXpansion}} \allowbreak\newblock// {{IEEE}}/{{CVF Conference}} on
  {{Computer Vision}} and {{Pattern Recognition}}. 2022.

\bibitem[French(1999)R.~French]{french1999}
{\EM French Robert}.
\newblock Catastrophic Forgetting in Connectionist Networks
  \allowbreak\newblock// Trends in Cognitive Sciences. 1999. 3, 4. 128--135.

\bibitem[Geiger(2019)R.~S. Geiger]{geiger2019}
{\EM Geiger R.~Stuart}.
\newblock {{ArXiV Archive}}: {{A}} Tidy and Complete Archive of Metadata for
  Papers on Arxiv.Org, 1993-2019. 2019.

\bibitem[Gu et~al.(2021)Y.~Gu, R.~Tinn, H.~Cheng, M.~Lucas, N.~Usuyama, X.~Liu,
  T.~Naumann, J.~Gao, H.~Poon]{gu2021}
{\EM Gu~Yu, Tinn Robert, Cheng Hao, Lucas Michael, Usuyama Naoto, Liu Xiaodong,
  Naumann Tristan, Gao Jianfeng, Poon Hoifung}.
\newblock Domain-{{Specific Language Model Pretraining}} for {{Biomedical
  Natural Language Processing}} \allowbreak\newblock// ACM Transactions on
  Computing for Healthcare. 2021. 3, 1. 2:1--2:23.

\bibitem[Gururangan et~al.(2020)S.~Gururangan, A.~Marasovi{\'c},
  S.~Swayamdipta, K.~Lo, I.~Beltagy, D.~Downey, N.~A. Smith]{gururangan2020}
{\EM Gururangan Suchin, Marasovi{\'c} Ana, Swayamdipta Swabha, Lo~Kyle, Beltagy
  Iz, Downey Doug, Smith Noah~A.}
\newblock Don't {{Stop Pretraining}}: {{Adapt Language Models}} to {{Domains}}
  and {{Tasks}} \allowbreak\newblock// Proceedings of the 58th {{Annual
  Meeting}} of the {{Association}} for {{Computational Linguistics}}. {Online}:
  {Association for Computational Linguistics}, 2020.  8342--8360.

\bibitem[Hadsell et~al.(2020)R.~Hadsell, D.~Rao, A.~A. Rusu,
  R.~Pascanu]{hadsell2020}
{\EM Hadsell Raia, Rao Dushyant, Rusu Andrei~A, Pascanu Razvan}.
\newblock Embracing {{Change}}: {{Continual Learning}} in {{Deep Neural
  Networks}} \allowbreak\newblock// Trends in Cognitive Sciences. 2020.

\bibitem[Han et~al.(2021)R.~Han, X.~Ren, N.~Peng]{han2021}
{\EM Han Rujun, Ren Xiang, Peng Nanyun}.
\newblock {{ECONET}}: {{Effective Continual Pretraining}} of {{Language
  Models}} for {{Event Temporal Reasoning}} \allowbreak\newblock// Proceedings
  of the 2021 {{Conference}} on {{Empirical Methods}} in {{Natural Language
  Processing}}. {Online and Punta Cana, Dominican Republic}: {Association for
  Computational Linguistics}, 2021.  5367--5380.

\bibitem[He et~al.(2016)K.~He, X.~Zhang, S.~Ren, J.~Sun]{he2016}
{\EM He~Kaiming, Zhang Xiangyu, Ren Shaoqing, Sun Jian}.
\newblock Deep {{Residual Learning}} for {{Image Recognition}}
  \allowbreak\newblock// 2016 {{IEEE Conference}} on {{Computer Vision}} and
  {{Pattern Recognition}} ({{CVPR}}). 2016.  770--778.

\bibitem[Hu et~al.(2021)D.~Hu, S.~Yan, Q.~Lu, L.~Hong, H.~Hu, Y.~Zhang, Z.~Li,
  X.~Wang, J.~Feng]{hu2021a}
{\EM Hu~Dapeng, Yan Shipeng, Lu~Qizhengqiu, Hong Lanqing, Hu~Hailin, Zhang
  Yifan, Li~Zhenguo, Wang Xinchao, Feng Jiashi}.
\newblock How {{Well Does Self-Supervised Pre-Training Perform}} with
  {{Streaming Data}}? \allowbreak\newblock// International {{Conference}} on
  {{Learning Representations}}. 2021.

\bibitem[Jang et~al.(2022)J.~Jang, S.~Ye, C.~Lee, S.~Yang, J.~Shin, J.~Han,
  G.~Kim, M.~Seo]{jang2022}
{\EM Jang Joel, Ye~Seonghyeon, Lee Changho, Yang Sohee, Shin Joongbo, Han
  Janghoon, Kim Gyeonghun, Seo Minjoon}.
\newblock {{TemporalWiki}}: {{A Lifelong Benchmark}} for {{Training}} and
  {{Evaluating Ever-Evolving Language Models}} \allowbreak\newblock//
  arXiv:2204.14211 [cs]. 2022.

\bibitem[Jang et~al.(2021)J.~Jang, S.~Ye, S.~Yang, J.~Shin, J.~Han, G.~Kim,
  S.~J. Choi, M.~Seo]{jang2021}
{\EM Jang Joel, Ye~Seonghyeon, Yang Sohee, Shin Joongbo, Han Janghoon, Kim
  Gyeonghun, Choi Stanley~Jungkyu, Seo Minjoon}.
\newblock Towards {{Continual Knowledge Learning}} of {{Language Models}}
  \allowbreak\newblock// International {{Conference}} on {{Learning
  Representations}}. 2021.

\bibitem[Jin et~al.(2021)X.~Jin, D.~Zhang, H.~Zhu, W.~Xiao, S.-W. Li, X.~Wei,
  A.~Arnold, X.~Ren]{jin2021a}
{\EM Jin Xisen, Zhang Dejiao, Zhu Henghui, Xiao Wei, Li~Shang-Wen, Wei Xiaokai,
  Arnold Andrew, Ren Xiang}.
\newblock Lifelong {{Pretraining}}: {{Continually Adapting Language Models}} to
  {{Emerging Corpora}} \allowbreak\newblock// arXiv:2110.08534 [cs]. 2021.

\bibitem[Kornblith et~al.(2019)S.~Kornblith, M.~Norouzi, H.~Lee,
  G.~Hinton]{kornblith2019}
{\EM Kornblith Simon, Norouzi Mohammad, Lee Honglak, Hinton Geoffrey}.
\newblock Similarity of {{Neural Network Representations Revisited}}
  \allowbreak\newblock// Proceedings of the 36th {{International Conference}}
  on {{Machine Learning}}. 2019.  3519--3529.

\bibitem[Lazaridou et~al.(2021)A.~Lazaridou, A.~Kuncoro, E.~Gribovskaya,
  D.~Agrawal, A.~Liska, T.~Terzi, M.~Gimenez, C.~d.~M. {d'Autume}, T.~Ko{\v
  c}isk{\'y}, S.~Ruder, D.~Yogatama, K.~Cao, S.~Young,
  P.~Blunsom]{lazaridou2021}
{\EM Lazaridou Angeliki, Kuncoro Adhiguna, Gribovskaya Elena, Agrawal Devang,
  Liska Adam, Terzi Tayfun, Gimenez Mai, {d'Autume} Cyprien de~Masson, Ko{\v
  c}isk{\'y} Tom{\'a}{\v s}, Ruder Sebastian, Yogatama Dani, Cao Kris, Young
  Susannah, Blunsom Phil}.
\newblock Mind the {{Gap}}: {{Assessing Temporal Generalization}} in {{Neural
  Language Models}} \allowbreak\newblock// Thirty-{{Fifth Conference}} on
  {{Neural Information Processing Systems}}. 2021.

\bibitem[Lee et~al.(2020)J.~Lee, W.~Yoon, S.~Kim, D.~Kim, S.~Kim, C.~H. So,
  J.~Kang]{lee2020}
{\EM Lee Jinhyuk, Yoon Wonjin, Kim Sungdong, Kim Donghyeon, Kim Sunkyu,
  So~Chan~Ho, Kang Jaewoo}.
\newblock {{BioBERT}}: A Pre-Trained Biomedical Language Representation Model
  for Biomedical Text Mining \allowbreak\newblock// Bioinformatics. 2020. 36,
  4. 1234--1240.

\bibitem[Lesort et~al.(2020)T.~Lesort, V.~Lomonaco, A.~Stoian, D.~Maltoni,
  D.~Filliat, N.~{D{\'i}az-Rodr{\'i}guez}]{lesort2020}
{\EM Lesort Timoth{\'e}e, Lomonaco Vincenzo, Stoian Andrei, Maltoni Davide,
  Filliat David, {D{\'i}az-Rodr{\'i}guez} Natalia}.
\newblock Continual Learning for Robotics: {{Definition}}, Framework, Learning
  Strategies, Opportunities and Challenges \allowbreak\newblock// Information
  Fusion. 2020. 58. 52--68.

\bibitem[Liu et~al.(2019)Y.~Liu, M.~Ott, N.~Goyal, J.~Du, M.~Joshi, D.~Chen,
  O.~Levy, M.~Lewis, L.~Zettlemoyer, V.~Stoyanov]{liu2019b}
{\EM Liu Yinhan, Ott Myle, Goyal Naman, Du~Jingfei, Joshi Mandar, Chen Danqi,
  Levy Omer, Lewis Mike, Zettlemoyer Luke, Stoyanov Veselin}.
\newblock {{RoBERTa}}: {{A Robustly Optimized BERT Pretraining Approach}}
  \allowbreak\newblock// arXiv:1907.11692 [cs]. 2019.

\bibitem[Lomonaco, Maltoni(2017)V.~Lomonaco, D.~Maltoni]{lomonaco2017}
{\EM Lomonaco Vincenzo, Maltoni Davide}.
\newblock {{CORe50}}: A {{New Dataset}} and {{Benchmark}} for {{Continuous
  Object Recognition}} \allowbreak\newblock// Proceedings of the 1st {{Annual
  Conference}} on {{Robot Learning}}.  78. 2017.  17--26.
\newblock (Proceedings of {{Machine Learning Research}}).

\bibitem[Lomonaco et~al.(2021)V.~Lomonaco, L.~Pellegrini, A.~Cossu, A.~Carta,
  G.~Graffieti, T.~L. Hayes, M.~De~Lange, M.~Masana, J.~Pomponi, G.~{van de
  Ven}, M.~Mundt, Q.~She, K.~Cooper, J.~Forest, E.~Belouadah, S.~Calderara,
  G.~I. Parisi, F.~Cuzzolin, A.~Tolias, S.~Scardapane, L.~Antiga, S.~Amhad,
  A.~Popescu, C.~Kanan, J.~{van de Weijer}, T.~Tuytelaars, D.~Bacciu,
  D.~Maltoni]{lomonaco2021}
{\EM Lomonaco Vincenzo, Pellegrini Lorenzo, Cossu Andrea, Carta Antonio,
  Graffieti Gabriele, Hayes Tyler~L., De~Lange Matthias, Masana Marc, Pomponi
  Jary, {van de Ven} Gido, Mundt Martin, She Qi, Cooper Keiland, Forest Jeremy,
  Belouadah Eden, Calderara Simone, Parisi German~I., Cuzzolin Fabio, Tolias
  Andreas, Scardapane Simone, Antiga Luca, Amhad Subutai, Popescu Adrian, Kanan
  Christopher, {van de Weijer} Joost, Tuytelaars Tinne, Bacciu Davide, Maltoni
  Davide}.
\newblock Avalanche: An {{End-to-End Library}} for {{Continual Learning}}
  \allowbreak\newblock// {{CLVision Workshop}} at {{CVPR}}. 2021.

\bibitem[{Lopez-Paz}, Ranzato(2017)D.~{Lopez-Paz}, M.~Ranzato]{lopez-paz2017}
{\EM {Lopez-Paz} David, Ranzato Marc'Aurelio}.
\newblock Gradient {{Episodic Memory}} for {{Continual Learning}}
  \allowbreak\newblock// {{NIPS}}. 2017.

\bibitem[Loureiro et~al.(2022)D.~Loureiro, F.~Barbieri, L.~Neves, L.~E. Anke,
  J.~{Camacho-Collados}]{loureiro2022}
{\EM Loureiro Daniel, Barbieri Francesco, Neves Leonardo, Anke Luis~Espinosa,
  {Camacho-Collados} Jose}.
\newblock {{TimeLMs}}: {{Diachronic Language Models}} from {{Twitter}}
  \allowbreak\newblock// arXiv:2202.03829 [cs]. 2022.

\bibitem[Madaan et~al.(2021)D.~Madaan, J.~Yoon, Y.~Li, Y.~Liu, S.~J.
  Hwang]{madaan2021}
{\EM Madaan Divyam, Yoon Jaehong, Li~Yuanchun, Liu Yunxin, Hwang Sung~Ju}.
\newblock Representational {{Continuity}} for {{Unsupervised Continual
  Learning}} \allowbreak\newblock// International {{Conference}} on {{Learning
  Representations}}. 2021.

\bibitem[McCloskey, Cohen(1989)M.~McCloskey, N.~J. Cohen]{mccloskey1989}
{\EM McCloskey Michael, Cohen Neal~J.}
\newblock Catastrophic {{Interference}} in {{Connectionist Networks}}: {{The
  Sequential Learning Problem}} \allowbreak\newblock// Psychology of
  {{Learning}} and {{Motivation}}.  24. 1989.  109--165.

\bibitem[Mehta et~al.(2020)N.~Mehta, K.~J. Liang, L.~Carin]{mehta2020}
{\EM Mehta Nikhil, Liang Kevin~J, Carin Lawrence}.
\newblock Bayesian {{Nonparametric Weight Factorization}} for {{Continual
  Learning}} \allowbreak\newblock// arXiv. 2020.  1--17.

\bibitem[Mehta et~al.(2021)S.~V. Mehta, D.~Patil, S.~Chandar,
  E.~Strubell]{mehta2021}
{\EM Mehta Sanket~Vaibhav, Patil Darshan, Chandar Sarath, Strubell Emma}.
\newblock An {{Empirical Investigation}} of the {{Role}} of {{Pre-training}} in
  {{Lifelong Learning}} \allowbreak\newblock// arXiv:2112.09153 [cs]. 2021.

\bibitem[Merity et~al.(2016)S.~Merity, C.~Xiong, J.~Bradbury,
  R.~Socher]{merity2016}
{\EM Merity Stephen, Xiong Caiming, Bradbury James, Socher Richard}.
\newblock Pointer {{Sentinel Mixture Models}} \allowbreak\newblock//
  arXiv:1609.07843 [cs]. 2016.

\bibitem[Nguyen et~al.(2020)T.~Nguyen, M.~Raghu, S.~Kornblith]{nguyen2020a}
{\EM Nguyen Thao, Raghu Maithra, Kornblith Simon}.
\newblock Do {{Wide}} and {{Deep Networks Learn}} the {{Same Things}}?
  {{Uncovering How Neural Network Representations Vary}} with {{Width}} and
  {{Depth}} \allowbreak\newblock// International {{Conference}} on {{Learning
  Representations}}. 2020.

\bibitem[Parisi et~al.(2019)G.~I. Parisi, R.~Kemker, J.~L. Part, C.~Kanan,
  S.~Wermter]{parisi2019}
{\EM Parisi German~I, Kemker Ronald, Part Jose~L, Kanan Christopher, Wermter
  Stefan}.
\newblock Continual Lifelong Learning with Neural Networks: {{A}} Review
  \allowbreak\newblock// Neural Networks. 2019. 113. 54--71.

\bibitem[Qin et~al.(2022)Y.~Qin, J.~Zhang, Y.~Lin, Z.~Liu, P.~Li, M.~Sun,
  J.~Zhou]{qin2022}
{\EM Qin Yujia, Zhang Jiajie, Lin Yankai, Liu Zhiyuan, Li~Peng, Sun Maosong,
  Zhou Jie}.
\newblock {{ELLE}}: {{Efficient Lifelong Pre-training}} for {{Emerging Data}}
  \allowbreak\newblock// Findings of ACL. 2022.

\bibitem[Ramasesh et~al.(2021)V.~V. Ramasesh, A.~Lewkowycz,
  E.~Dyer]{ramasesh2021}
{\EM Ramasesh Vinay~Venkatesh, Lewkowycz Aitor, Dyer Ethan}.
\newblock Effect of Scale on Catastrophic Forgetting in Neural Networks
  \allowbreak\newblock// International {{Conference}} on {{Learning
  Representations}}. 2021.

\bibitem[Rongali et~al.(2021)S.~Rongali, A.~Jagannatha, B.~P.~S. Rawat,
  H.~Yu]{rongali2021}
{\EM Rongali Subendhu, Jagannatha Abhyuday, Rawat Bhanu Pratap~Singh, Yu~Hong}.
\newblock Continual {{Domain-Tuning}} for {{Pretrained Language Models}}
  \allowbreak\newblock// arXiv:2004.02288 [cs]. 2021.

\bibitem[Ruder et~al.(2019)S.~Ruder, M.~E. Peters, S.~Swayamdipta,
  T.~Wolf]{ruder2019}
{\EM Ruder Sebastian, Peters Matthew~E., Swayamdipta Swabha, Wolf Thomas}.
\newblock Transfer {{Learning}} in {{Natural Language Processing}}
  \allowbreak\newblock// Proceedings of the 2019 {{Conference}} of the {{North
  American Chapter}} of the {{Association}} for {{Computational Linguistics}}:
  {{Tutorials}}. {Minneapolis, Minnesota}: {Association for Computational
  Linguistics}, 2019.  15--18.

\bibitem[Van~Horn et~al.(2018)G.~Van~Horn, O.~Mac~Aodha, Y.~Song, Y.~Cui,
  C.~Sun, A.~Shepard, H.~Adam, P.~Perona, S.~Belongie]{vanhorn2018}
{\EM Van~Horn Grant, Mac~Aodha Oisin, Song Yang, Cui Yin, Sun Chen, Shepard
  Alex, Adam Hartwig, Perona Pietro, Belongie Serge}.
\newblock The {{INaturalist Species Classification}} and {{Detection Dataset}}
  \allowbreak\newblock// Proceedings of the {{IEEE Conference}} on {{Computer
  Vision}} and {{Pattern Recognition}}. 2018.  8769--8778.

\bibitem[Vaswani et~al.(2017)A.~Vaswani, N.~Shazeer, N.~Parmar, J.~Uszkoreit,
  L.~Jones, A.~N. Gomez, {\L}.~Kaiser,
  I.~Polosukhin]{vaswaniAttentionAllYou2017}
{\EM Vaswani Ashish, Shazeer Noam, Parmar Niki, Uszkoreit Jakob, Jones Llion,
  Gomez Aidan~N, Kaiser {\L}ukasz, Polosukhin Illia}.
\newblock Attention Is {{All}} You {{Need}} \allowbreak\newblock// Advances in
  {{Neural Information Processing Systems}} 30. 2017.  5998--6008.

\bibitem[Wu et~al.(2021)T.~Wu, M.~Caccia, Z.~Li, Y.-F. Li, G.~Qi,
  G.~Haffari]{wu2021a}
{\EM Wu~Tongtong, Caccia Massimo, Li~Zhuang, Li~Yuan-Fang, Qi~Guilin, Haffari
  Gholamreza}.
\newblock Pretrained {{Language Model}} in {{Continual Learning}}: {{A
  Comparative Study}} \allowbreak\newblock// International {{Conference}} on
  {{Learning Representations}}. 2021.

\bibitem[Zhang et~al.(2020)R.~Zhang, R.~Gangi~Reddy, M.~A. Sultan, V.~Castelli,
  A.~Ferritto, R.~Florian, E.~Sarioglu~Kayi, S.~Roukos, A.~Sil,
  T.~Ward]{zhang2020}
{\EM Zhang Rong, Gangi~Reddy Revanth, Sultan Md~Arafat, Castelli Vittorio,
  Ferritto Anthony, Florian Radu, Sarioglu~Kayi Efsun, Roukos Salim, Sil Avi,
  Ward Todd}.
\newblock Multi-{{Stage Pre-training}} for {{Low-Resource Domain Adaptation}}
  \allowbreak\newblock// Proceedings of the 2020 {{Conference}} on {{Empirical
  Methods}} in {{Natural Language Processing}} ({{EMNLP}}). {Online}:
  {Association for Computational Linguistics}, 2020.  5461--5468.

\end{thebibliography}



\appendix

\section{Extended Experimental Setup} \label{app:setup}
Here, we describe the experimental setup we adopted in our work for both the NLP environment and the CV environment. All our experiments were run on a single A100 GPU with 80 GB of memory, on a server with 96 cores.
\paragraph{NLP}
The continual pre-training dataset of \texttt{scientific abstracts} is taken from GitHub\footnote{R. Stuart Geiger (2020), ArXiV Archive: A Tidy and Complete Archive of Metadata for Papers on arxiv.org, Zenodo: \url{http://doi.org/10.5281/zenodo.1463242}}. We selected $10$ ArXiv classes to build our continual pre-training stream, namely `hep-ph', `astro-ph', `hep-th', `quant-ph', `cond-mat.mes-hall', `gr-qc', `cond-mat.mtrl-sci', `cond-mat.str-el', `cond-mat.stat-mech' and `astro-ph.SR'. For both pre-training and downstream fine-tuning, we selected $10,000$ abstracts for each of the $10$ classes for the training set and $1,000$ for the test set. Hence, an abstract present in one of the training/test set of continual pre-training or downstream fine-tuning is not present in the other partitions. We chose similar abstract categories since being able to distinguish very different kinds of abstracts may greatly simplify the problem (e.g., one term may be enough to classify the entire abstract). We will publicly release our version of the scientific abstract dataset used in the experiments. The dataset can be easily loaded via Huggingface. \\
In order to select new tokens for the expansion of RoBERTa vocabulary at each experience of continual pre-training, we trained from scratch a tokenizer on the WikiText dataset \citep{merity2016}. This tokenizer quickly approximates the tokens present in Wikipedia. We also train a tokenizer on our \texttt{scientific abstracts} dataset and ranked the tokens which were occurring in the latter but not in the former tokenizer. That is, the domain tokens related to the \texttt{scientific abstracts} datasets. We selected $426$ new tokens for joint training experiments (Appendix \ref{app:downstream}) and $39/42/28/30/10$ for each of the $5$ experiences of continual pre-training.\\
We added tokens to the tokenizer such that new tokens have precedence over already existing tokens during tokenization process. Within new tokens, we sorted inversely by token length and the precedence is given by the order of addition (First In First Out). The list of new tokens is embedded in the released code. We also ran few experiments (not reported here) by adding with the same procedure sub-word tokens (BPE encoding) instead of word tokens. We did not find significant differences in the results, which do not seem to depend on which specific new tokens are selected, as long as they provide domain knowledge about the task.\\
The FC dataset \texttt{QNLI} is available from Huggingface as part of the GLUE benchmark \url{https://huggingface.co/datasets/glue}. The \texttt{sentiment analysis} from tweets dataset is also taken from Huggingface at \url{https://huggingface.co/datasets/emotion}.
\texttt{Senteval} benchmark is taken from the official codebase at \url{https://github.com/facebookresearch/SentEval}.

During linear evaluation, we removed the feedforward layer right before the classifier. We observed that keeping it frozen yielded a very low training performance. On the other side, fine-tuning it together with the linear classifier did not show the issue but resulted in a non linear fine-tuning procedure, making it difficult to compare results against the CV setup.
Therefore, linear evaluation is performed by taking the representation built for the special CLF token by the last hidden layer of the transformer and decoding it with a trained linear classifier.

\paragraph{Computer Vision}
We adopted the Masked Image Modeling task for self-supervised pre-training with BEiT. Following the original BEiT paper, we leveraged the DALL-E encoder, which is kept fixed during continual pre-training. A simple example of masked image modeling can be found at \url{https://github.com/NielsRogge/Transformers-Tutorials/blob/master/BEiT/Understanding_BeitForMaskedImageModeling.ipynb}. Experiments which continually pre-train also the encoder may constitute interesting future works.\\
Following the original Pytorch code at \url{https://github.com/pytorch/vision/blob/main/references/classification/presets.py}, for continual pre-training and fine-tuning on FC dataset with ResNet we used a chain of augmentations: RandomResizedCrop with bilinear interpolation, RandomHorizontalFlip and normalization of mean and standard deviation. On the test sets, we resized the image to $256x256$, applied center crop and normalization. ViT uses the same setup without normalization. BEiT applies the ViT setup on the FC dataset only.

For all CKA experiments, we used the Python library from \url{https://github.com/AntixK/PyTorch-Model-Compare}, which provides the unbiased minibatch estimator of the CKA.

\section{Additional Results}
\subsection{SentEval Results} \label{app:senteval}
Table \ref{tab:cl-senteval} shows the complete set of results for the \texttt{SentEval} benchmark. We compare the performance of continual pre-training after $5$ experiences on \texttt{scientific abstracts} against two baselines (GloVe and fastText) and the original pre-trained model. For RoBERTa, we also provide the results in case of vocabulary expansion. 
We used one hidden layer of $50$ units for probing tasks and logistic regression for the transfer tasks. 

\begin{table}
    \caption{Accuracy on $10$ transfer and $10$ probing tasks from \texttt{SentEval}. For comparison, we report the performance of the pre-trained models at the end of pre-training on the last experience (e5) of \texttt{scientific abstracts} dataset.}
    \label{tab:cl-senteval}
\centering
\tabcolsep=0.13cm
\begin{tabular}{lcc|ccc|cc}
\toprule
 & & & \multicolumn{3}{c}{\textbf{RoBERTa}} & \multicolumn{2}{c}{\textbf{BERT}} \\ \midrule 
\textbf{Task} & \textbf{GloVe} & \textbf{fastText} & \textbf{Base} & \textbf{Pretr.} & \textbf{Pretr. NT} & \textbf{Base} & \textbf{Pretr.}\\ \midrule
\textbf{CR} & 78.70 & 80.20 & 88.34 & 85.38 & 86.20 & 86.01 & 83.66 \\ \midrule
\textbf{MR} & 77.40 & 78.20 & 84.35 & 80.95 & 80.65 & 80.46 & 76.37 \\ \midrule
\textbf{MPQA} & 87.70 & 88.00 & 86.12 & 82.34 & 82.04 & 87.83 & 84.22 \\ \midrule
\textbf{SUBJ} & 91.20 & 91.80 & 95.28 & 93.34 & 93.36 & 94.79 & 93.19\\ \midrule
\textbf{SST2} & 80.30 & 82.30 & 89.46 & 85.67 & 85.17 & 84.51 & 80.62 \\ \midrule
\textbf{SST5} & 44.70 & 45.10 & 51.27 & 46.88 & 46.65 & 45.48 & 43.21\\ \midrule
\textbf{TREC} & 83.00 & 83.40 & 93.20 & 90.20 & 90.40 & 92.80 & 88.40\\ \midrule
\textbf{MRPC} & 72.70 & 74.40 & 74.20 & 74.78 & 74.67 & 75.07 & 73.39 \\ \midrule
\textbf{SNLI} & 65.97 & 68.80 & 72.18 & 70.26 & 70.69 & 70.59 & 68.88 \\ \midrule
\textbf{SICK-E} & 78.50 & 78.90 & 80.29 & 79.78 & 79.16 & 79.74 & 78.63 \\ \midrule\midrule
\textbf{Length} & 71.76 & 64.20 & 87.03 & 87.33 & 86.17 & 86.11 & 87.58\\ \midrule
\textbf{Word Content} & 80.61 & 82.10 & 59.68 & 60.44 & 62.63 & 59.28 & 62.60 \\ \midrule
\textbf{Depth} & 36.50 & 36.38 & 43.93 & 44.67 & 44.21 & 41.41 & 43.80 \\ \midrule
\textbf{Top Constituents} & 66.09 & 66.34 & 75.23 & 76.02 & 75.91 & 75.46 & 77.72 \\ \midrule
\textbf{Bigram Shift} & 49.90 & 49.67 & 90.84 & 85.89 & 85.75 & 88.96 & 85.96 \\ \midrule
\textbf{Tense} & 85.34 & 87.18 & 88.56 & 88.14 & 87.88 & 89.06 & 88.80 \\ \midrule
\textbf{Subj Number} & 79.26 & 80.78 & 86.89 & 87.81 & 87.44 & 85.53 & 86.44 \\ \midrule
\textbf{Obj Number} & 77.66 & 80.29 & 84.49 & 84.46 & 84.80 & 83.44 & 83.42 \\ \midrule
\textbf{Odd Man Out} & 53.15 & 49.96 & 68.65 & 62.45 & 61.67 & 65.86 & 60.99 \\ \midrule
\textbf{Coordination Inversion} & 54.13 & 52.23 & 73.87 & 70.13 & 70.33 & 72.36 & 69.65 \\ \bottomrule
\end{tabular}
\end{table}

\begin{table}[ht]
    \centering
    \caption{Accuracy on the entire downstream dataset of \texttt{scientific abstracts} classification after joint training on the entire pre-training dataset of \texttt{scientific abstracts}. The \emph{scratch} term indicates that the model is randomly initialized at the beginning and not pre-trained on Wikipedia.}
    \begin{tabular}{lcc}
        \toprule
        \textbf{Model} & \textbf{Accuracy} & \textbf{1-epoch Accuracy} \\ 
        \toprule
        \textbf{RoBERTa Base} & 82.25 & 79.27  \\ \midrule
        \textbf{BERT Base} & 82.57 & 79.37 \\ \midrule\midrule
        \textbf{RoBERTa NT} & 81.84 & 77.88 \\ \midrule
        \textbf{RoBERTa Pr.} & 82.26 & 81.01 \\ \midrule
        \textbf{BERT Pr.} & 83.49 & 82.62 \\ \midrule
        \textbf{RoBERTa Pr. NT} & 83.51 & 81.94  \\ \midrule\midrule
        \textbf{RoBERTa scratch} & 80.48 & 75.79 \\ \midrule
        \textbf{RoBERTa scratch Pr.} & 82.50 & 81.50  \\
        \bottomrule
    \end{tabular}
    \label{tab:joint}
\end{table}

\begin{table}[ht]
    \centering
    \caption{Accuracy on the downstream dataset of \texttt{scientific abstracts} classification after continual pre-training. The split used for downstream classification and pre-training contains different documents. The digit next to the model indicates the last experience the model has been trained on (e.g., 5 means that the model has been pre-trained on all $5$ experiences sequentially).}
    \begin{tabular}{lcc}
        \toprule
        \textbf{Model} & \textbf{Accuracy} & \textbf{1-epoch Accuracy} \\ 
        \toprule
        \textbf{RoBERTa Pr. 1} & 82.59 & 79.88 \\ \midrule
        \textbf{BERT Pr. 1} & 82.64 & 80.91 \\ \midrule
        \textbf{RoBERTa Pr. NT 1} & 82.37 & 80.58  \\ \midrule\midrule
        \textbf{RoBERTa Pr. 5} & 83.24 & 81.19 \\ \midrule
        \textbf{BERT Pr. 5} & 83.08 & 81.84 \\ \midrule
        \textbf{RoBERTa Pr. NT 5} & 83.06 & 81.22 \\
        \bottomrule
    \end{tabular}
    \label{tab:cl-entire-abstract}
\end{table}

\begin{table}[ht]
    \centering
    \caption{ACC on \texttt{scientific abstracts} classification for 5 experiences with RoBERTa. Pre-trained only on the first experience of \texttt{scientific abstracts} dataset. Replay memory size is 500. Joint training from Table \ref{tab:joint}. ACC around $20.00$ means complete forgetting (only the last task is correctly classified).}
    \begin{tabular}{l|c|ccc}
        \toprule
        \textbf{Model} & \textbf{Joint} & \textbf{Naive} & \textbf{Replay} &  \textbf{DSLDA} \\ 
        \toprule
        \textbf{RoBERTa Base} & 80.00 & 19.95 & 52.94 & 69.22\\
        \textbf{RoBERTa Pr.} & 82.26 & 19.90 & 50.78 & 72.03 \\
        \textbf{RoBERTa Pr. NT} & 83.51 & 19.90 & 51.37 & 73.32 \\
        \bottomrule
    \end{tabular}
    \label{tab:cl-strategies}
\end{table}

\begin{table}[ht]
    \caption{Fine-tuning accuracy on the entire dataset of \texttt{CORe50} with large transformers. Pre-training has been performed sequentially over each experience of \texttt{iNaturalist}.}
    \label{tab:cl-inaturalist2core-large}
    \centering
    \begin{tabular}{lccccc|ccccc}
\toprule
\textbf{Model} & \multicolumn{5}{c}{\textbf{Accuracy}} & \multicolumn{5}{c}{\textbf{1-epoch Accuracy}}  \\ \toprule
\textbf{ViT Base} & \multicolumn{5}{c}{92.95} &  \multicolumn{5}{c}{90.77} \\ \midrule
\textbf{BEiT Base} & \multicolumn{5}{c}{90.41} &  \multicolumn{5}{c}{89.41} \\ \midrule
Exp.      & e1 & e2 & e3 & e4 & e5         & e1 & e2 & e3 & e4 & e5        \\ \midrule
\textbf{ViT Pr.}            & 91.50  & 89.37 & 89.93 & 89.12 & 87.72           & 91.39  & 89.22  & 89.30 & 89.12 & 87.70                   \\ \midrule
\textbf{BEiT Pr.}            & 89.78 & 89.90 & 89.18 & 88.50 & 90.09              & 86.81 & 85.94 & 87.50 & 88.50 & 88.50                \\ \bottomrule
\end{tabular}
\end{table}

\begin{table}[ht]
    \caption{Linear evaluation accuracy on the entire dataset of \texttt{CORe50} with large Transformers. Pre-training has been performed sequentially over each experience of \texttt{iNaturalist}.}
    \label{tab:cl-inaturalist2core-linear-large}
    \centering
    \begin{tabular}{lccccc}
\toprule
\textbf{Model} & \multicolumn{5}{c}{\textbf{Accuracy}}  \\ \toprule
\textbf{ViT Base} & \multicolumn{5}{c}{82.39} \\ \midrule
\textbf{BEiT Base} & \multicolumn{5}{c}{52.04} \\ \midrule
Exp.      & e1 & e2 & e3 & e4 & e5       \\ \midrule
\textbf{ViT Pr.}            & 85.62 & 73.75 & 73.73 & 75.89 & 68.27                         \\ \midrule
\textbf{BEiT Pr.}            & 56.67 & 55.62 & 56.12 & 55.74 & 56.76                 \\ \bottomrule
\end{tabular}
\end{table}

\subsection{Effect of Pre-Training on the Downstream Domain Task} \label{app:downstream}
Table \ref{tab:joint} shows the accuracy on the entire dataset of \texttt{scientific abstracts} classification after pre-training on the entire datasets of \texttt{scientific abstracts} (held-out sets). Therefore, this setup uses only one step of pre-training to assess its effectiveness on the performance on the downstream task. We show that pre-training is beneficial to the final performance with respect to the original model pre-trained on Wikipedia.

Similarly, Table \ref{tab:cl-entire-abstract} shows the impact of $1$ and $5$ steps of continual pre-training on the dataset of \texttt{scientific abstracts} classification. Each fine-tuning step is performed on the corresponding split of the \texttt{scientific abstract dataset}. Again, we see a moderate improvement in the final performance. \\
It is important to note that the improvement, although small, is nonetheless present even if each experience of continual pre-training contains a smaller set of samples with respect to the pre-training dataset typically used in the NLP literature, like Wikipedia. For each experience, we have $20,000$ samples. This aspect is particularly important for continual learning, where the model is not updated one-shot with a large dataset, but in multiple steps with few samples.

\subsection{Results With Traditional CL scenario} \label{app:cl}
Table \ref{tab:cl-strategies} shows that in a traditional CL setup, fine-tuning a single model on \texttt{scientific abstracts} classification tasks continuously leads to large forgetting on the same \texttt{scientific abstracts} classification task (held-out dataset), unless CL strategies are employed. We measure the popular ACC metric \citep{lopez-paz2017} which computes the accuracy on all tasks after training on the last task. The lower its value, the larger the forgetting effect. This shows that, although in the traditional CL scenario we always have a model ready to tackle all the previous tasks without retraining, the loss in terms of performance (accuracy in this case) is very large with respect to the continual pre-training scenario.

\subsection{CKA Plots} \label{app:cka}
\begin{figure}
     \centering
     \begin{subfigure}[b]{0.24\textwidth}
         \centering
         \includegraphics[width=\textwidth]{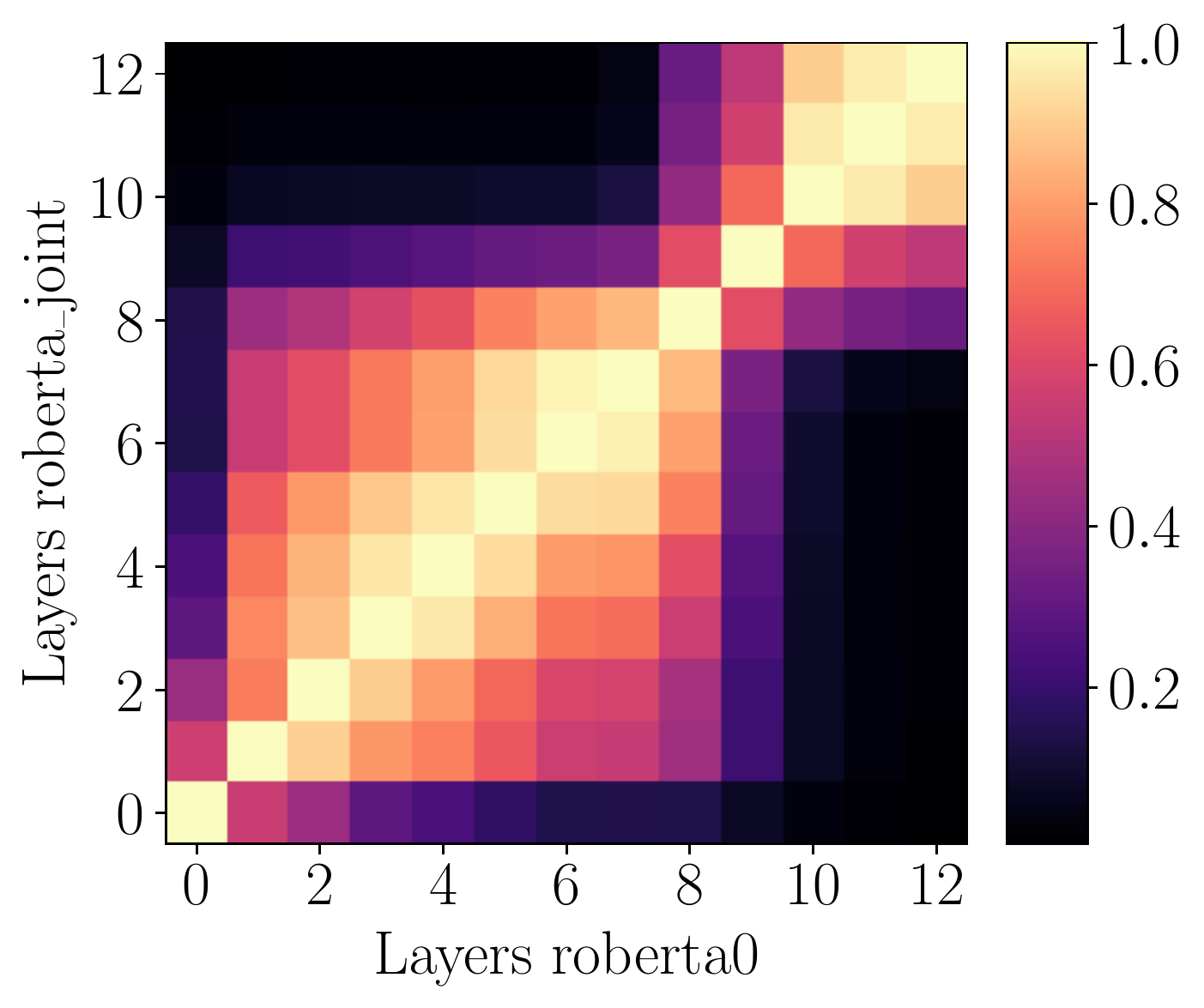}
         \caption{RoBERTa QNLI 1}
     \end{subfigure}
     \hfill
     \begin{subfigure}[b]{0.24\textwidth}
         \centering
         \includegraphics[width=\textwidth]{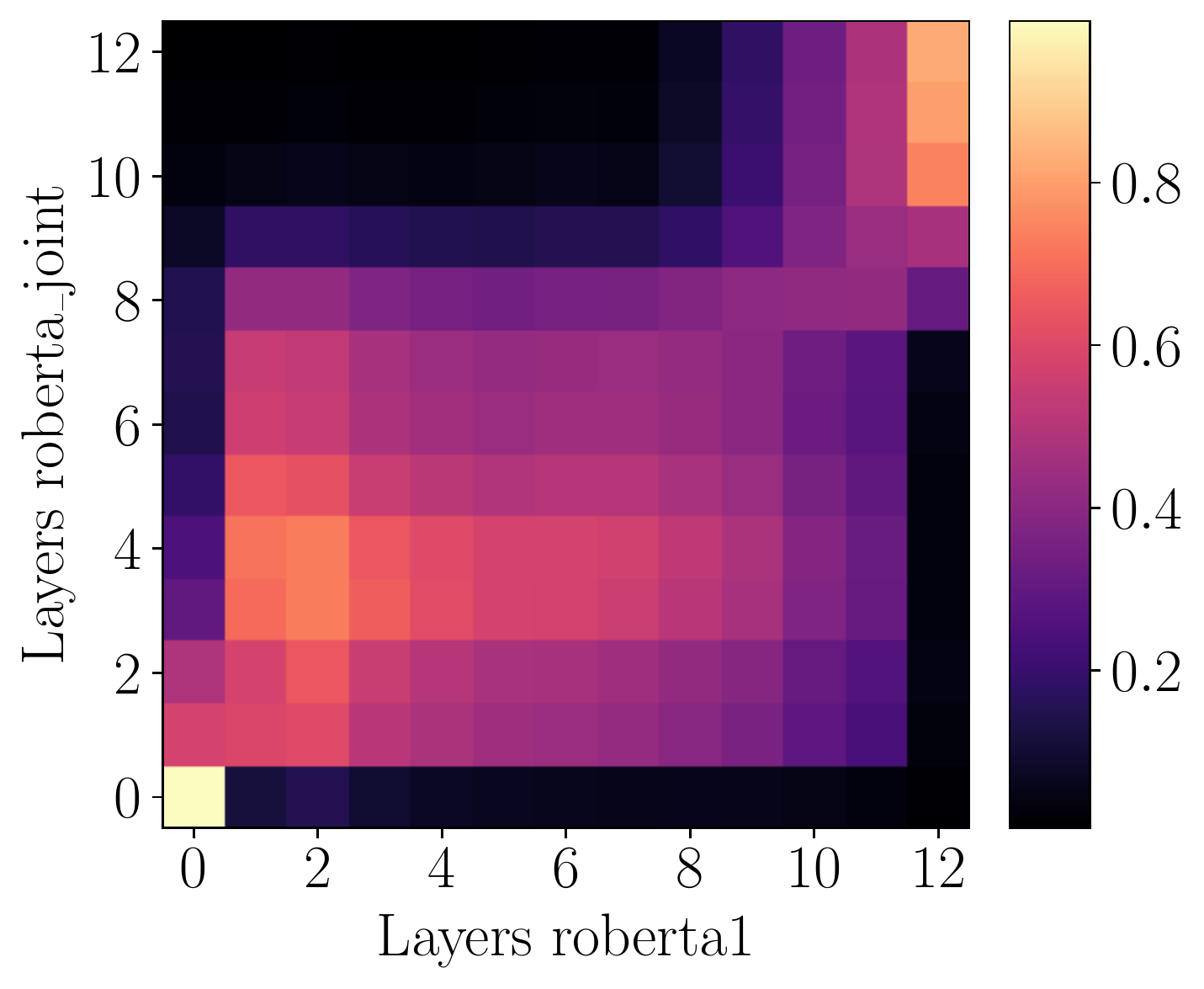}
         \caption{RoBERTa QNLI 2}
     \end{subfigure}
     \hfill
     \begin{subfigure}[b]{0.24\textwidth}
         \centering
         \includegraphics[width=\textwidth]{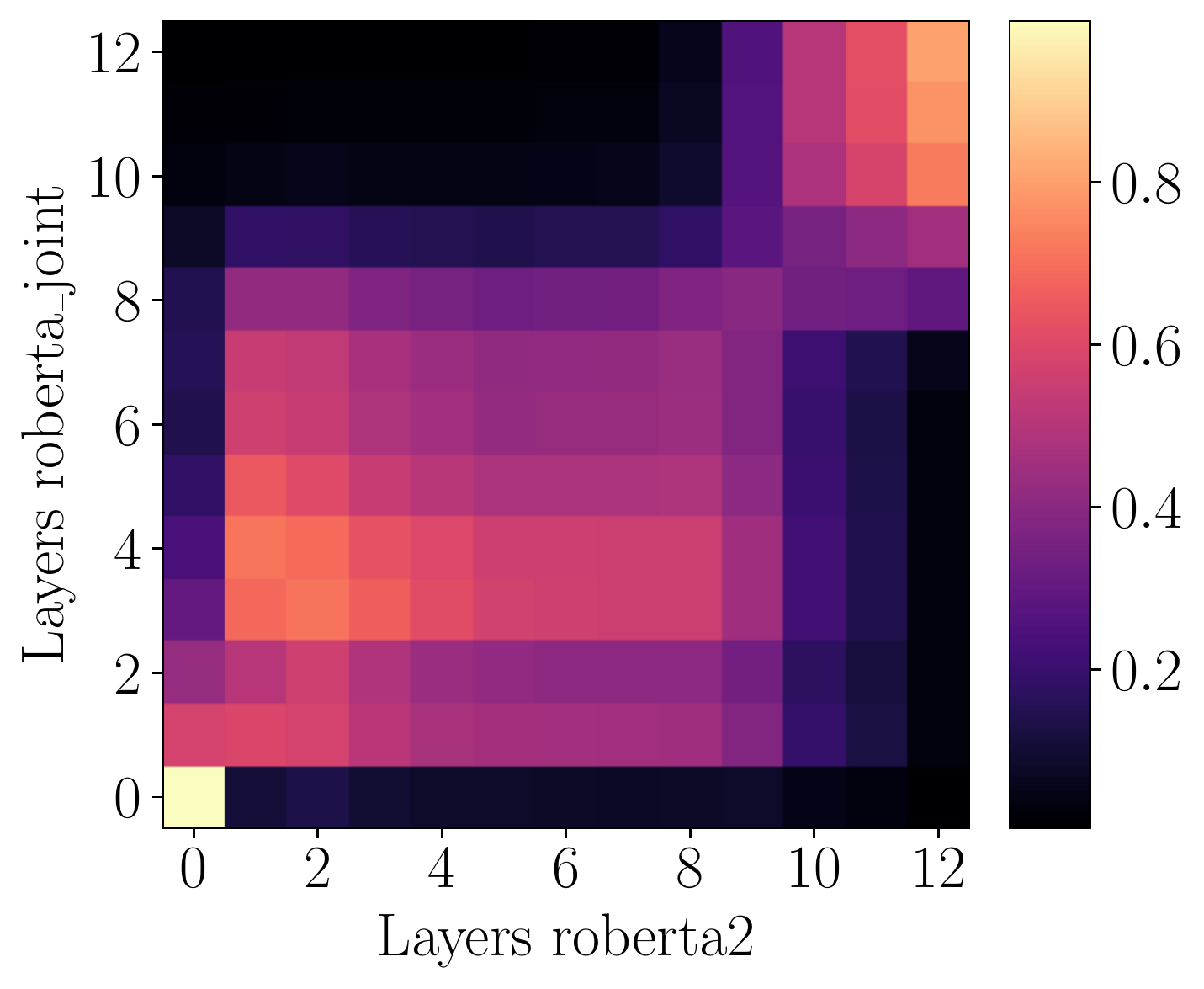}
         \caption{RoBERTa QNLI 3}
     \end{subfigure}
     \hfill
     \begin{subfigure}[b]{0.24\textwidth}
         \centering
         \includegraphics[width=\textwidth]{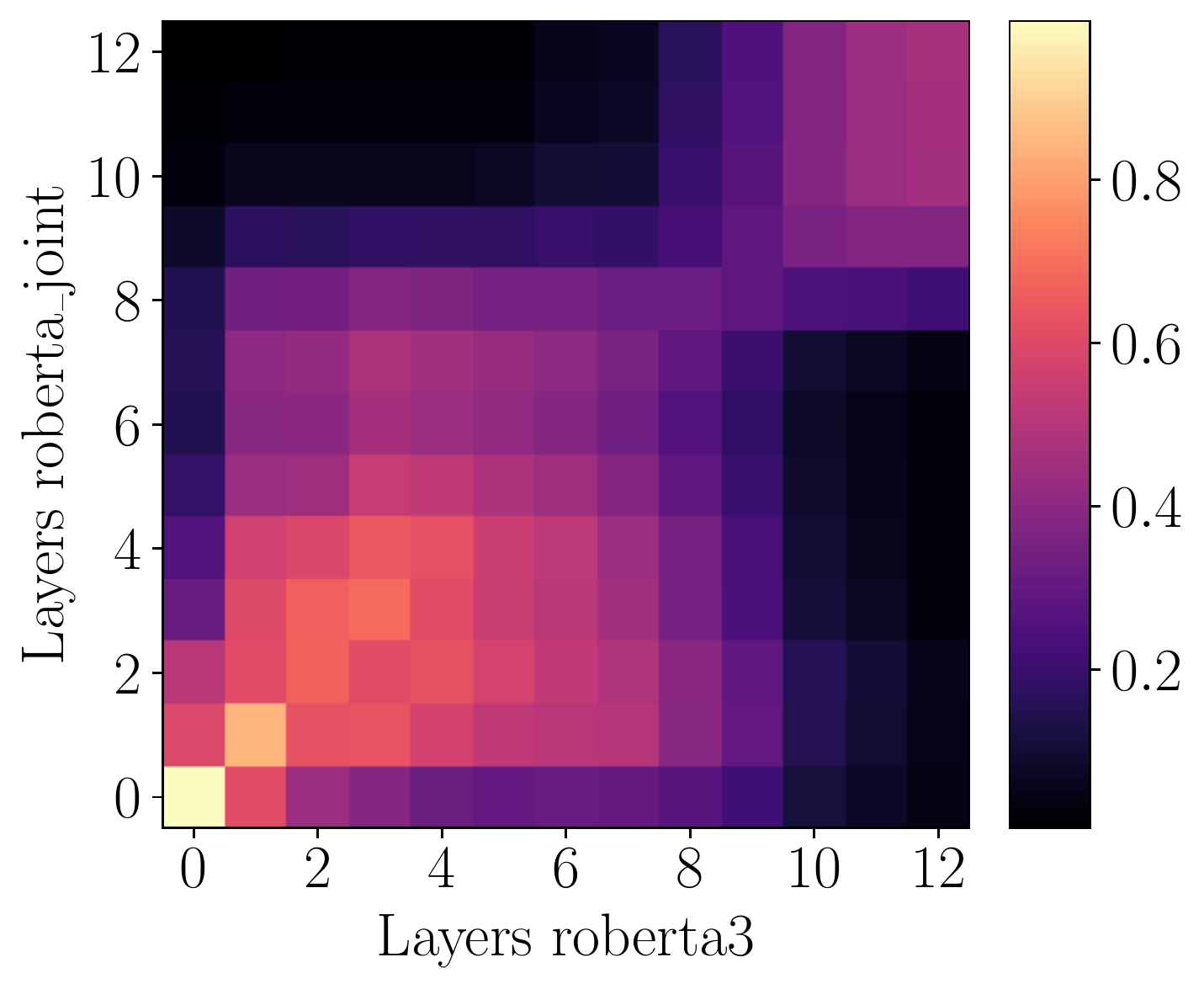}
         \caption{RoBERTa QNLI 4}
     \end{subfigure}
     \begin{subfigure}[b]{0.24\textwidth}
         \centering
         \includegraphics[width=\textwidth]{img/roberta_qnli/matrix4.pdf}
         \caption{RoBERTa QNLI 5}
     \end{subfigure}
     \hfill
     \begin{subfigure}[b]{0.24\textwidth}
         \centering
         \includegraphics[width=\textwidth]{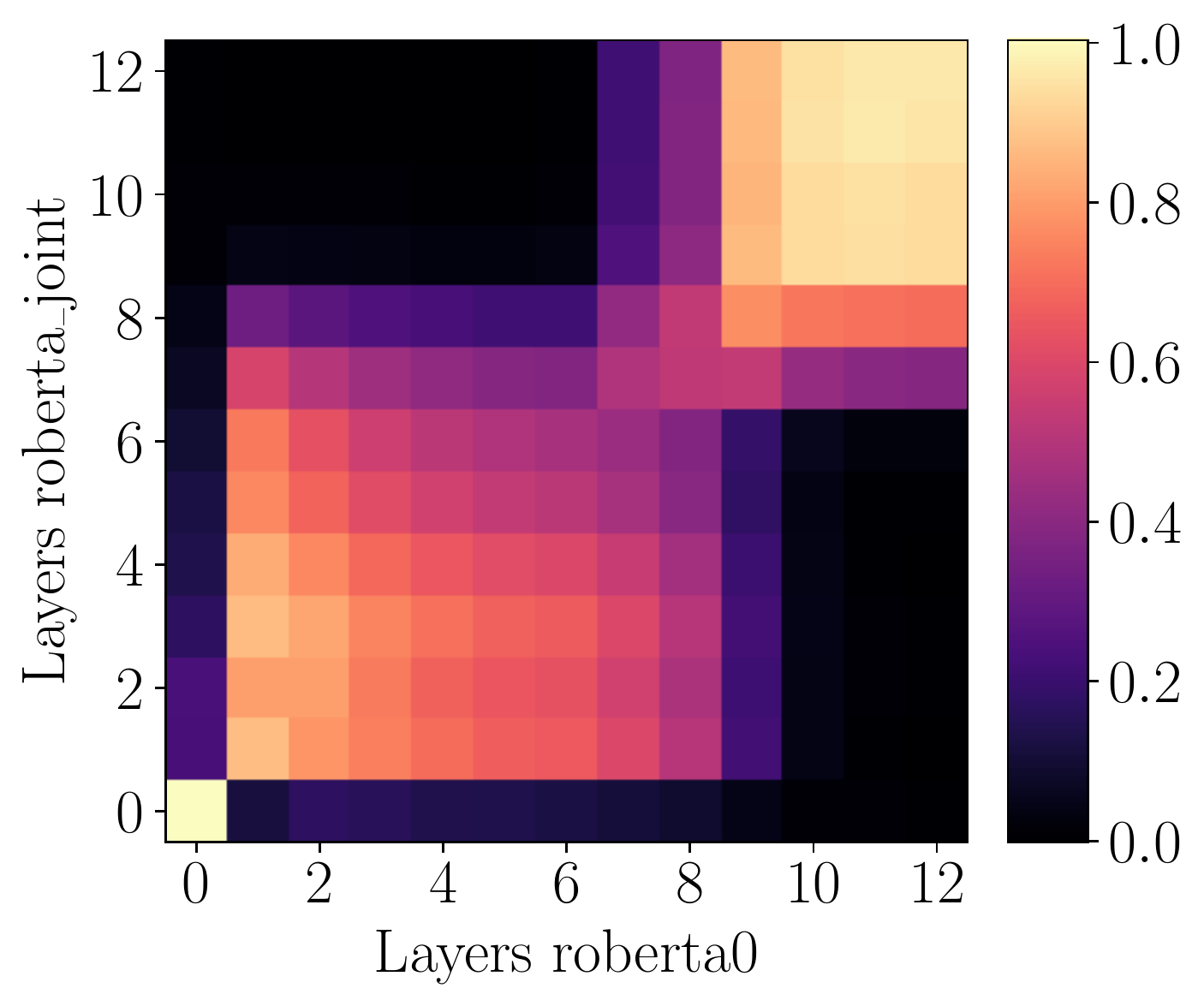}
         \caption{RoBERTa Tweets 1}
     \end{subfigure}
     \hfill
     \begin{subfigure}[b]{0.24\textwidth}
         \centering
         \includegraphics[width=\textwidth]{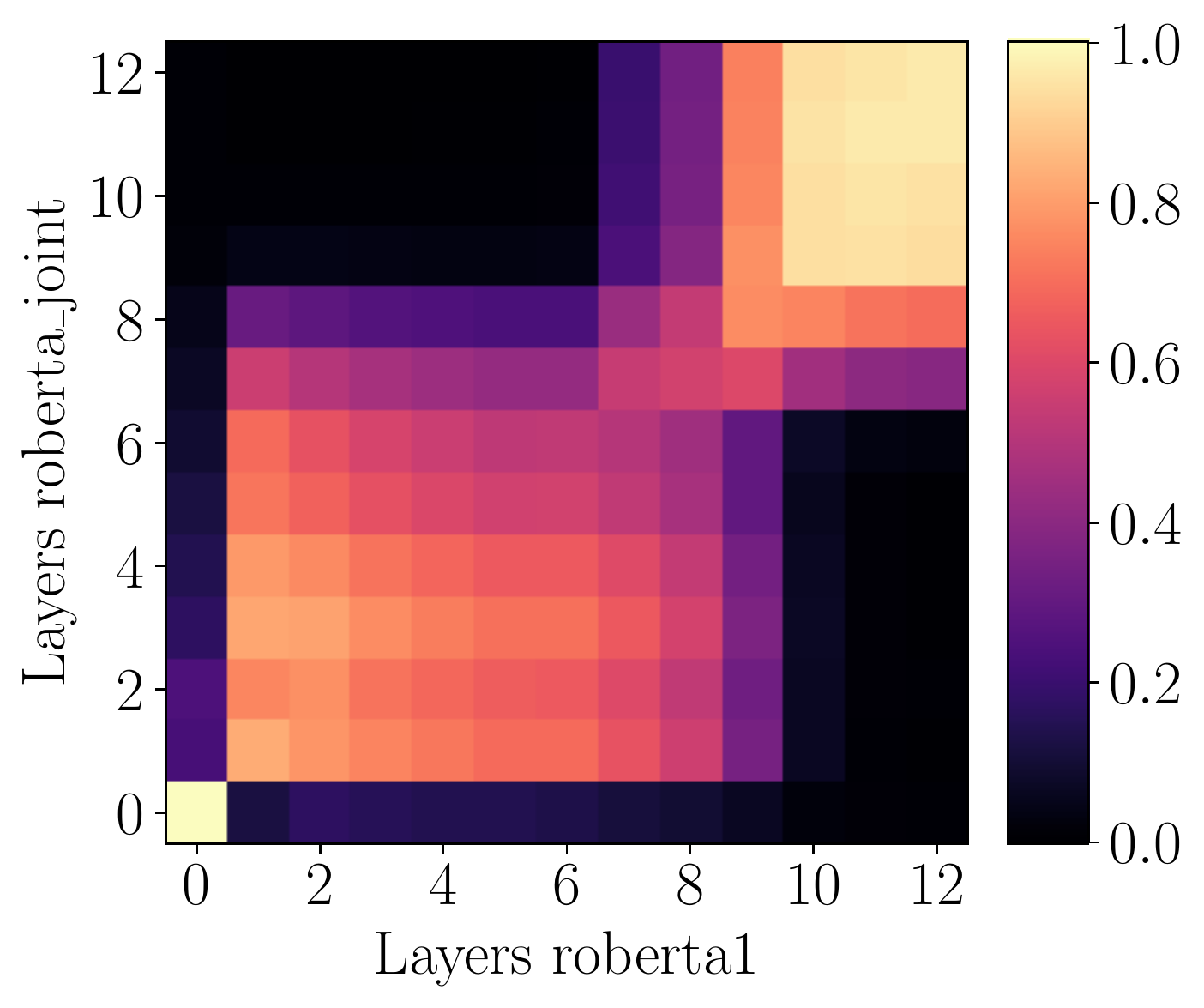}
         \caption{RoBERTa Tweets 2}
     \end{subfigure}
     \hfill
     \begin{subfigure}[b]{0.24\textwidth}
         \centering
         \includegraphics[width=\textwidth]{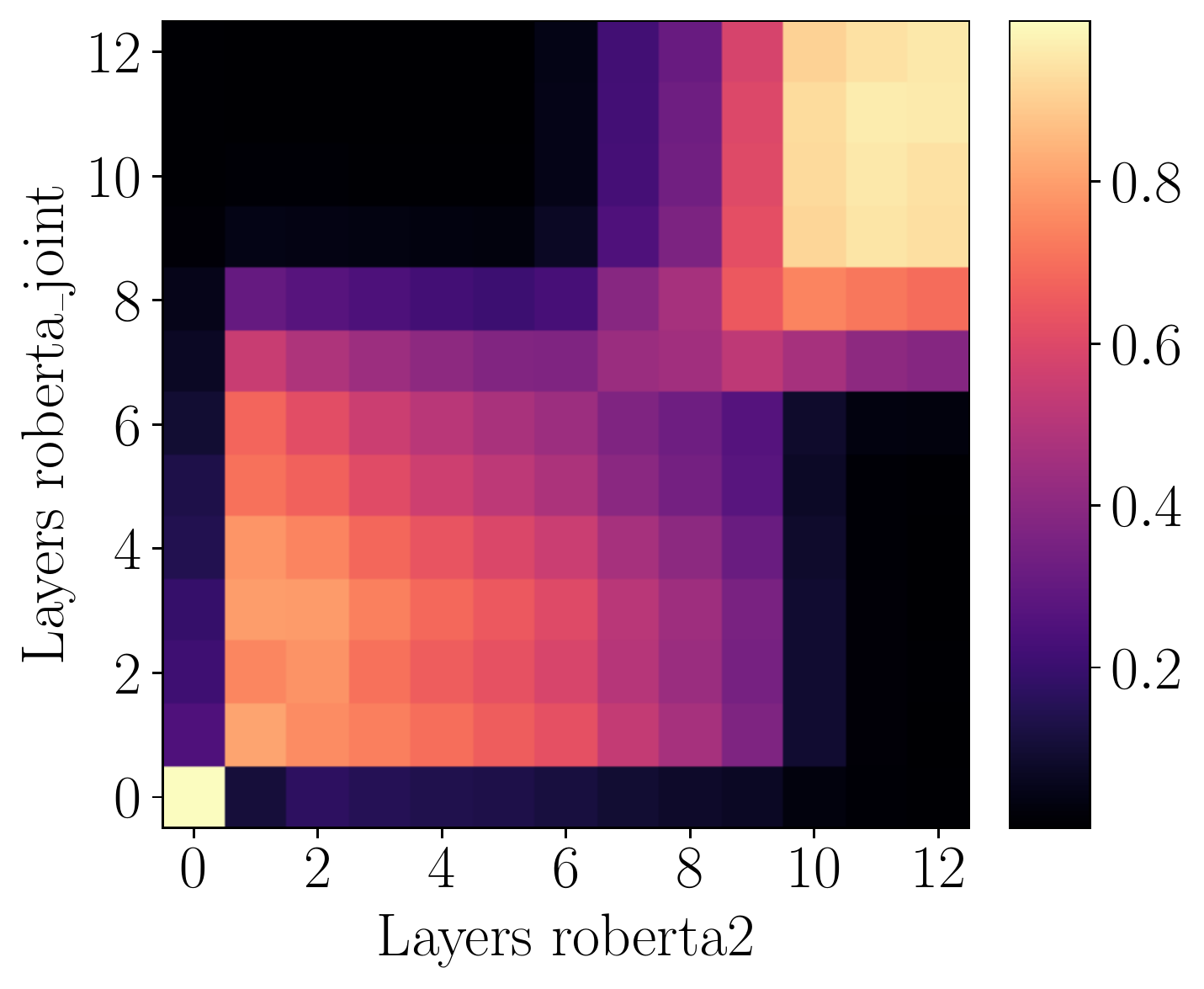}
         \caption{RoBERTa Tweets 3}
     \end{subfigure}
     \begin{subfigure}[b]{0.24\textwidth}
         \centering
         \includegraphics[width=\textwidth]{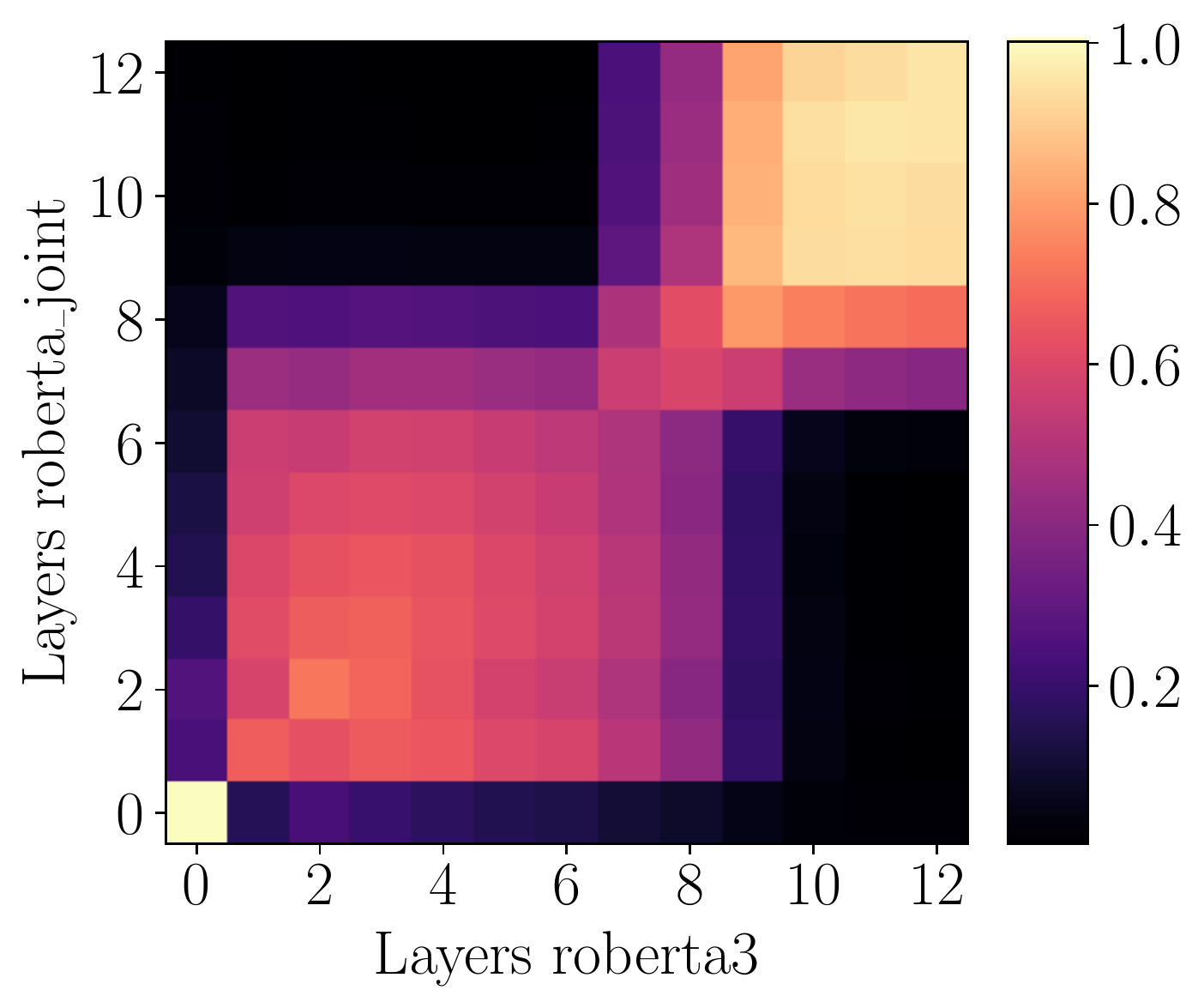}
         \caption{RoBERTa Tweets 4}
     \end{subfigure}
     \hfill
     \begin{subfigure}[b]{0.24\textwidth}
         \centering
         \includegraphics[width=\textwidth]{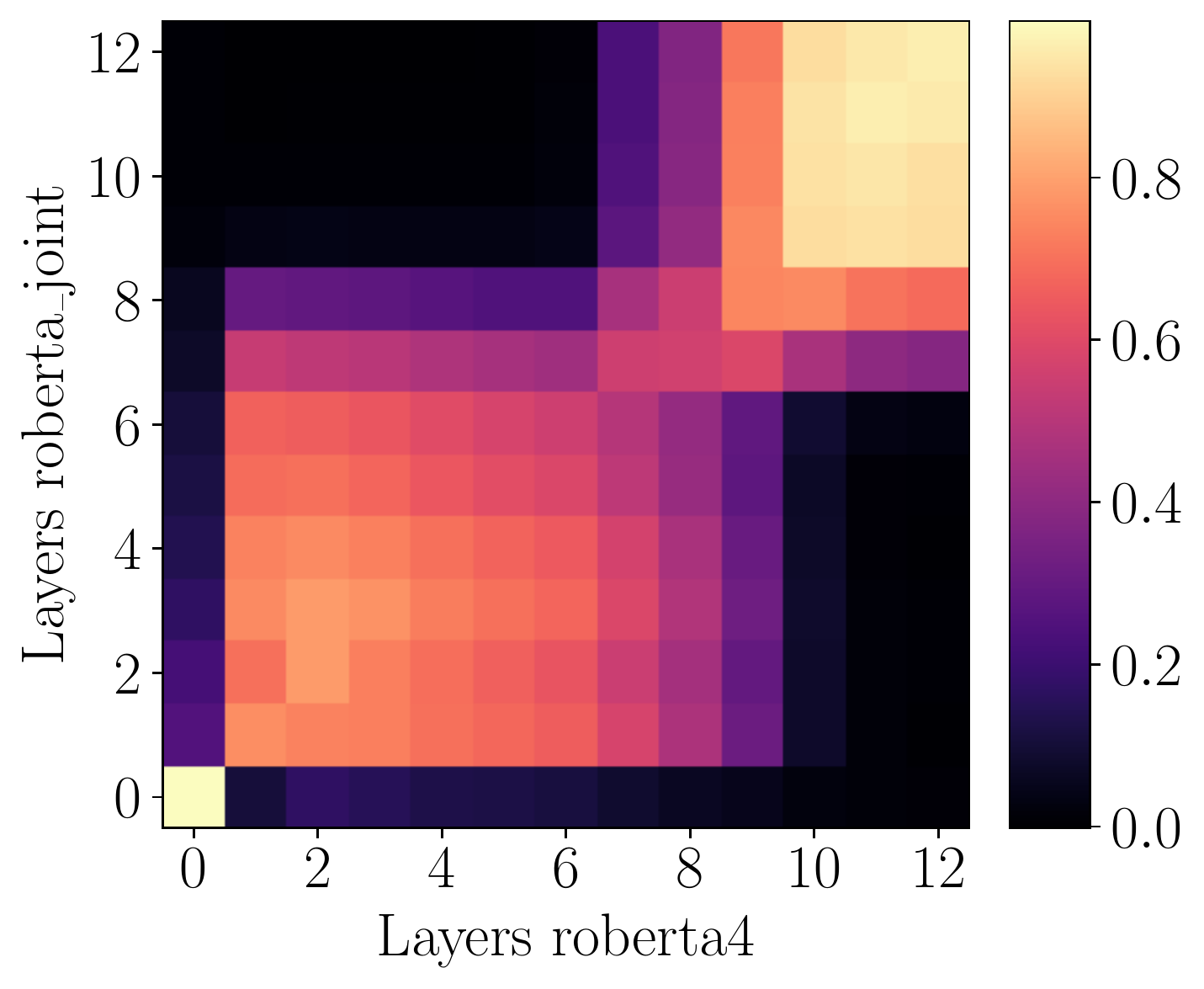}
         \caption{RoBERTa Tweets 5}
     \end{subfigure}
     \hfill
     \begin{subfigure}[b]{0.24\textwidth}
         \centering
         \includegraphics[width=\textwidth]{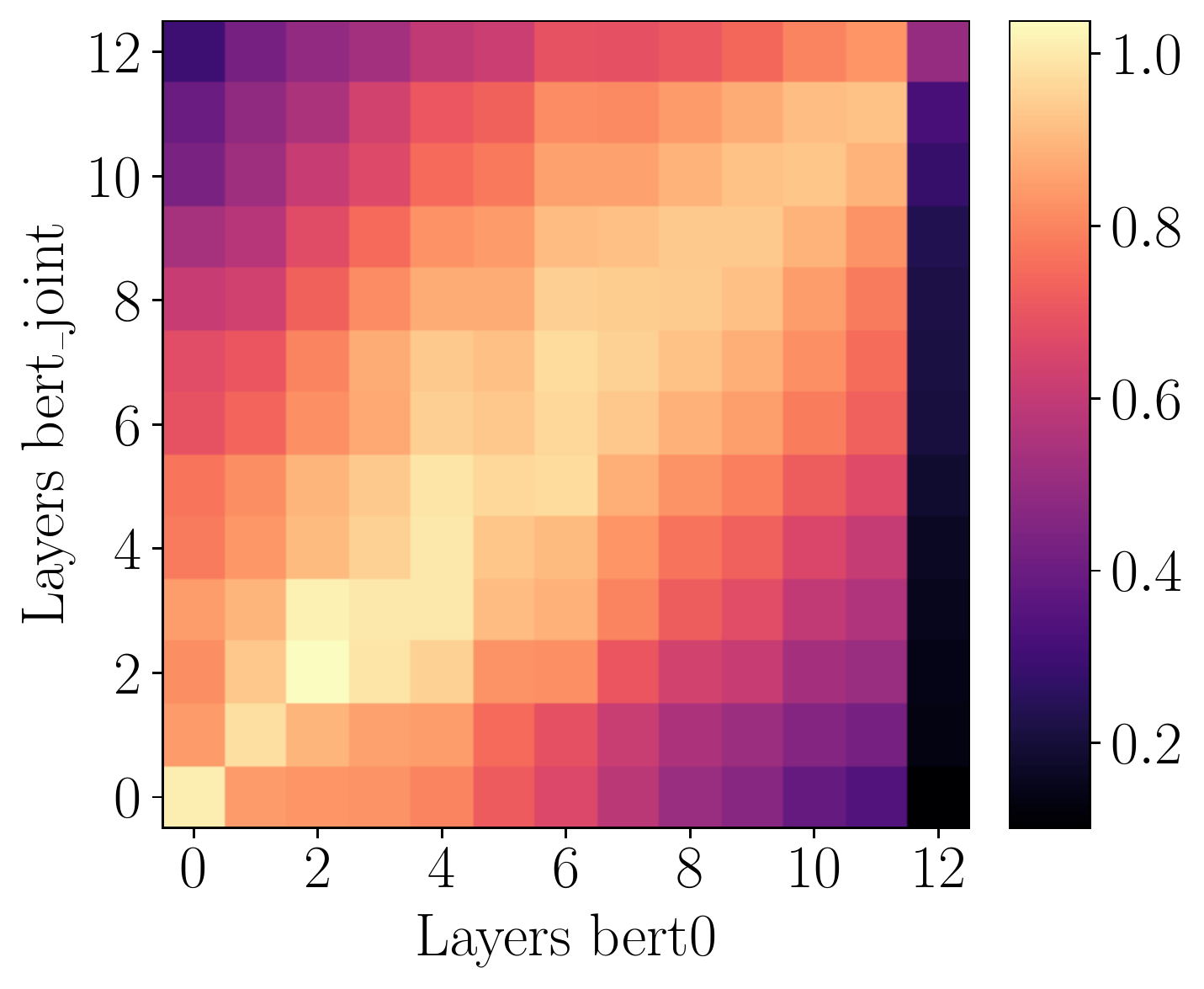}
         \caption{BERT QNLI 1}
     \end{subfigure}
     \hfill
     \begin{subfigure}[b]{0.24\textwidth}
         \centering
         \includegraphics[width=\textwidth]{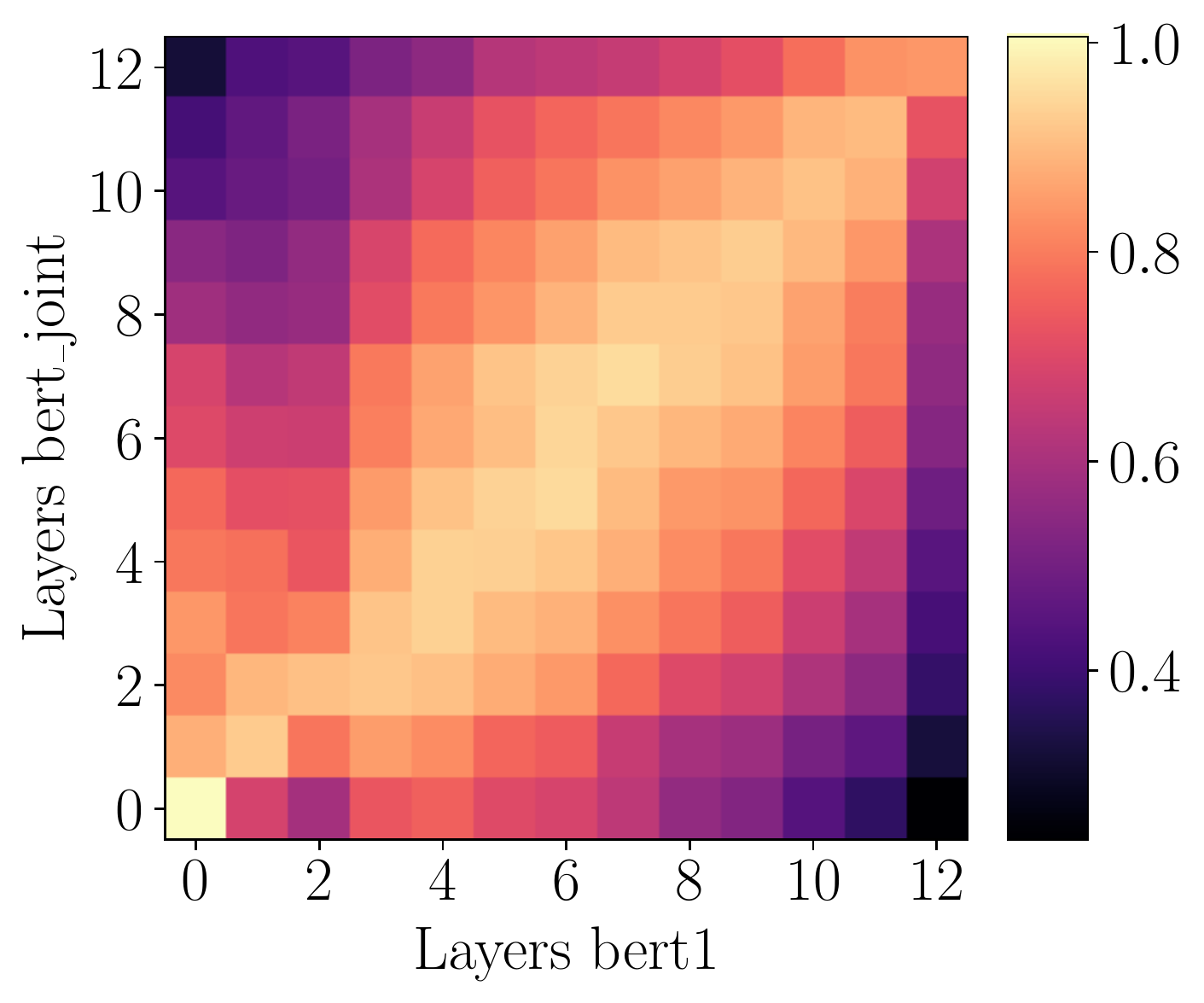}
         \caption{BERT QNLI 2}
     \end{subfigure}
     \begin{subfigure}[b]{0.24\textwidth}
         \centering
         \includegraphics[width=\textwidth]{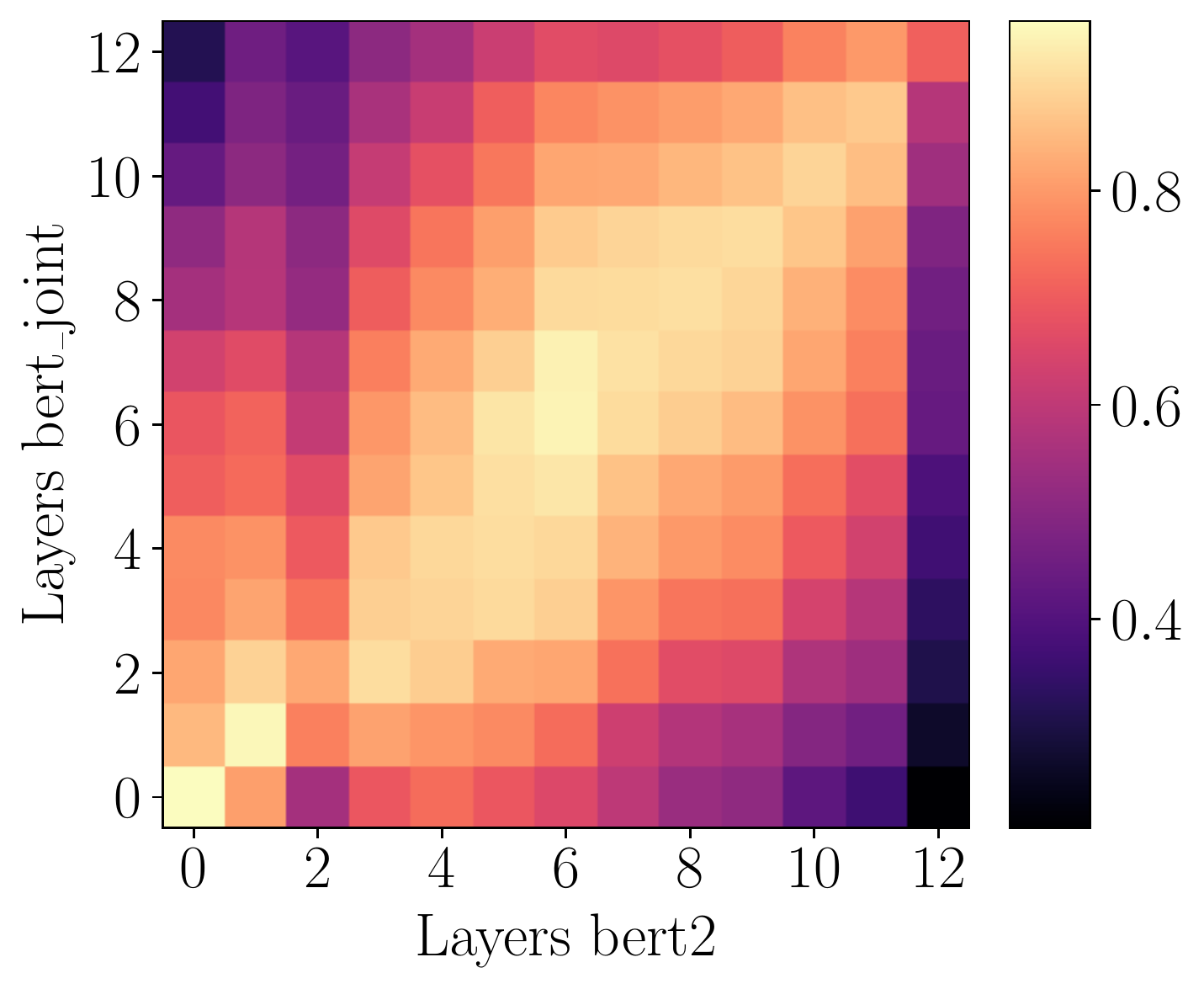}
         \caption{BERT QNLI 3}
     \end{subfigure}
     \begin{subfigure}[b]{0.24\textwidth}
         \centering
         \includegraphics[width=\textwidth]{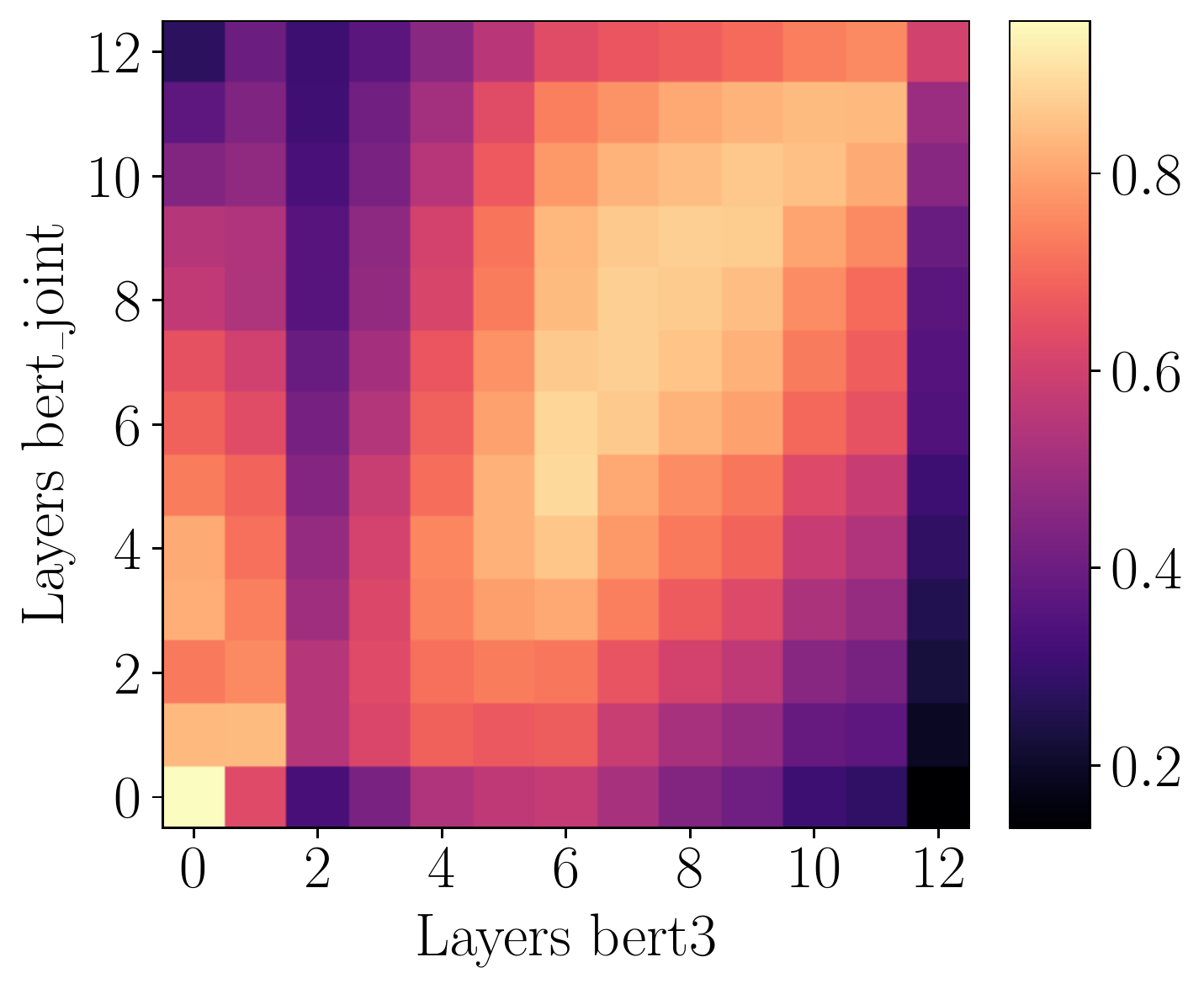}
         \caption{BERT QNLI 4}
     \end{subfigure}
     \hfill
     \begin{subfigure}[b]{0.24\textwidth}
         \centering
         \includegraphics[width=\textwidth]{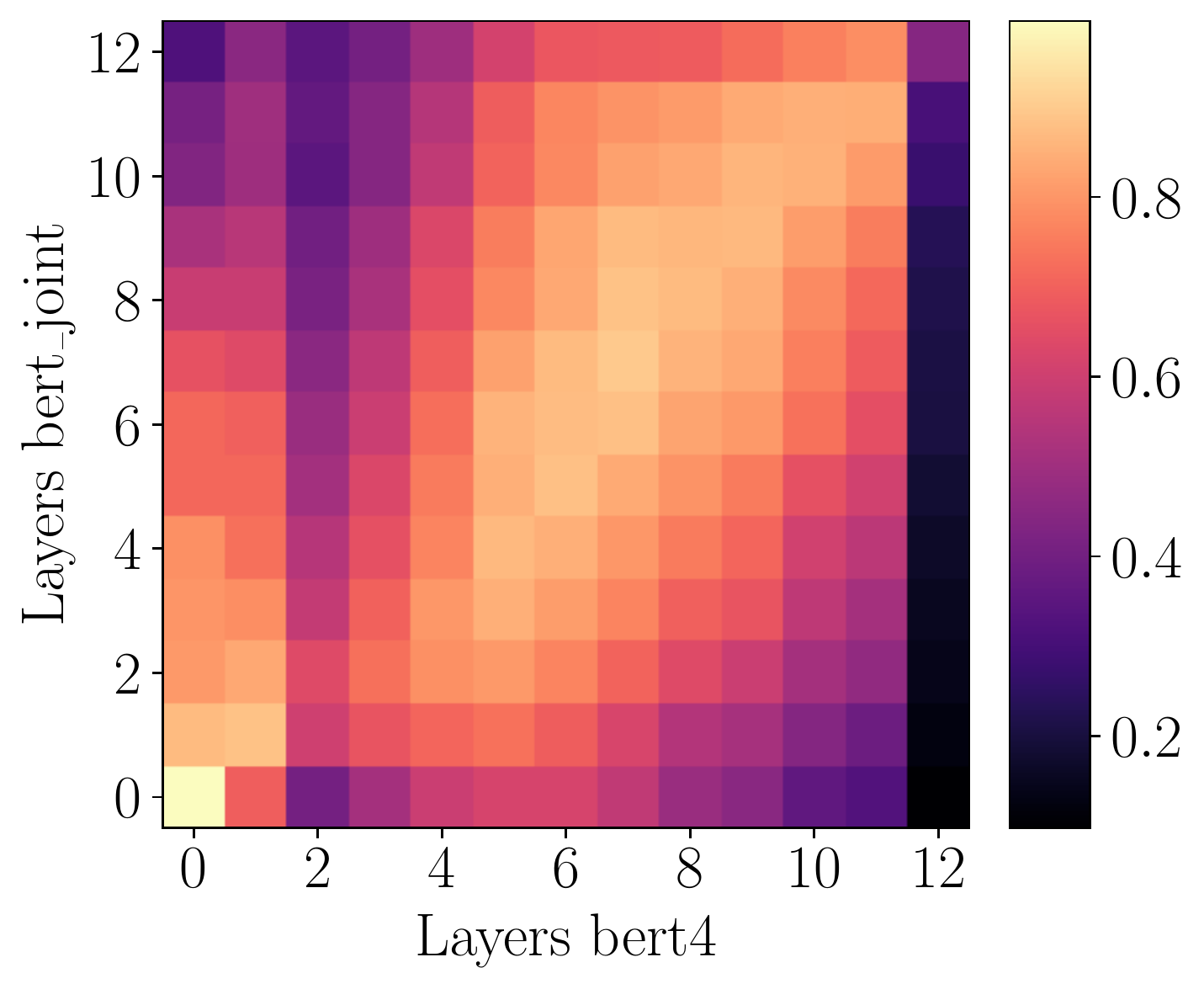}
         \caption{BERT QNLI 5}
     \end{subfigure}
     \hfill
     \begin{subfigure}[b]{0.24\textwidth}
         \centering
         \includegraphics[width=\textwidth]{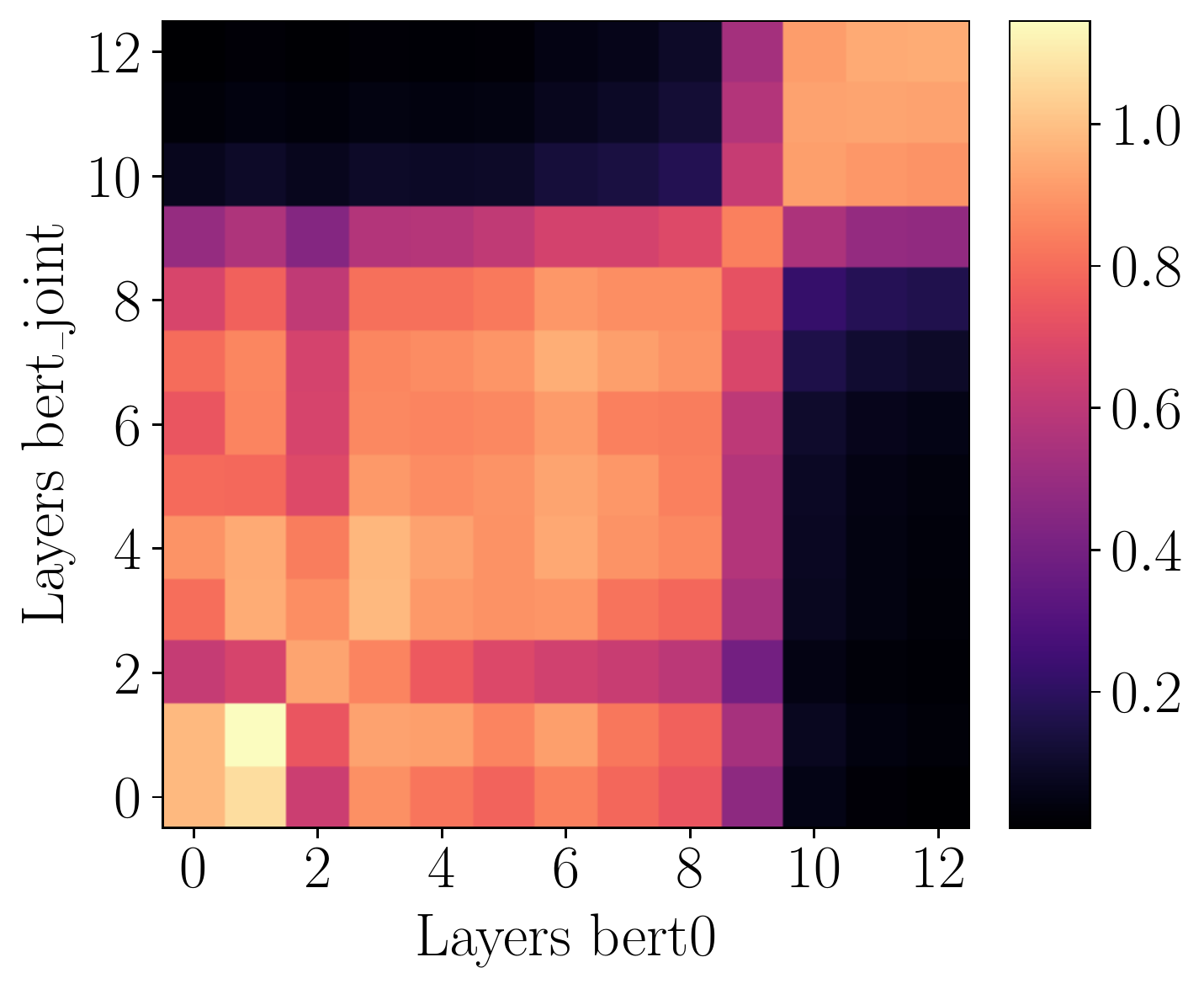}
         \caption{BERT Tweets 1}
     \end{subfigure}
     \hfill
     \begin{subfigure}[b]{0.24\textwidth}
         \centering
         \includegraphics[width=\textwidth]{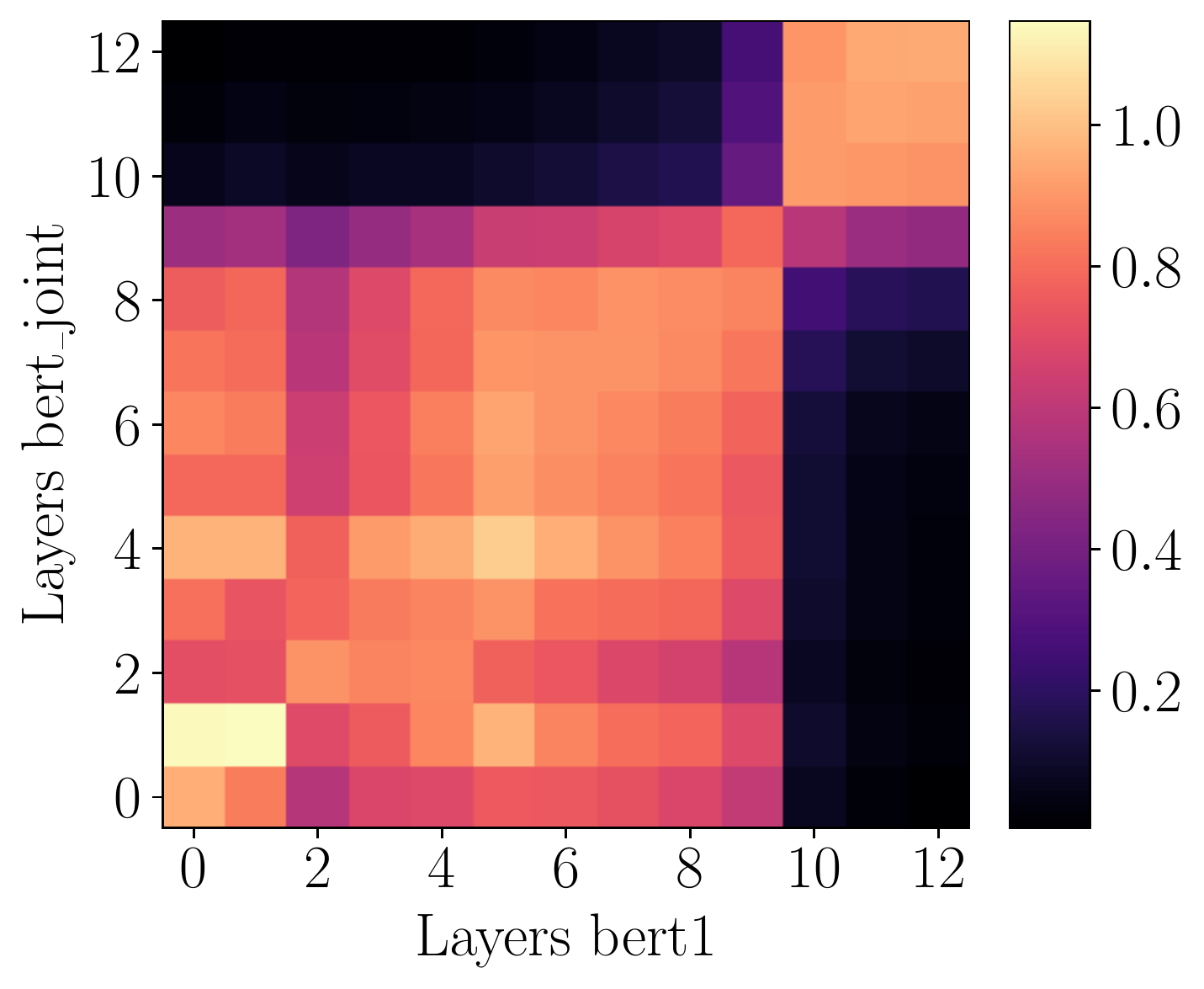}
         \caption{BERT Tweets 2}
     \end{subfigure}
     \begin{subfigure}[b]{0.24\textwidth}
         \centering
         \includegraphics[width=\textwidth]{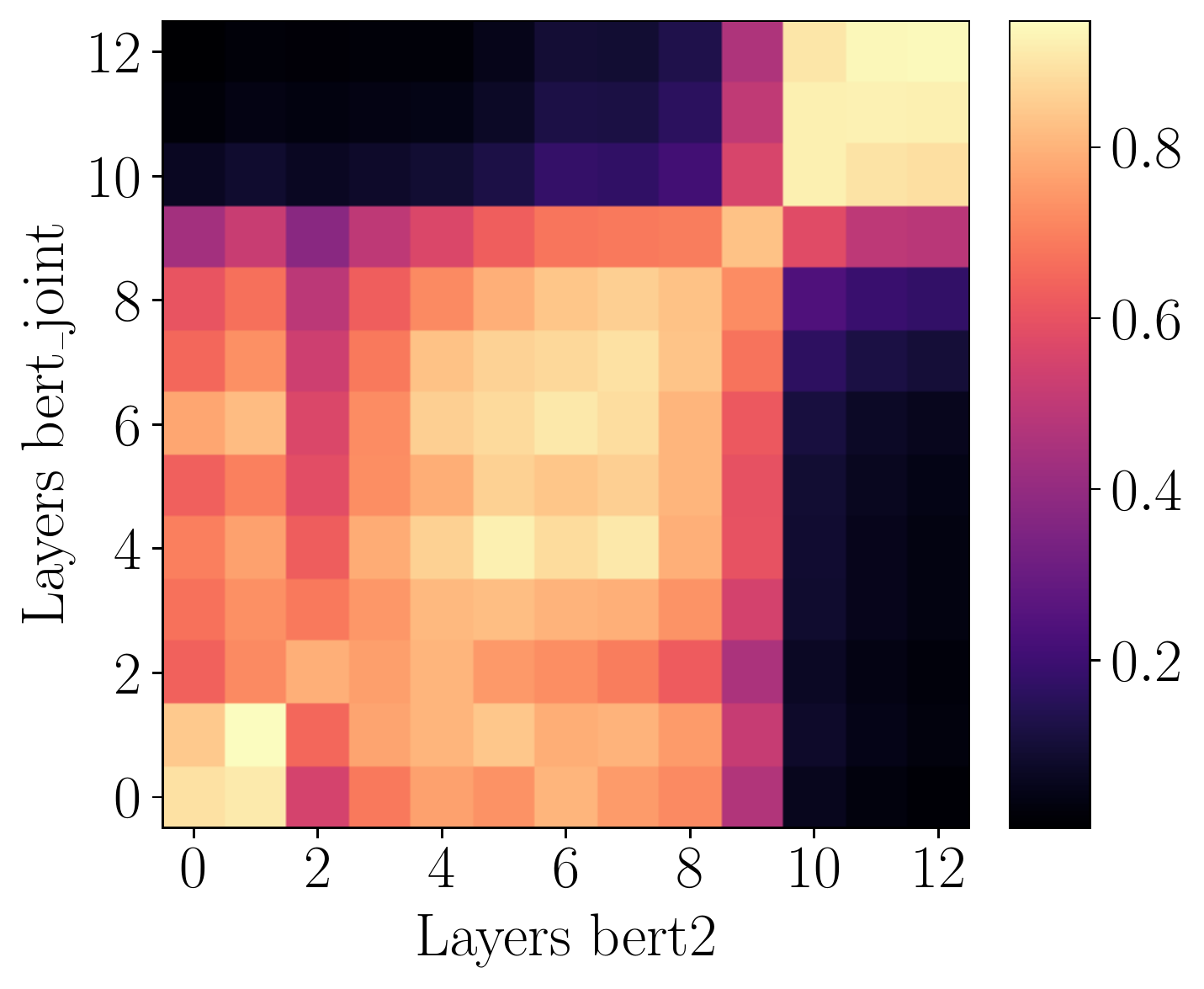}
         \caption{BERT Tweets 3}
     \end{subfigure}
     \begin{subfigure}[b]{0.24\textwidth}
         \centering
         \includegraphics[width=\textwidth]{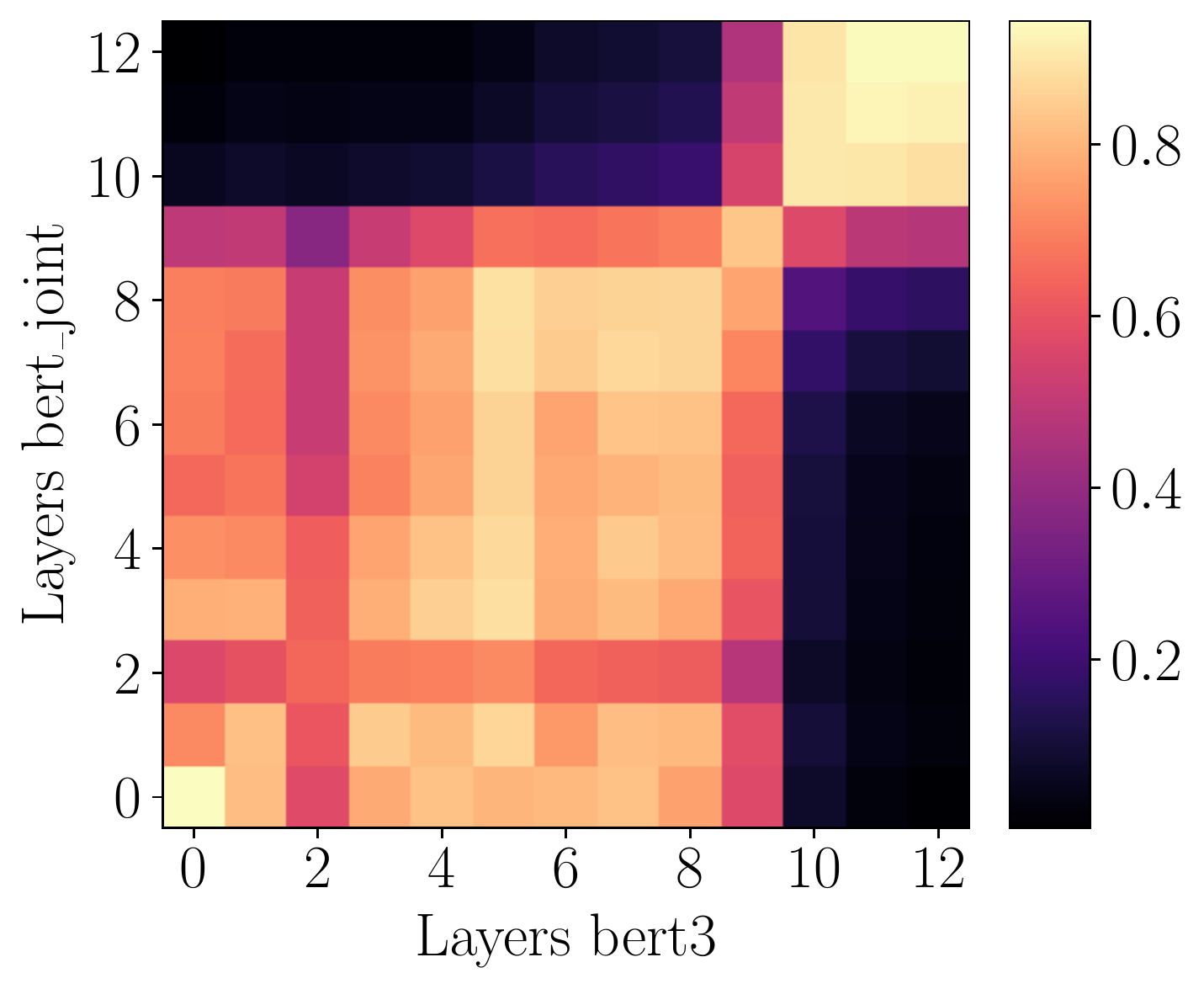}
         \caption{BERT Tweets 4}
     \end{subfigure}
     \hfill
     \begin{subfigure}[b]{0.24\textwidth}
         \centering
         \includegraphics[width=\textwidth]{img/bert_tweets/matrix4.pdf}
         \caption{BERT Tweets 5}
     \end{subfigure}
    \caption{CKA for RoBERTa and BERT. Pre-trained models after each experience are compared with the original pre-trained model.}
    \label{fig:cka-nlp}
    \end{figure}

\begin{figure}
    \centering
     \begin{subfigure}[b]{0.24\textwidth}
         \centering
         \includegraphics[width=\textwidth]{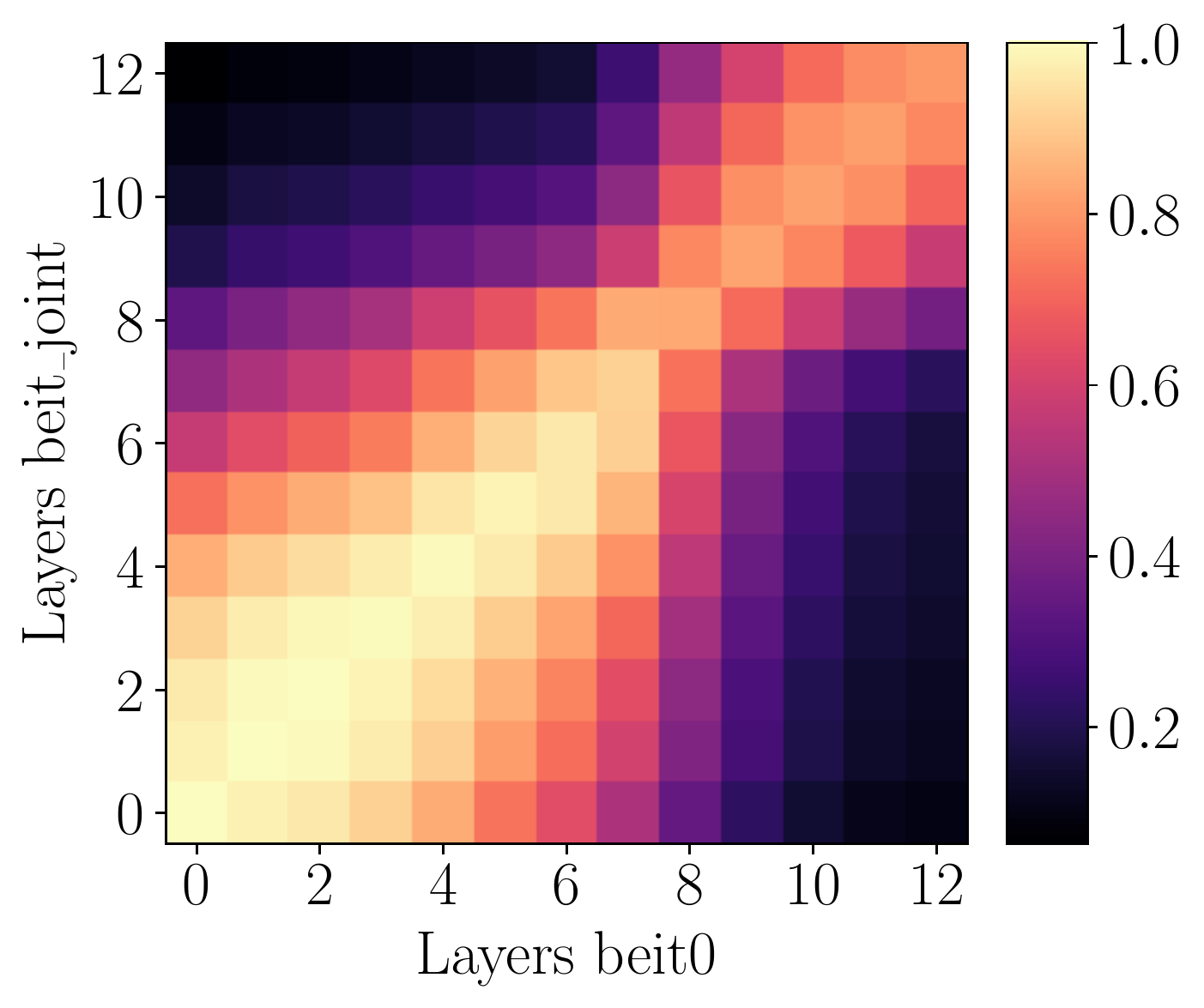}
         \caption{BEiT 1}
     \end{subfigure}
    \hfill
     \begin{subfigure}[b]{0.24\textwidth}
         \centering
         \includegraphics[width=\textwidth]{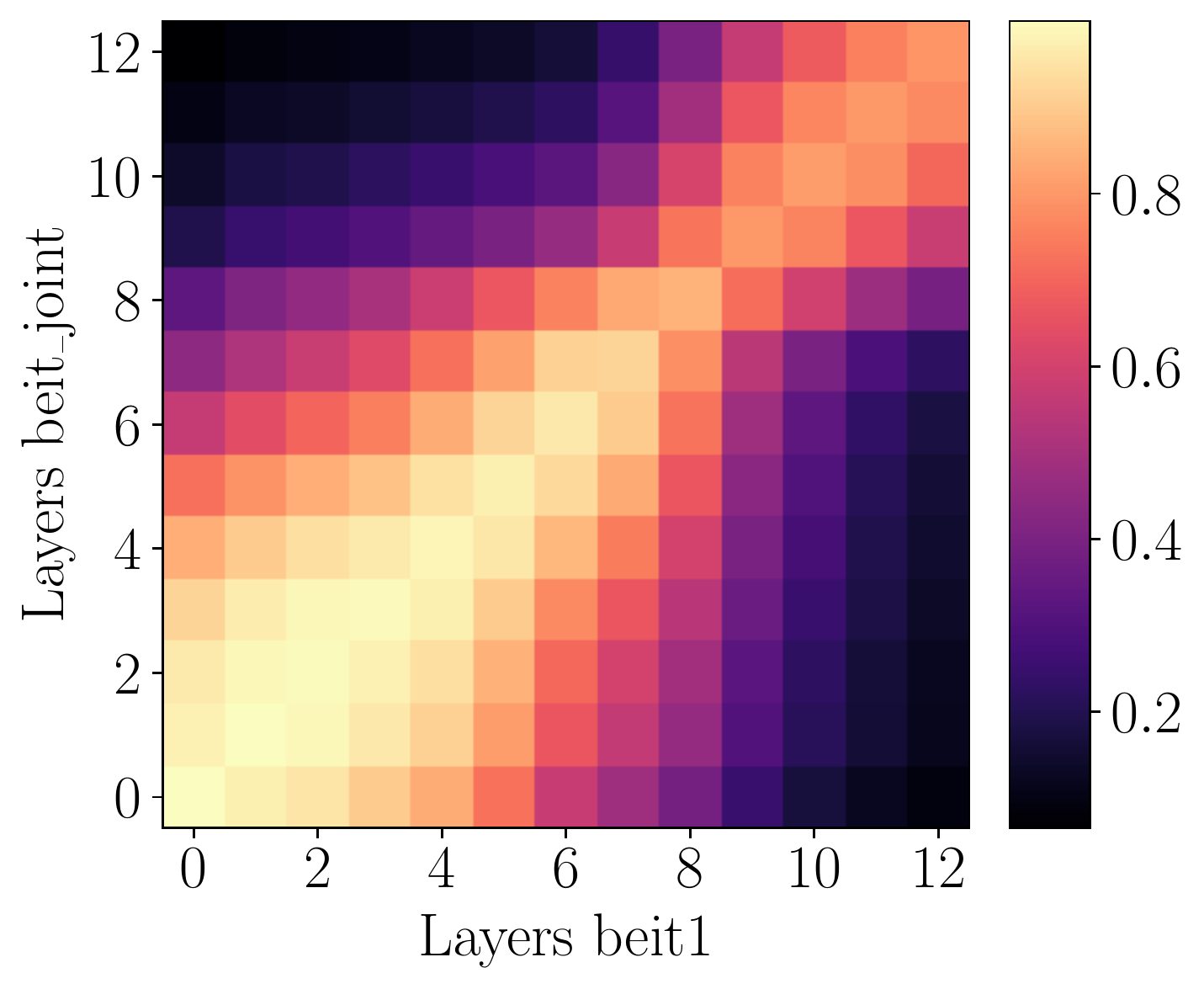}
         \caption{BEiT 2}
     \end{subfigure}
     \begin{subfigure}[b]{0.24\textwidth}
         \centering
         \includegraphics[width=\textwidth]{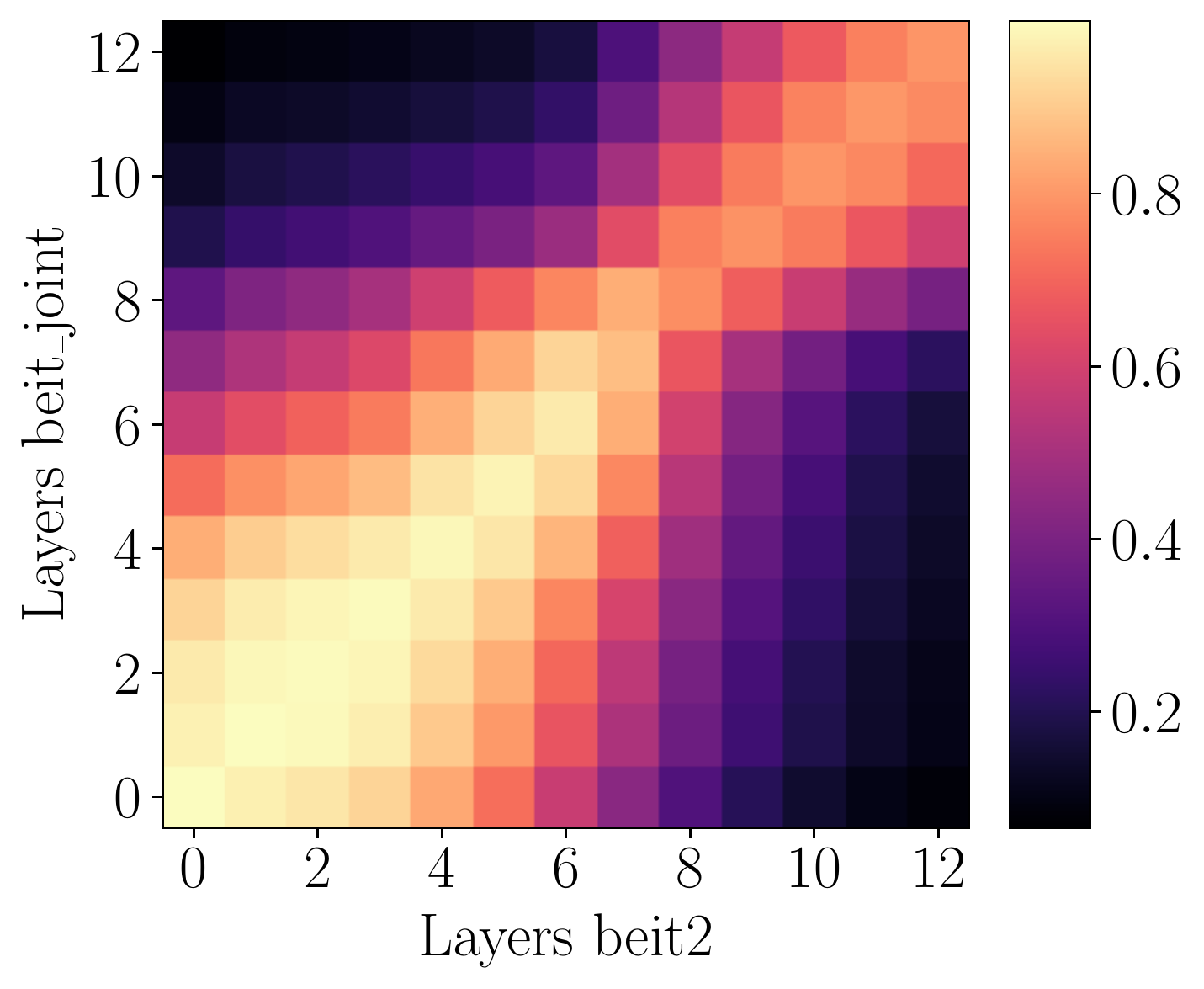}
         \caption{BEiT 3}
     \end{subfigure}
     \begin{subfigure}[b]{0.24\textwidth}
         \centering
         \includegraphics[width=\textwidth]{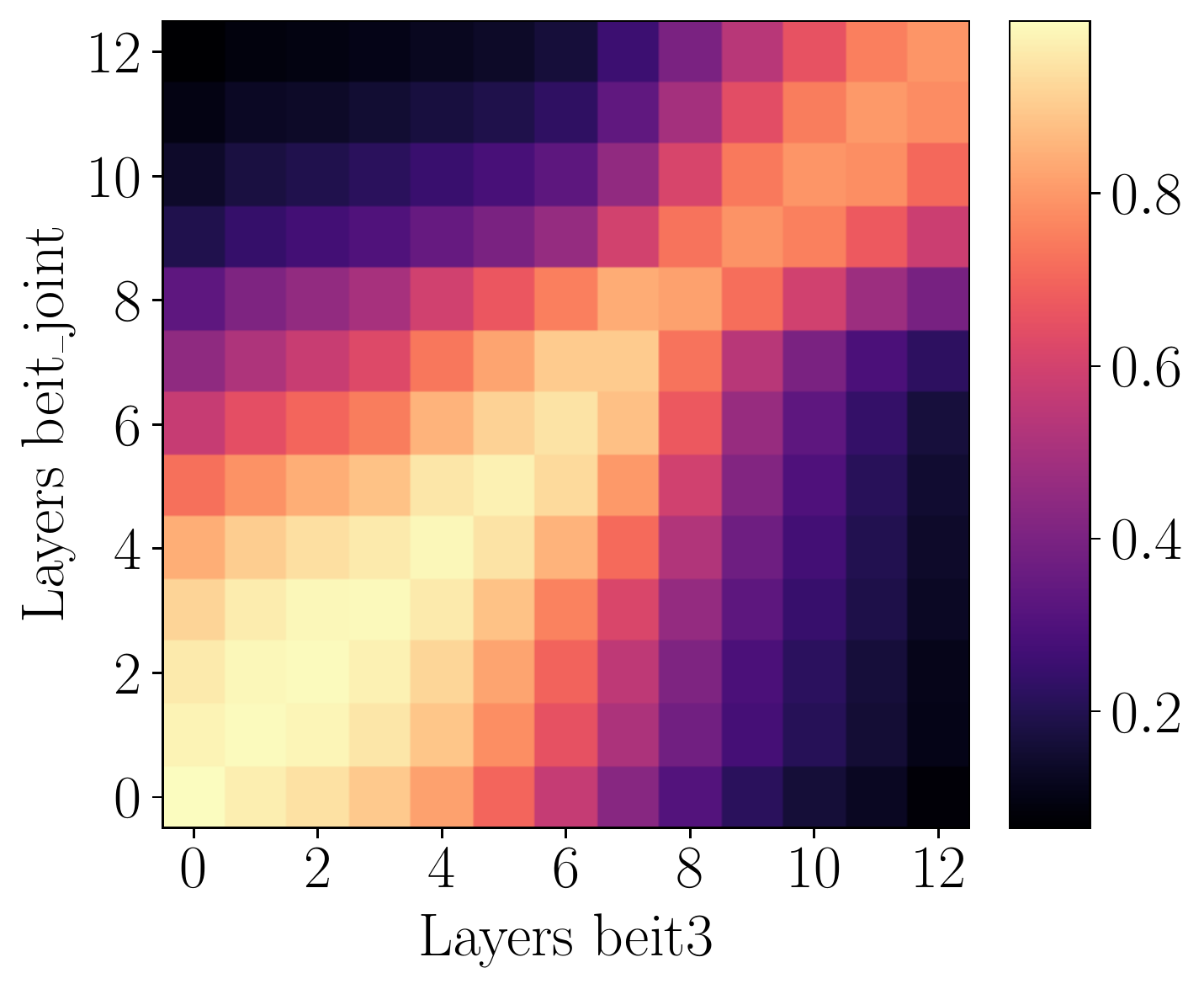}
         \caption{BEiT 4}
     \end{subfigure}
     \hfill
     \begin{subfigure}[b]{0.24\textwidth}
         \centering
         \includegraphics[width=\textwidth]{img/beit/matrix4.pdf}
         \caption{BEiT 5}
     \end{subfigure}
     \hfill
     \begin{subfigure}[b]{0.24\textwidth}
         \centering
         \includegraphics[width=\textwidth]{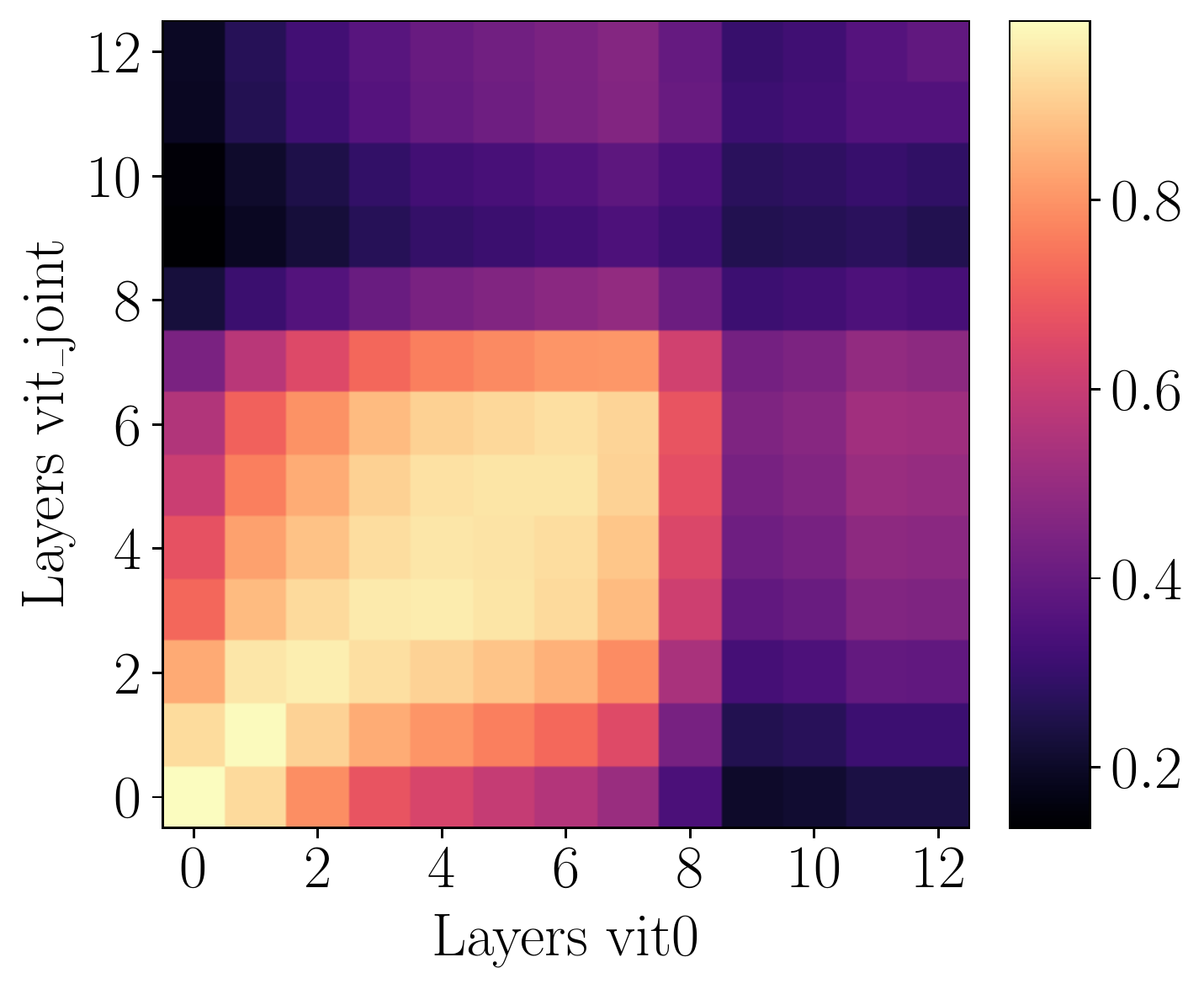}
         \caption{ViT 1}
     \end{subfigure}
     \begin{subfigure}[b]{0.24\textwidth}
         \centering
         \includegraphics[width=\textwidth]{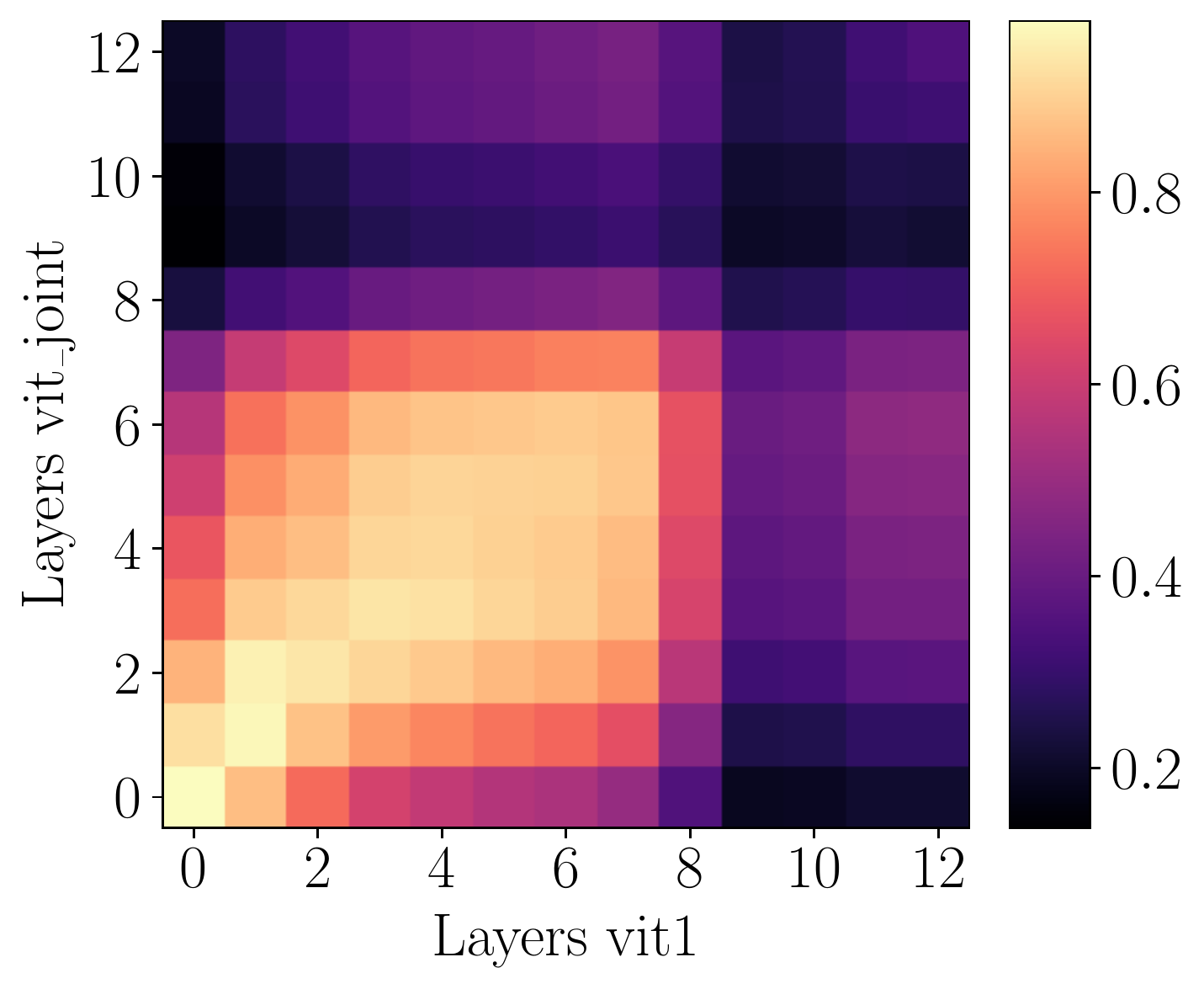}
         \caption{ViT 2}
     \end{subfigure}
     \begin{subfigure}[b]{0.24\textwidth}
         \centering
         \includegraphics[width=\textwidth]{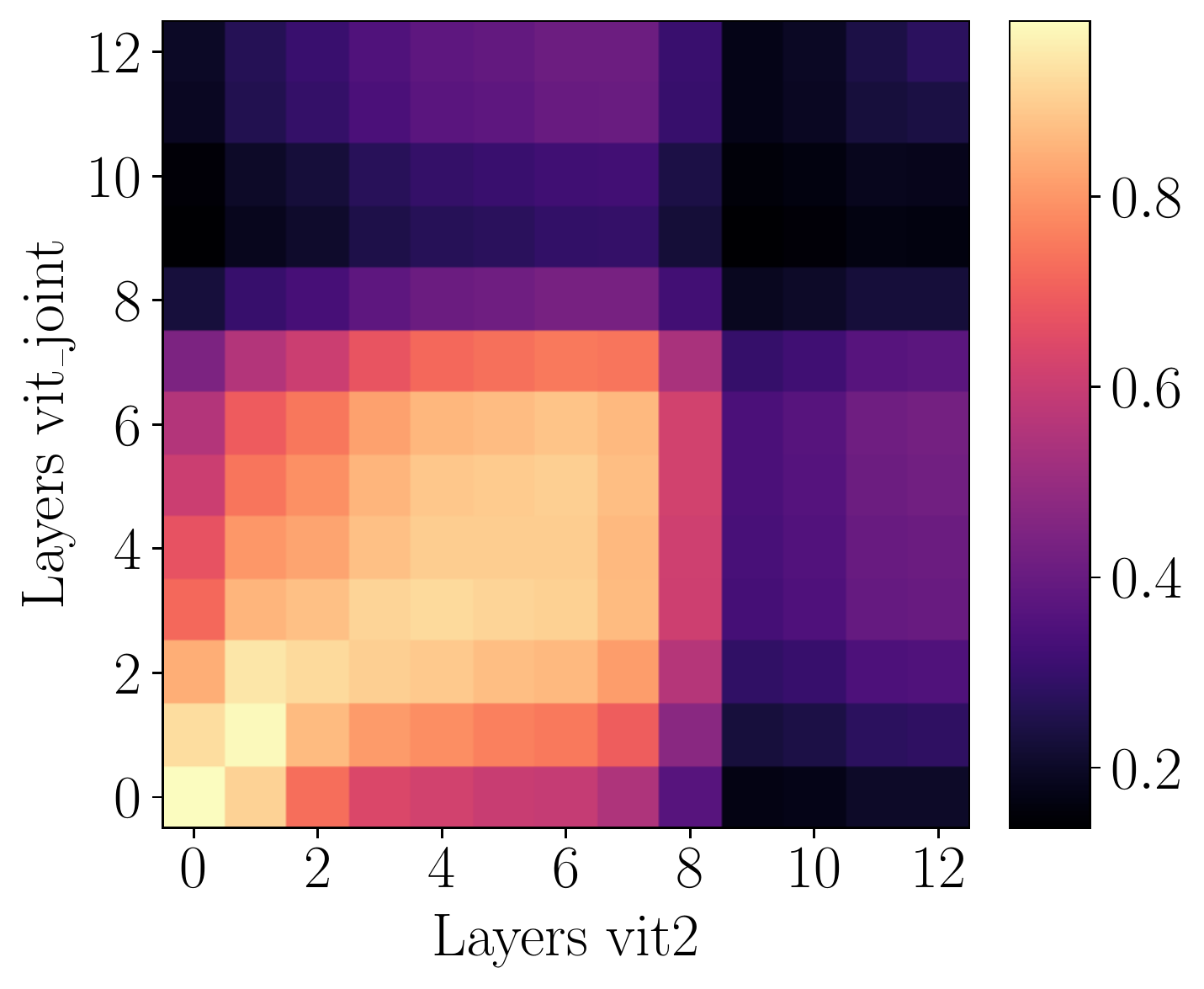}
         \caption{ViT 3}
     \end{subfigure}
     \begin{subfigure}[b]{0.24\textwidth}
         \centering
         \includegraphics[width=\textwidth]{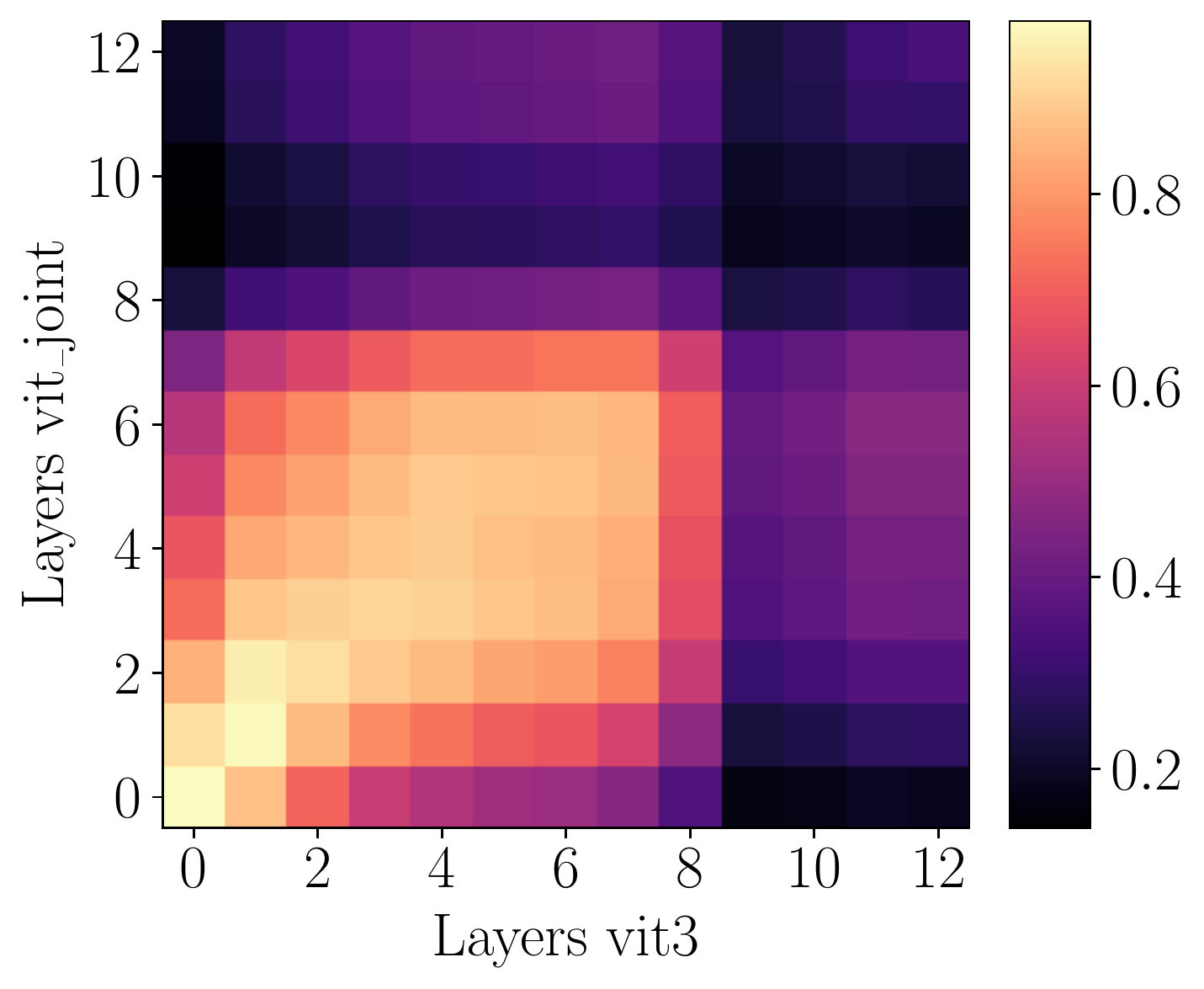}
         \caption{ViT 4}
     \end{subfigure}
     \begin{subfigure}[b]{0.24\textwidth}
         \centering
         \includegraphics[width=\textwidth]{img/vit/matrix4.pdf}
         \caption{ViT 5}
     \end{subfigure}
     \begin{subfigure}[b]{0.24\textwidth}
         \centering
         \includegraphics[width=\textwidth]{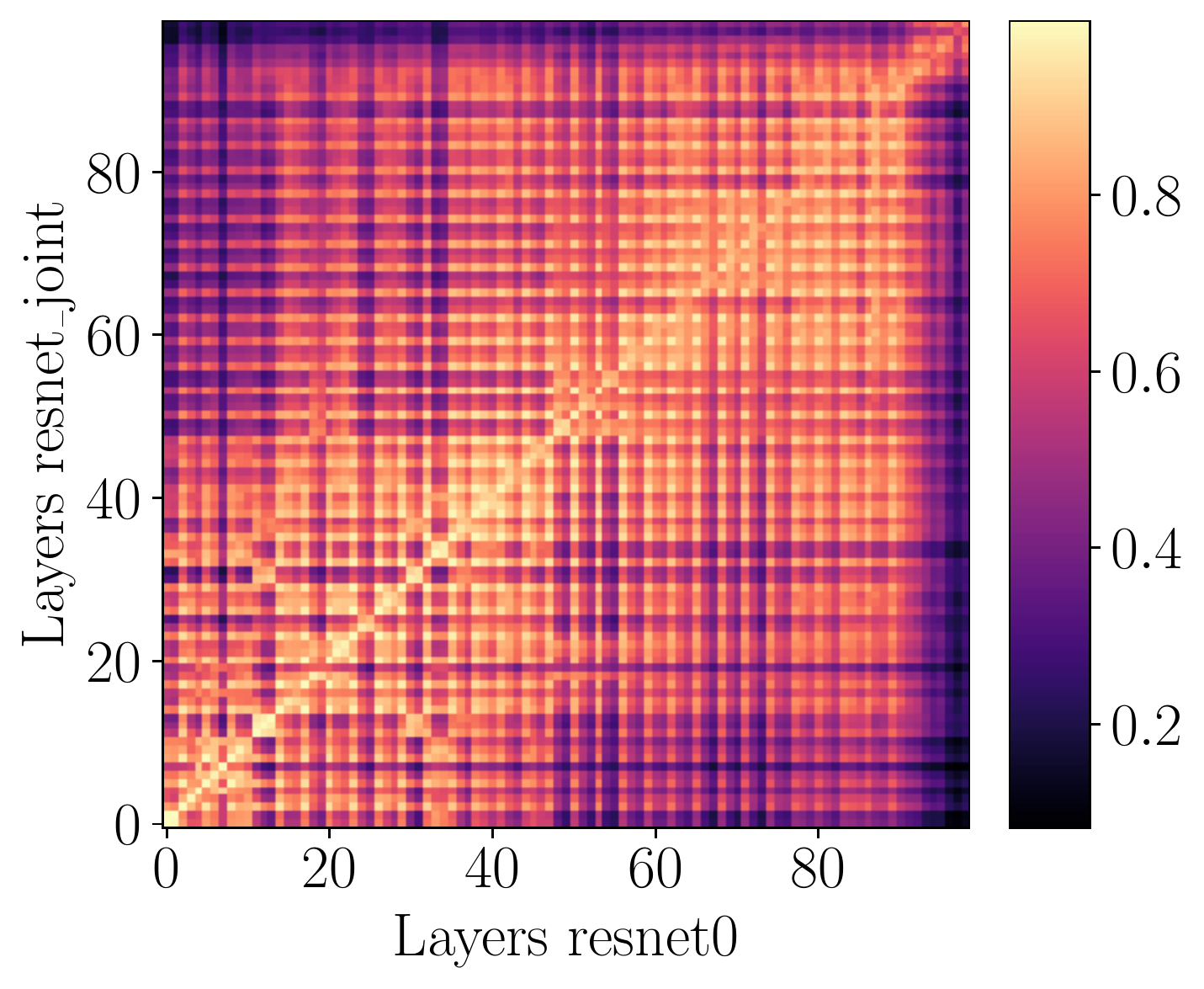}
         \caption{ResNet 1}
     \end{subfigure}
     \begin{subfigure}[b]{0.24\textwidth}
         \centering
         \includegraphics[width=\textwidth]{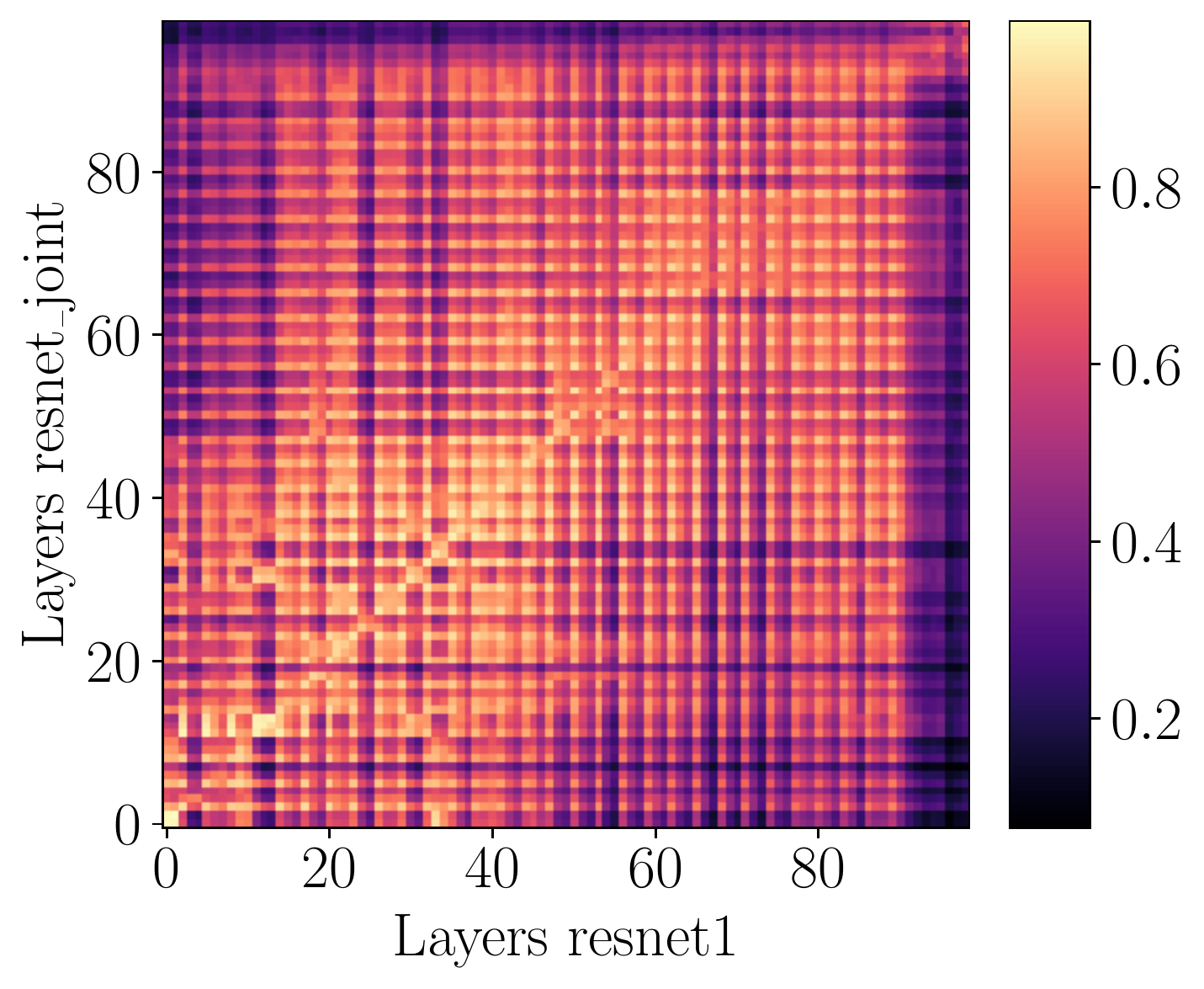}
         \caption{ResNet 2}
     \end{subfigure}
     \begin{subfigure}[b]{0.24\textwidth}
         \centering
         \includegraphics[width=\textwidth]{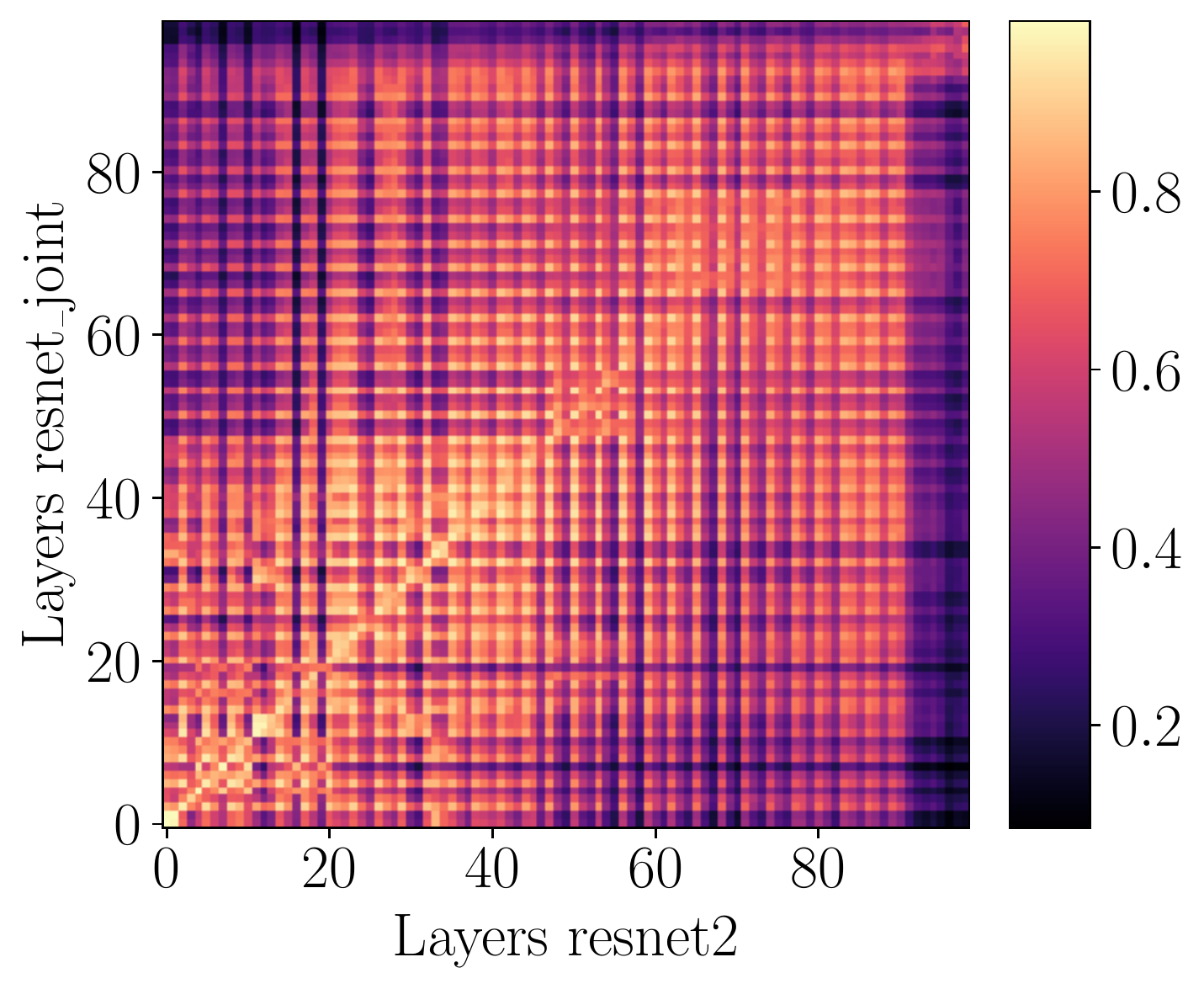}
         \caption{ResNet 3}
     \end{subfigure}
     \begin{subfigure}[b]{0.24\textwidth}
         \centering
         \includegraphics[width=\textwidth]{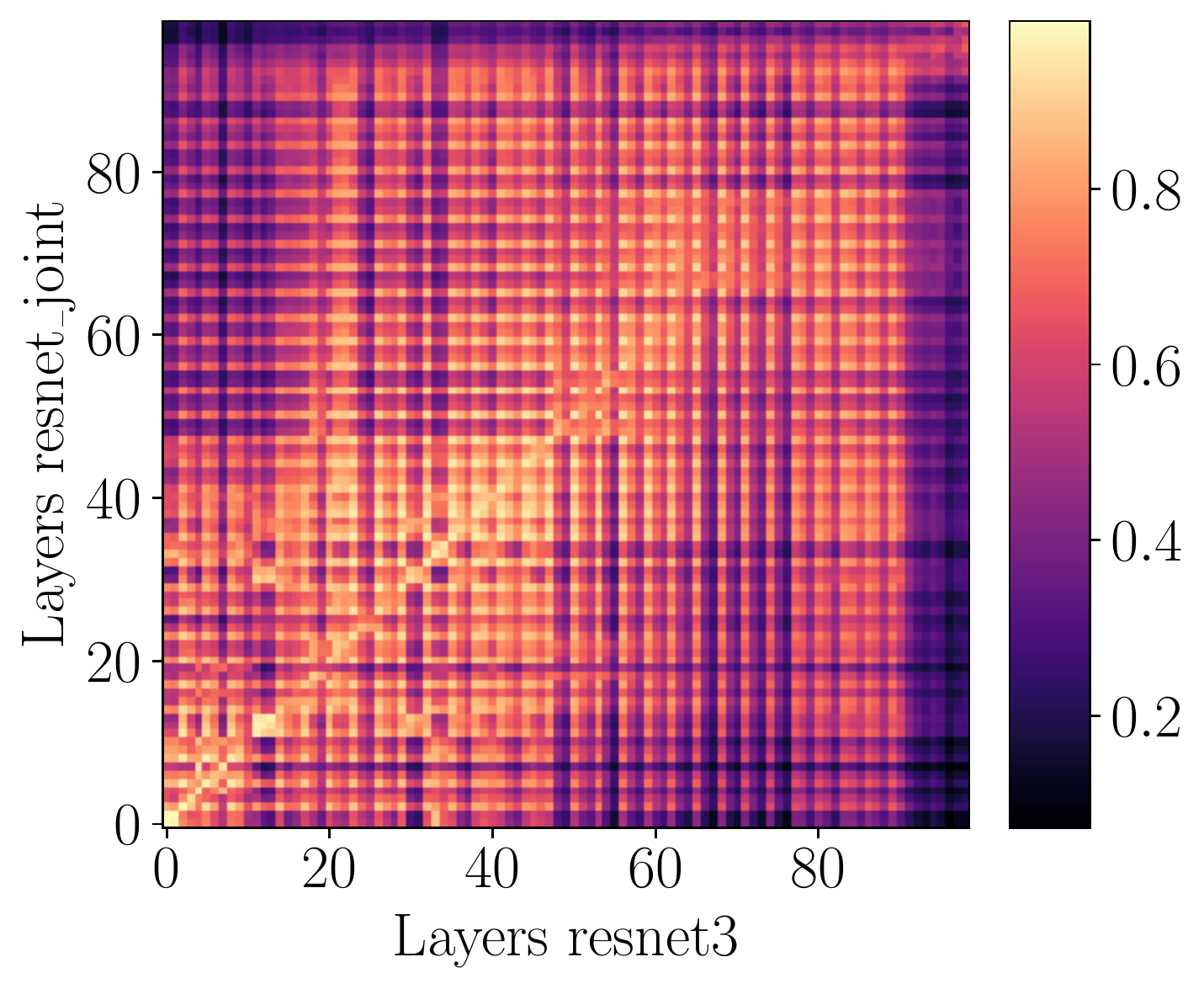}
         \caption{ResNet 4}
     \end{subfigure}
     \begin{subfigure}[b]{0.24\textwidth}
         \centering
         \includegraphics[width=\textwidth]{img/resnet/matrix4.pdf}
         \caption{ResNet 5}
     \end{subfigure}
        \caption{CKA for BEiT, Vit and ResNet. Pre-trained models after each experience are compared with the original pre-trained model.}
        \label{fig:cka-vision}
\end{figure}

CKA is computed incrementally in minibatches, following \cite{nguyen2020a}. We provide the full set of CKA plots in Figure \ref{fig:cka-nlp} for the NLP environment and in Figure \ref{fig:cka-vision} for the CV environment. We include the CKA against the original pre-trained model and its continually pre-trained version after each experience of continual pre-training. The upper-right corner of each image represents the upper layers of the models and its correlation is very low only for ViT and ResNet, while it stays large for BEiT, RoBERTa and BERT on all FC datasets.

\subsection{Experiments with Larger CV Models} \label{app:larger}
We report in Table \ref{tab:cl-inaturalist2core-large} and Table \ref{tab:cl-inaturalist2core-linear-large} the performance obtained by larger Vision Transformers models with $24$ transformers layers for fine-tuning and linear evaluation, respectively. The results are in line with our main findings with smaller models, except for the ViT, which shows a smaller degree of forgetting. However, the training curves for the large ViT shows an unstable trend: the best accuracy is reached usually after one epoch, after which the value quickly degrades to a lower performance. We believe that future works investigating the impact of model depth on our results may shed a light on this phenomenon.

\section{Continual Pre-Training Pseudocode} \label{app:pseudo}
Algorithm \ref{alg:pre-training} provides a high-level description of the continual pre-training scenario, showing the steps of continual pre-training, downstream fine-tuning and catastrophic forgetting evaluation against the FC dataset. To obtain the configuration we used in linear evaluation, it is sufficient to change \emph{fine-tune} with \emph{linear-eval} in Line \ref{alg:FC}.

\begin{algorithm}[htbp]
    \centering
    \caption{Continual Pre-training scenario}\label{alg:pre-training}
    \begin{algorithmic}[1]
        \Require Pre-trained model $h_0^{pr}$, stream of experiences $\mathcal{S} = (e_1, e_2, e_3, \ldots)$, FC dataset $\mathcal{D}^{fc}$.
        
        \State $h_0^{fc} \gets$ fine-tune($h^{pr}_0, \mathcal{D}^{fc}$) \Comment{Evaluate model on FC dataset before continual pre-training}
        
        \For{$e_i$ $\in \mathcal{S}$}
            \State $\mathcal{D}_i^{pr}$, $\mathcal{D}_i^{ds}$ $\gets$ split($\mathcal{D}_i$)
            
            \State $h_i^{pr} \gets$ pre-train($h_{i-1}^{pr}$, $\mathcal{D}_i^{pr}$) \Comment{Choose appropriate pre-train objective}
            
            \State $h_i^{ds} \gets$ fine-tune($h_{i}^{pr}$, $\mathcal{D}_i^{ds}$)
            
            \State $h_i^{fc} \gets$ fine-tune($h_{i}^{pr}$, $\mathcal{D}^{fc}$) \Comment{Evaluate model on FC dataset} \label{alg:FC}
            
            \State Compare performance of $h_i^{fc}$ with $h_0^{fc}$ to assess forgetting.
        \EndFor
        \State \Return $y$
    \end{algorithmic}
\end{algorithm}

\section{Extended Related Works} \label{app:extended-rel}
The continual pre-training scenario appeared very recently in the literature. In this section, we provide a more detailed description of the existing works exploring continual pre-training and the differences with respect to our work. Section \ref{sec:related} already provides a brief description but, due to lack of space, we were unable to thoroughly discuss the few existing studies.\\
Among existing works, the CL scenario used in \citep{jin2021a} constitutes the most similar setup with respect to our definition of continual pre-training. Like us, the authors used a dataset of research papers as pre-training stream and leveraged RoBERTa in their experiments. Differently from us, though, their work is focused on NLP tasks and on the impact that different CL strategies have on the final performance, rather than on the kind of pre-training protocol and on its impact on a separate FC task. Moreover, the downstream tasks used to measure performance are strongly related to the pre-training stream, making it difficult to understand the impact of each pre-training step on catastrophic forgetting. The results they provided show that the amount of forgetting does not depend on the specific CL strategy used. In line with our findings, a naive fine-tuning approach is robust and does not show a catastrophic loss in performance. \\
The Continual Knowledge Learning (CKL) framework \citep{jang2021} shares some similarities with the continual pre-training scenario adopted in our work. The CKL considers a pre-trained model updated continuously and, throughout its training, focuses on different objectives: recognizing invariant knowledge which does not change over time, incorporating new knowledge not present before and updating knowledge which is outdated. The proposed benchmark is entirely based on NLP: it consists of a continual pre-training dataset of news, a "time-invariant knowledge" dataset hand-crafted from relations dataset and an "updated knowledge" and "new knowledge" datasets built from scratch through Amazon Mechanical Turk and validated by a set of external experts. The empirical evaluation provided in the paper is based on a new metric, called FUAR, which condenses the performance of the pre-trained model in these three tasks into a single number. The experiments are conducted on the T5 transformer endowed with existing CL strategies. The authors found out that that parameter expansion methods are amongst the best performing ones, although they require a larger number of parameters with respect to static alternatives. \\
The study of \cite{hu2021a} focused on the impact of self-supervised pre-training on streaming data subjected to different types of drifts (some of them ascribable to existing CL scenarios like domain-incremental, data-incremental, class-incremental). The authors adopted the MoCo-v2 self-supervised technique for pre-training and a vast set of downstream tasks to measure forgetting, all belonging to CV. Importantly for our work, the authors discussed the problem of catastrophic forgetting. However, differently from our work, the evaluation is performed on the same data used for pre-training instead of relying on a separate downstream task. In our opinion, reporting results on a FC dataset better fits the continual pre-training scenario and delivers a clearer picture of the effect of continual pre-training. Nonetheless, the results obtained by \cite{hu2021a} are compatible with our findings, showing that self-supervised pre-training reduces features drift and mitigates forgetting. The CKA analysis provided by the authors, similar to ours, supports the experimental results.

\end{document}